\documentclass[11pt]{article}

\usepackage[preprint]{acl}

\usepackage{times}
\usepackage{latexsym}

\usepackage[T1]{fontenc}

\usepackage[utf8]{inputenc}

\usepackage{microtype}

\usepackage{inconsolata}

\usepackage{graphicx}

\usepackage{url}
\usepackage{booktabs}
\usepackage{multirow}
\usepackage{array}
\usepackage{makecell}
\usepackage{amsmath}
\usepackage{amssymb}
\usepackage{diagbox}
\usepackage{pifont}
\usepackage{algorithm}
\newcommand{\cmark}{\textcolor{green}{\ding{51}}}%
\newcommand{\xmark}{\textcolor{red}{\ding{55}}}%

\usepackage{xcolor}
\usepackage{tcolorbox}
\tcbuselibrary{listings, skins}

\definecolor{eclipseStrings}{RGB}{42,0,255}
\definecolor{eclipseKeywords}{RGB}{127,0,85}
\colorlet{numb}{magenta!60!black}

\lstdefinelanguage{json}{
    basicstyle=\normalfont\ttfamily\footnotesize,
    commentstyle=\color{eclipseStrings}, 
    stringstyle=\color{eclipseStrings},  
    numbers=left,
    numberstyle=\scriptsize,
    stepnumber=1,
    numbersep=8pt,
    showstringspaces=false,
    breaklines=true,
    frame=lines,
    string=[s]{"}{"},
    comment=[l]{:\ "},
    morecomment=[l]{:"},
    literate=
     *{0}{{{\color{numb}0}}}{1}
      {1}{{{\color{numb}1}}}{1}
      {2}{{{\color{numb}2}}}{1}
      {3}{{{\color{numb}3}}}{1}
      {4}{{{\color{numb}4}}}{1}
      {5}{{{\color{numb}5}}}{1}
      {6}{{{\color{numb}6}}}{1}
      {7}{{{\color{numb}7}}}{1}
      {8}{{{\color{numb}8}}}{1}
      {9}{{{\color{numb}9}}}{1}
}

\newtcblisting{jsonbox}[2][]{
  colback=gray!5!white,       
  colframe=gray!75!black,    
  title={#2},                 
  fonttitle=\bfseries,       
  listing only,
  listing options={
    language=json,           
    basicstyle=\ttfamily\footnotesize,
    breaklines=true,
    columns=fullflexible
  },
  #1
}

\usepackage{tcolorbox}
\tcbuselibrary{skins, breakable}

\newtcolorbox{promptbox}[1][]{
  colback=gray!10!white,    
  colframe=gray!60!black,   
  title={\textsc{Prompt}},  
  fonttitle=\bfseries\small,
  fontupper=\small\sffamily,
  boxrule=0.8pt,
  arc=2pt,
  #1
}

\newtcolorbox{outputbox}[1][]{
  colback=white,            
  colframe=blue!40!black,   
  title={\textsc{Model Output}},
  fonttitle=\bfseries\small,
  fontupper=\small\ttfamily, 
  boxrule=0.8pt,
  arc=2pt,
  #1
}

\newcommand{\evaluator}[1]{\noindent\textbf{\color{blue!60!black}Evaluator:} #1\par\vspace{0.1cm}}

\newcommand{\kimi}[1]{\noindent\textbf{\color{red!60!black}Kimi-K2-Thinking:} #1\par\vspace{0.1cm}}
\newcommand{\qwen}[1]{\noindent\textbf{\color{red!60!black}Qwen3-235B-A22B-Instruct-2507:} #1\par\vspace{0.1cm}}

\DeclareCaptionStyle{ruled}{labelfont=normalfont,labelsep=colon,strut=off} 
\floatstyle{ruled}
\newfloat{listing}{tb}{lst}{}
\floatname{listing}{Listing}

\usepackage{caption}
\DeclareCaptionLabelFormat{continued}{#1~#2 (cont.)}

\title{LiveFact: A Dynamic, Time-Aware Benchmark for LLM-Driven Fake News Detection}

\author{
Cheng Xu\textsuperscript{1,4} \quad 
Changhong Jin\textsuperscript{1} \quad 
Yingjie Niu\textsuperscript{1} \quad 
Nan Yan\textsuperscript{2,4} \\ 
\textbf{Yuke Mei\textsuperscript{4} \quad 
Shuhao Guan\textsuperscript{1} \quad 
Liming Chen\textsuperscript{3} \quad 
M-Tahar Kechadi\textsuperscript{1}} \\
  \textsuperscript{1}\,University College Dublin \quad 
  \textsuperscript{2}\,Georgia Institute of Technology \\ 
  \textsuperscript{3}\,Dalian University of Technology \quad 
  \textsuperscript{4}\,Bebxy\\
  \texttt{cheng.xu1@ucdconnect.ie \quad tahar.kechadi@ucd.ie} \\
}

\begin{document}
\maketitle
\begin{abstract}
The rapid development of Large Language Models (LLMs) has transformed fake news detection and fact-checking tasks from simple classification to complex reasoning. However, evaluation frameworks have not kept pace. Current benchmarks are static, making them vulnerable to benchmark data contamination (BDC) and ineffective at assessing reasoning under temporal uncertainty. To address this, we introduce \textbf{LiveFact}\footnote{\url{https://github.com/bebxy/livefact}} a continuously updated benchmark that simulates the real-world "fog of war" in misinformation detection. LiveFact uses dynamic, temporal evidence sets to evaluate models on their ability to reason with evolving, incomplete information rather than on memorized knowledge. We propose a dual-mode evaluation: Classification Mode for final verification and Inference Mode for evidence-based reasoning, along with a component to monitor BDC explicitly. Tests with 22 LLMs show that open-source Mixture-of-Experts models, such as Qwen3-235B-A22B, now match or outperform proprietary state-of-the-art systems. More importantly, our analysis finds a significant "reasoning gap." Capable models exhibit epistemic humility by recognizing unverifiable claims in early data slices-an aspect traditional static benchmarks overlook. LiveFact sets a sustainable standard for evaluating robust, temporally aware AI verification.
\end{abstract}

\begin{figure}[t]
    \centering
    \includegraphics[width=1\linewidth]{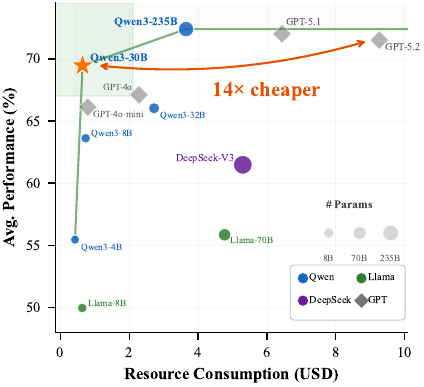}
    \caption{Cost-performance trade-off on LiveFact (Nov. 2025). Qwen3-235B-A22B-Instruct-2507 achieves the best performance (72.4\%), while Qwen3-30B-A3B-Instruct-2507 provides optimal cost-efficiency at 14× lower cost than comparable GPT models.}
    \label{fig:main}
\end{figure}

\section{Introduction}

Fake news detection has long been a foundational task in Natural Language Processing (NLP) and information verification, evolving from simple linguistic analysis to complex system verifications \citep{guo2022survey}. 
Early research treated this as a binary classification problem, relying heavily on feature engineering, such as writing style analysis, word-frequency statistics, and sentiment patterns \citep{zhou2020survey}. As social media platforms proliferated, the scope expanded to include propagation analysis, where researchers modeled the spread of information across networks to identify anomalies \citep{rocha2021impact,sadeghi2022fake,mishra2020fake,chen2023rumor}. These traditional approaches laid the groundwork for automated verification, but they were largely constrained by reliance on static features that cannot capture the semantic depth required for effective detection of fake news \citep{xu2025ssa,xu2025analysis}.

The advent of Large Language Models (LLMs) has fundamentally transformed this landscape \citep{openai2024gpt4,anthropic2024introducing,geminiteam2024gemini}, shifting the paradigm from pattern recognition to semantic reasoning \citep{zhang2024llm,havrilla2024glore,tong2025generate}. Modern LLMs possess unprecedented capabilities in knowledge retrieval, context understanding, and logical inference, enabling them to approach fact-checking with a level of sophistication that mirrors human cognition \citep{Zhang2025is,xiao2026fact,ma2024fake}. The task has consequently evolved from simple "true/false" classification into a complex reasoning challenge. Systems are now required to verify claims against multi-hop evidence, identify contradictions in long-form context, and provide interpretable explanations for their verdicts \citep{liao2023muser,atanasova2020generating,alhindi2018evidence}. This shift has raised the ceiling for what is possible in automated fake news detection.

However, there is a critical misalignment between the rapid advancement of LLMs and the stagnation of the benchmarks used to evaluate them \cite{xu2025triplefact}. The field of fake news detection has not kept pace with the generative AI revolution, leading to a "capability-evaluation gap." Current mainstream benchmarks remain static, offering fixed snapshots of information that cannot challenge the dynamic capabilities of modern models (e.g., OpenAI's latest GPT-5.2 model has a knowledge cutoff of August 2025\footnote{\url{https://platform.openai.com/docs/models/gpt-5.2}}). They typically present a "God-view" scenario in which all relevant evidence is perfectly aggregated, ignoring the temporal nuances and information scarcity, which define real-world fake news detection. Furthermore, because these static datasets are frequently ingested into the massive training corpora of LLMs, they carry high risks of Benchmark Data Contamination (BDC) \citep{xu2024benchmark,sun2025the}, making it difficult to discern whether a model is genuinely reasoning or merely reciting memorized training data \citep{lee2022deduplicating,sainz2023nlp,zhou2023dont,mcintosh2024inadequacies,deng2024unveiling}.

The root cause of these limitations lies in the static nature of dataset construction, which directly conflicts with the evolving nature of news and the continuous pre-training of LLMs \citep{white2025livebench,chen2025recent}. To bridge this gap, we must fundamentally rethink evaluation paradigms. A robust solution requires transitioning from static archives to dynamic, living benchmarks that continuously update with the news cycle. Furthermore, evaluation must move beyond simple fact retrieval to test reasoning under uncertainty, simulating the "fog of war" where evidence is incomplete or evolving. Only by controlling for time and BDC can we accurately measure an LLM's true utility as a fake news detection assistant.

To address these issues, we introduce \textbf{LiveFact}, a benchmark designed to rigorously evaluate fake news detection with LLMs in a dynamic environment. LiveFact specifically contributes the following:
\begin{enumerate}
   \item {\bf Dynamic Benchmarking:} We continuously update the dataset on a monthly basis using the latest news, ensuring zero-shot evaluation on unseen events and minimizing BDC risk.
   \item {\bf Sustainable Contamination Monitoring:} We employ Semantic Sensitivity Amplifier (SSA) framework \citep{xu2025ssa} with entity shift mechanism to explicitly measure and mitigate memorization risks.
   \item {\bf Fine-Grained Temporal Evidence:} We utilize dynamic time-sliced evidence sets (e.g., $E^{(-3)}$ vs. $E^{(+3)}$) to simulate real-world information evolution.
   \item {\bf Dual-Mode Evaluation:} We employ both Inference Mode (reasoning based on available evidence) and Classification Mode (knowledge-based verification) to disentangle reasoning skills from internal knowledge.
\end{enumerate}

\begin{table*}[t]
    \centering
    \resizebox{0.98\textwidth}{!}{%
    \begin{tabular}{ll|lccccc}
    \toprule
        \textbf{Benchmark} & \textbf{Author} & \textbf{Type} & \textbf{Update} & \textbf{Evidence} & \textbf{Temporal} & \textbf{BDC Control} & \textbf{Evaluation Focus} \\
        \midrule
        LIAR & \citet{wang2017liar} & Static & \xmark & \xmark & \xmark & \xmark & Classification \\
        LIAR-PLUS & \citet{alhindi2018evidence} & Static & \xmark & \cmark & \xmark & \xmark & Classification \\
        FEVER & \citet{thorne2018fever} & Static & \xmark & \cmark & \xmark & \xmark & Classification \\
        HotpotQA & \citet{yang2018hotpotqa} & Static & \xmark & \cmark & \xmark & \xmark & Reasoning\\
        NELA-GT-2018 & \citet{nørregaard2019nela} & Static & \xmark & \xmark & \xmark & \xmark & Classification \\
        FakeNewsNet & \citet{shu2020fakenewsnet} & Static & \xmark & \xmark & \xmark & \xmark & Classification \\
        HoVer & \citet{jiang2020hover} & Static & \xmark & \cmark & \xmark & \xmark & Reasoning\\
        PolitiHop & \citet{ostrowski2021multi} & Static & \xmark & \cmark & \xmark & \xmark & Reasoning \\
        CHECKED & \citet{yang2021checked} & Static & \xmark & \cmark & \cmark & \xmark & Classification \\
        Weibo23 & \citet{liu2024bc} & Static & \xmark & \cmark & \xmark & \xmark & Classification \\
        MCFEND & \citet{li2024mcfend} & Static & \xmark & \cmark & \xmark & \xmark & Classification \\
        FineFake & \citet{zhou2024finefake} & Static & \xmark & \cmark & \xmark & \xmark & Classification \\
        MultiHoax & \citet{shafiei2025multihoax} & Static & \xmark & \cmark & \xmark & \xmark & Reasoning \\
        MMFakeBench & \citet{liu2025mmfakebench} & Static & \xmark & \cmark & \xmark & \xmark & Classification \\
        MPPFND & \citet{zhao2025mppfnd} & Static & \xmark & \cmark & \cmark & \xmark & Classification  \\
        AdvFake & \citet{chen2025real} & Dynamic & \xmark & \cmark & \cmark & Medium & Classification \\
        TripleFact & \citet{xu2025triplefact} & Dynamic & \xmark & \cmark & \xmark & Medium & Classification \\
        LiveFact & Ours & Dynamic & \cmark & \cmark & \cmark & High & Reasoning \& Classification \\
         \bottomrule
    \end{tabular}
    }
    \caption{Comparison of LiveFact with existing representative fake news detection benchmarks.}
    \label{tab:relatedWork}
\end{table*}

\section{Related Work}
\label{sec:related_work}

The early era of fake news detection was defined by static datasets and feature-based learning. Pioneering benchmarks such as LIAR \citep{wang2017liar} and FakeNewsNet \citep{shu2020fakenewsnet} provided labeled short statements from political contexts, enabling classifiers that rely on linguistic cues and metadata. Subsequent iterations, such as LIAR-PLUS \citep{alhindi2018evidence} and LIAR2 \citep{xu2024enhanced,xu2023fuzzy}, attempted to add evidence context, while other datasets focused on social context and propagation graphs \citep{meyers2020fake,gong2023fake,zhu2024propagation,si2023exploring,huang2025multi,cao2025enhancing,zheng2025mrr}. While foundational, these benchmarks are not adequate for LLMs. They are typically small-scale, domain-specific (often limited to U.S. politics), and, most critically, static. Thus, modern LLMs, which have likely seen this data during pre-training, can solve them via memorization rather than reasoning \citep{white2025livebench}. Furthermore, the evidence in these older datasets is often simplistic, lacking the multi-hop complexity required to evaluate current generative models \citep{ostrowski2021multi}.

As LLMs have become widely used, research has shifted toward more complex, reasoning-intensive frameworks. MultiHoax \citep{shafiei2025multihoax} introduced false-premise questions to test critical thinking, while MMCV \citep{wang2025piecing} extended verification into the multimodal domain. MUSER \citep{liao2023muser} and MSynFD \citep{xiao2024msynfd} introduced multi-step retrieval and syntax-aware mechanisms to handle complex narrative structures. Similarly, EX-FEVER \citep{ma2024ex} emphasized explainability, requiring models to generate reasoning paths. While these works represent a significant leap forward in task complexity, they still rely on static data snapshots, leaving them vulnerable to the rapid obsolescence and contamination inherent in fixed datasets.

Despite these advancements, we face a "validity crisis" stemming from BDC and the lack of temporal realism \citep{xu2025dcr,zhang2024causal}. Static benchmarks cannot guarantee that a model is reasoning on unseen data and do not capture the temporal constraints of real-world verification \citep{walter2020fact}.

Recently, AdvFake \citep{chen2025real} tried to address BDC problem dynamically through real-time adversarial generation to stress-test RAG systems. While sharing our dynamic philosophy, LiveFact differs fundamentally: rather than focusing on adversarial deception, we isolate temporal reasoning and epistemic humility through strictly controlled time-sliced evidence (the "Fog of War").

The benchmark most closely aligned with LiveFact's philosophy is LiveBench\footnote{\url{https://livebench.ai/}} \citep{white2025livebench}, which uses a regularly updated design to mitigate contamination. However, LiveBench is not designed for fake news detection; instead, it evaluates general LLM capabilities such as coding, mathematics, and data analysis. Consequently, it lacks the specific evidential reasoning structures required to verify complex claims in the misinformation domain. In the realm of fake news detection, the most recent attempt is TripleFact \citep{xu2025triplefact}. However, it remains a highly limited conceptual framework. Due to strict data copyright issues, the TripleFact dataset cannot be publicly released, and it lacks critical components such as continuous updates and temporal evidence slicing. Additionally, other recent benchmarks like FreeEval \citep{yu2024freeeval} and TreeEval \citep{li2025treeeval} incorporate advanced evaluation methodologies, but they do not address the specific constraints of the fake news detection domain.

To the best of our knowledge, as shown in Table \ref{tab:relatedWork}, LiveFact is the first fake news detection benchmark to simultaneously address dynamic updates, contamination control, and temporal reasoning, filling the critical gaps left by existing works.

\begin{figure*}
    \centering
    \includegraphics[width=1\linewidth]{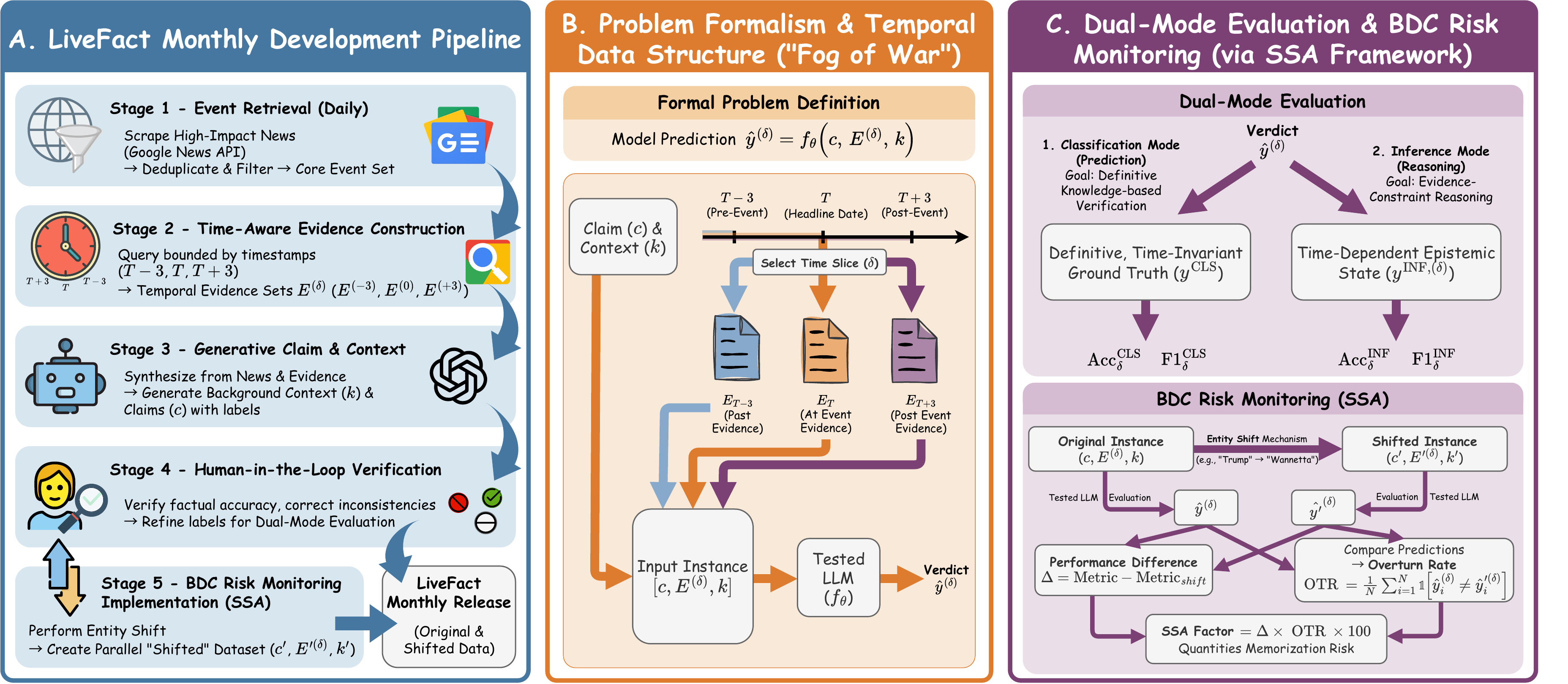}
    \caption{The overall framework of the LiveFact Benchmark. (A) The Monthly Development Pipeline illustrates the continuous process of acquiring real-time events, generating claims and context via LLMs, and performing human verification. (B) The Problem Formalism specifies the task as conditional reasoning under temporal constraints, utilizing time-sliced evidence sets (e.g., Pre-Event $E^{(-3)}$ vs. Post-Event $E^{(+3)}$) to simulate the "fog of war." (C) The Evaluation framework details the Dual-Mode approach (separating prediction vs. reasoning capabilities) and the integration of the SSA Framework to quantify BDC risk via entity shift mechanism.}
    \label{fig:livefact}
\end{figure*}

\section{The LiveFact Benchmark}

\subsection{Motivation}

LiveFact is designed around three core principles that address the limitations identified in Section~\ref{sec:related_work}. First, to combat BDC, we implement \textit{continuous monthly updates} using the latest news events, ensuring evaluation data remain unseen during model pre-training. Second, to assess genuine reasoning rather than knowledge retrieval, we introduce a \textit{temporal evidence structure} that simulates the "fog of war"---providing models with evidence sets of varying completeness ($E^{(\delta)}$) to test their ability to recognize information gaps. Third, to disentangle reasoning from memorization, we employ a \textit{dual-mode evaluation}: Classification Mode tests factual accuracy against ground truth, while Inference Mode tests whether models appropriately predict "Ambiguous" when evidence is insufficient. These principles are formalized in the following section.

\subsection{Problem Definition}

We formalize the LiveFact task as a conditional reasoning problem under temporal constraints. Let $\mathcal{E} = \{e_1, e_2, \ldots, e_M\}$ denote a set of $M$ news events, where each event $e_j$ is associated with a headline date $T_j$. For each event $e_j$, we construct a set of claims $\mathcal{C}_j = \{c_{j,1}, c_{j,2}, \ldots\}$, where each claim $c \in \mathcal{C}_j$ is a statement about event $e_j$. We denote the full dataset as $\mathcal{D} = \{(c_i, e(c_i), k_i, y_i)\}_{i=1}^{N}$, consisting of $N$ claim instances, where $e(c_i)$ denotes the event associated with claim $c_i$, $k_i$ is the static background context describing the fundamental entities involved (e.g., "Donald Trump is the current President of the United States..."), and $y_i \in \mathcal{Y}$ is the ground-truth label with $\mathcal{Y} = \{\textsc{Real}, \textsc{Fake}, \textsc{Ambiguous}\}$.

To simulate the "fog of war" inherent in real-time verification, we construct time-sliced evidence sets based on the headline date of each event. For an event with headline date $T$, we define $E^{(\delta)}$ as the evidence set containing only information available up to day $T + \delta$, where $\delta \in \{-3, 0, +3\}$ represents the temporal offset in days. For instance, $E^{(-3)}$ contains evidence published at least three days before the event, $E^{(0)}$ includes evidence up to the headline date, and $E^{(+3)}$ incorporates evidence available three days after the event.

Given a LLMs $f_\theta$ parameterized by $\theta$, the model predicts a verdict based on the claim, time-sliced evidence, and background context:
\begin{equation}
    \hat{y}_i^{(\delta)} = f_\theta\bigl(c_i,\, E_i^{(\delta)},\, k_i\bigr)
\end{equation}
where $E_i^{(\delta)}$ denotes the evidence set at temporal offset $\delta$ for the event associated with claim $c_i$, and $\hat{y}_i^{(\delta)} \in \mathcal{Y}$ is the predicted label.

We evaluate models under two distinct modes:
\begin{itemize}
    \item \textbf{Classification Mode} ($m = \textsc{cls}$): The ground-truth label $y_i^{\textsc{cls}}$ is the definitive, time-invariant verification label representing the ultimate factual status of the claim.
    \item \textbf{Inference Mode} ($m = \textsc{inf}$): The ground-truth label $y_i^{\textsc{inf},(\delta)}$ is dynamic, determined by whether the partial evidence $E_i^{(\delta)}$ is sufficient to support a correct judgment at the given temporal offset.
\end{itemize}

Let $m \in \{\textsc{cls}, \textsc{inf}\}$ denote the evaluation mode, and let $y_i^{m,(\delta)}$ denote the corresponding ground-truth label (where $y_i^{\textsc{cls},(\delta)} = y_i^{\textsc{cls}}$ is time-invariant for Classification Mode). We measure performance using accuracy and macro-averaged F1 score:
\begin{equation}
    \mathrm{Acc}_{\delta}^{m} = \frac{1}{N} \sum_{i=1}^{N} \mathbb{1}\bigl[\hat{y}_i^{(\delta)} = y_i^{m,(\delta)}\bigr]
\end{equation}
\begin{equation}
    \mathrm{F1}_{\delta}^{m} = \frac{1}{|\mathcal{Y}|} \sum_{y \in \mathcal{Y}} \frac{2 \cdot P_y^{m,(\delta)} \cdot R_y^{m,(\delta)}}{P_y^{m,(\delta)} + R_y^{m,(\delta)}}
\end{equation}
where $\mathbb{1}[\cdot]$ is the indicator function that returns 1 if the condition is true and 0 otherwise, and $P_y^{m,(\delta)}$ and $R_y^{m,(\delta)}$ denote precision and recall for class $y \in \mathcal{Y}$ under mode $m$ at temporal offset $\delta$, respectively.

To monitor the risk of BDC, we integrate the SSA framework~\citep{xu2025ssa}. For each instance, we generate a counterfactual claim $c'_i$ via an Entity Shift mechanism, along with corresponding shifted evidence $E'^{(\delta)}_i$ and context $k'_i$. Key entities are replaced with semantically equivalent but fictional alternatives (e.g., "Trump" $\rightarrow$ "Wannetta") to isolate reasoning from memorization. Let $\hat{y}'^{(\delta)}_i = f_\theta(c'_i, E'^{(\delta)}_i, k'_i)$ denote the prediction on the shifted instance.

We define the \textit{Overturn Rate} (OTR) as the proportion of instances where the model's prediction changes solely due to entity replacement:
\begin{equation}
    \mathrm{OTR} = \frac{1}{N} \sum_{i=1}^{N} \mathbb{1}\bigl[\hat{y}_i^{(\delta)} \neq \hat{y}'^{(\delta)}_i\bigr]
\end{equation}
Let $\Delta$ denote the performance gap between the original and shifted evaluations:
\begin{equation}
    \Delta = \mathrm{Metric} - \mathrm{Metric}_{\textit{shift}}
\end{equation}
where $\mathrm{Metric}$ refers to any evaluation metric (e.g., accuracy or F1 score). The final contamination indicator is:
\begin{equation}
    \mathrm{SSA\text{ }Factor} = \Delta \times \mathrm{OTR} \times 100
\end{equation}
A high SSA-Factor indicates that the model relies on spurious correlations or memorized entity-specific knowledge rather than robust evidence-based reasoning.

\subsection{LiveFact Development Pipeline}

As shown in Figure \ref{fig:livefact}, the construction of LiveFact is a rigorous, multi-stage process designed to ensure data freshness, accuracy, and depth. The pipeline is automated for monthly updates while incorporating human oversight to guarantee quality. For this inaugural iteration, our data collection spanned November 2025.

\subsubsection{Stage 1: Event Retrieval}

We initiate the pipeline by collecting the latest high-impact news events. Using the Google News API, we systematically scrape trending events from the "World" section daily at 00:00 GMT. This ensures a consistent global scope for our dataset. To maintain dataset quality and reduce redundancy, we apply a rigorous deduplication and filtering process, removing overlapping storylines and low-relevance items. For the November 2025 cycle, this process yielded a core Event Set of 737 distinct events.

\subsubsection{Stage 2: Time-Aware Evidence Construction}

To support our temporal reasoning analysis ("Fog of War"), we construct evidence sets anchored to specific time offsets relative to the event's headline date ($T$). We specifically target three distinct temporal offsets: $\delta=-3$ (three days prior), $\delta=0$ (the event headline date), and $\delta=+3$ (three days post-event). Leveraging Google APIs, we query for information strictly bounded by these timestamps, resulting in three discrete evidence sets for each event: $E^{(-3)}$, $E^{(0)}$, and $E^{(+3)}$. In total, we gathered 25,064 distinct pieces of evidence across the 737 events, providing a rich informational backdrop for temporal analysis. Further technical specifications are detailed in Appendix \ref{sec:evidence}.

\subsubsection{Stage 3: Claim and Context Generation}

Once the raw data are aggregated, we employ OpenAI's o4-mini\footnote{\url{https://openai.com/index/introducing-o3-and-o4-mini/}}, one of the most advanced reasoning models available, to synthesize the testing components. We input the main news headline and all associated time-stamped evidence into the model to generate two key outputs: (1) a concise background Context for the entities involved, and (2) a set of Claims categorized into three labels: Real, Ambiguous, and Fake. For the current dataset, this stage produced a total of 4,392 test sample claims, distributed initially as: "Real" (1,468), "Fake" (1,451), and "Ambiguous" (1,473). This balanced generation ensures the model is tested against diverse truth values. Detailed generation prompts and parameters are provided in Appendix \ref{sec:generation}.

\subsubsection{Stage 4: Human-in-the-Loop Verification}

To ensure the reliability of our ground truth, all generated content undergoes a strict human review process. Annotators verify the factual accuracy of the generated claims against the retrieved evidence and correct any inconsistencies.

Crucially, this stage involves refining the labels for our Inference Mode. In this mode, the ground truth is dynamically adjusted based on whether the provided evidence slice ($E^{(-3)}$, $E^{(0)}$, or $E^{(+3)}$) is sufficient for a human expert to make a correct judgment. For instance, a claim that is factually "Real" might be labeled "Ambiguous" at $T-3$ if the confirming evidence had not yet emerged. This adjustment results in significant shifts in label distribution across time slices, as shown in Table \ref{tab:disctribution}.

\begin{table}[h]
    \centering
    \resizebox{0.48\textwidth}{!}{%
    \begin{tabular}{l|ccccc}
    \toprule
         \textbf{Evidence Set} & \textbf{Real} & \textbf{Fake} & \textbf{Ambiguous} & \textbf{Total Claims} \\
        \midrule
        $E^{(-3)}$ & 346 & 348 & 3,698 & 4,392 \\
        $E^{(0)}$ & 1,404 & 1,390 & 1,598 & 4,392 \\
        $E^{(+3)}$ & 1,460 & 1,443 & 1,489 & 4,392 \\
    \bottomrule
    \end{tabular}
    }
    \caption{Dynamic Label Distribution for Inference Mode (November 2025)}
    \label{tab:disctribution}
\end{table}

As illustrated, the $\delta=-3$ set is heavily skewed toward "Ambiguous," reflecting the high uncertainty of the "Fog of War" period. As time progresses to $\delta=0$ and $\delta=+3$, the distribution normalizes as more evidence becomes available. This evolving distribution is a key feature of LiveFact, designed to test temporal adaptability. The complete protocol for manual review is documented in Appendix \ref{sec:generation}.

\begin{figure}[t]
    \centering
    \includegraphics[width=1.0\linewidth]{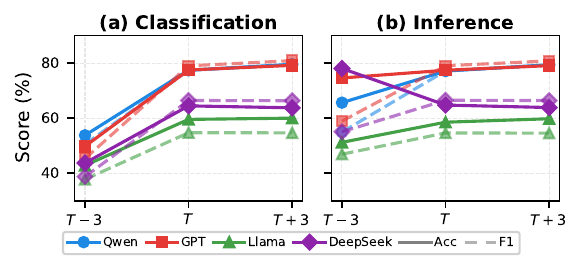}
    \caption{Temporal performance evolution for select model families across $\delta \in \{-3, 0, +3\}$. Panel (a) shows Classification Mode, where scores drop at $\delta=-3$ due to the lack of definitive evidence. Panel (b) shows Inference Mode, where robust models recover accuracy by correctly predicting "Ambiguous," narrowing the performance gap.}
    \label{fig:temporal_evolution}
\end{figure}

\subsubsection{Stage 5: BDC Risk Monitoring}

Finally, we integrate the contamination monitoring component by adopting the workflow from the SSA framework \citep{xu2025ssa}. To avoid over-reliance on OpenAI models that could lead to potential preference leakage \citep{li2025preference}, we employ the Qwen3-235B-A22B model \citep{yang2025qwen3} at this stage, we perform Entity Shift on the finalized Event Set, Claim Set, and Evidence Set. This process replaces specific named entities with fictional or neutral counterparts while preserving the narrative structure. By evaluating models on this parallel "shifted" dataset, we calculate the performance difference and OTR to derive the SSA Factor, providing a quantitative measure of BDC risk for each release cycle. Further details regarding this stage are provided in Appendix \ref{sec:ssa_implementation}.

\subsection{Evaluation Metrics}

To rigorously evaluate model performance across different modes and time slices, we employ a comprehensive set of 12 distinct metrics.

We calculate Accuracy (Acc) and F1-Macro Score (F1) for both evaluation modes (Classification, $m=\textsc{cls}$, and Inference, $m=\textsc{inf}$) at each of the three temporal offsets ($\delta \in \{-3, 0, +3\}$).

While Accuracy provides a general view of performance, it can be misleading when classes are imbalanced. As demonstrated in Stage 4 (Table \ref{tab:disctribution}), the manual adjustment for Inference Mode creates a severe class imbalance at $\delta=-3$, where "Ambiguous" labels account for nearly 85\% of the data. A model that simply guesses "Ambiguous" for every query at $\delta=-3$ would achieve a high accuracy score without performing any genuine reasoning. The F1-Macro score treats all classes equally, preventing the majority class from dominating the score. This allows us to rationally evaluate whether a model can correctly identify the rare "Real" or "Fake" instances that are verifiable at early stages, ensuring a robust assessment of reasoning capabilities under unbalanced conditions.

\begin{table*}[t]
    \centering
    \resizebox{1.00\textwidth}{!}{%
    \begin{tabular}{l|ccc|ccc|ccc|ccc|c}
    \toprule
        \multirow{2}{*}{\backslashbox[55mm]{\textbf{Model}}{\textbf{Eval. Mode}}} & \multicolumn{6}{c|}{\textbf{Classification}} & \multicolumn{6}{c|}{\textbf{Inference}} & \multirow{2}{*}{\textbf{Avg.}} \\
        \cmidrule(lr){2-4} \cmidrule(lr){5-7} \cmidrule(lr){8-10} \cmidrule(lr){11-13}
        & $\mathrm{Acc}_{-3}^{\textsc{cls}}$ & $\mathrm{Acc}_{0}^{\textsc{cls}}$ & $\mathrm{Acc}_{+3}^{\textsc{cls}}$ 
        & $\mathrm{F1}_{-3}^{\textsc{cls}}$ & $\mathrm{F1}_{0}^{\textsc{cls}}$ & $\mathrm{F1}_{+3}^{\textsc{cls}}$
        & $\mathrm{Acc}_{-3}^{\textsc{inf}}$ & $\mathrm{Acc}_{0}^{\textsc{inf}}$ & $\mathrm{Acc}_{+3}^{\textsc{inf}}$
        & $\mathrm{F1}_{-3}^{\textsc{inf}}$ & $\mathrm{F1}_{0}^{\textsc{inf}}$ & $\mathrm{F1}_{+3}^{\textsc{inf}}$ \\
        \midrule
        Qwen3-235B-A22B-Instruct-2507 & 52.25 & \underline{79.76} & \textbf{82.08} & 48.63 & \textbf{79.84} & \underline{82.02} & 66.67 & \textbf{79.53} & \textbf{81.99} & 54.47 & \underline{79.64} & \underline{81.94} & \textbf{72.40} \\
        Qwen3-30B-A3B-Instruct-2507 & \underline{55.24} & 75.05 & 77.00 & \underline{52.39} & 74.87 & 76.61 & 64.55 & 74.64 & 76.78 & 55.37 & 74.66 & 76.40 & 69.46\\
        Qwen3-32B & 45.20 & 70.54 & 71.86 & 39.10 & 68.99 & 69.93 & \textbf{82.81} & 71.47 & 71.86 & \underline{60.94} & 69.70 & 69.91 & 66.03 \\
        Qwen3-8B & 50.84 & 68.88 & 69.83 & 49.12 & 67.79 & 68.44 & 60.29 & 68.67 & 69.65 & 53.76 & 67.82 & 68.30 & 63.62 \\
        Qwen3-4B-Instruct-2507 & 45.58 & 62.20 & 64.80 & 38.62 & 61.84 & 64.33 & 37.75 & 61.57 & 64.66 & 38.10 & 61.81 & 64.27 & 55.46 \\
        \midrule
        Llama-3.3-70B-Instruct & 44.90 & 69.01 & 69.76 & 38.96 & 63.71 & 64.17 & 23.70 & 66.85 & 69.44 & 33.50 & 62.32 & 63.93 & 55.85 \\
        Llama-3.1-70B$^{\dagger}$ & 34.77 & 33.45 & 33.47 & 19.52 & 16.75 & 16.80 & 7.90 & 31.99 & 33.29 & 5.10 & 16.20 & 16.73 & 22.16 \\
        Llama-3.1-8B-Instruct & 40.69 & 50.05 & 50.20 & 36.53 & 45.67 & 45.11 & 78.69 & 50.16 & 50.09 & 60.26 & 46.89 & 45.14 & 49.96 \\
        Llama-3.2-3B$^{\dagger}$ & 33.42 & 33.42 & 33.42 & 16.70 & 16.70 & 16.70 & 7.88 & 31.97 & 33.24 & 4.87 & 16.15 & 16.63 & 22.76 \\
        Llama-3.2-1B$^{\dagger}$ & 33.38 & 33.31 & 33.31 & 16.74 & 16.72 & 16.69 & 7.90 & 31.85 & 33.13 & 4.89 & 16.16 & 16.62 & 21.72 \\
        \midrule
        DeepSeek-V3.1 & 43.67 & 64.44 & 63.73 & 38.73 & 66.39 & 66.31 & 78.03 & 64.73 & 63.83 & 55.01 & 66.48 & 66.38 & 61.48 \\
        \midrule
        Kimi-K2-Thinking$^\star$ & 45.97 & 57.15 & 54.21 & 45.34 & 61.71 & 59.22 & 58.13 & 56.56 & 54.08 & 42.07 & 61.57 & 59.15 & 54.60 \\
        \midrule
        gpt-oss-120b$^\star$ & \textbf{55.83} & \textbf{79.94} & \underline{81.81} & \textbf{52.82} & \underline{79.79} & 81.49 & 62.23 & \underline{78.89} & \underline{81.60} & 50.83 & 78.97 & 81.31 & \underline{72.13} \\
        gpt-oss-20b$^\star$ & 47.84 & 65.55 & 67.42 & 47.16 & 61.33 & 62.32 & 41.46 & 64.32 & 67.28 & 36.53 & 61.07 & 62.34 & 57.05 \\
        \midrule
        gpt-5.2-2025-12-11 & 47.63 & 76.34 & 77.32 & 42.88 & 78.20 & 79.56 & \underline{80.71} & 76.25 & 77.30 & \textbf{64.15} & 78.30 & 79.57 & 71.52 \\
        gpt-5.1-2025-11-13 & 51.89 & 78.60 & 81.01 & 47.66 & 79.76 & \textbf{82.17} & 68.44 & 78.51 & 80.83 & 53.68 & \textbf{79.72} & \textbf{81.99} & 72.02 \\
        gpt-4o-2024-08-06 & 48.84 & 72.29 & 73.98 & 44.45 & 71.38 & 72.83 & 74.61 & 72.95 & 74.13 & 54.99 & 71.94 & 72.98 & 67.11 \\
        gpt-4o-mini-2024-07-18 & 51.94 & 71.02 & 71.22 & 49.06 & 69.54 & 69.43 & 73.86 & 71.17 & 71.22 & 55.92 & 69.67 & 69.45 & 66.12 \\
    \bottomrule
    \end{tabular}
    }
    \caption{Performance comparison of 18 LLMs on LiveFact (November 2025). Models marked with $^\dagger$ are base models without instruction tuning (detailed in Table \ref{tab:llm_overview}). The best result in each column is \textbf{bolded}, and the second-best result is \underline{underlined}. The last column, Avg., represents the average of all results tested for this model. $E^{-3}$ CLS evaluates models ($\mathrm{Acc}_{-3}^{\textsc{cls}}$, $\mathrm{F1}_{-3}^{\textsc{cls}}$) under forced-choice conditions prior to event occurrence, highlighting prior bias rather than evidence-based verification. The asterisk $^\star$ denotes the non-standard evaluation setting required to accommodate verbose Chain-of-Thought (CoT) outputs.}
    \label{tab:result}
\end{table*}

\section{Experiments}
\label{sec:exp}

\subsection{Experiment Setup}

For our main experimental evaluation, we selected 18 representative models to ensure broad coverage of current LLM capabilities. This selection includes 14 open-source models (spanning the Qwen3, Llama 3, DeepSeek, Kimi, and GPT-OSS families) and 4 closed-source proprietary models (from the OpenAI GPT series). This diverse cohort ranges in scale from 1 billion to 1 trillion parameters, covering everything from lightweight edge-deployable models to flagship commercial systems. The specific architectural details, parameter counts, and training methodologies for all 18 models are provided in Appendix \ref{sec:models}. All models were evaluated on the LiveFact November 2025 dataset. Detailed evaluation settings, including inference hyperparameters and prompt structures, are outlined in Appendix \ref{sec:evaluation}.

\subsection{General Results \& Analysis}

Table \ref{tab:result} presents the comprehensive performance of all 18 models across both Classification and Inference modes. The experimental results demonstrate a significant shift in the competitive landscape of LLMs for complex reasoning tasks. The Qwen3-235B-A22B-Instruct model not only leads the open-source sector but also outperforms proprietary flagship models, including gpt-5.1, achieving the highest average score of 72.40\%. This dominance of Mixture-of-Experts (MoE) architectures—seen also in the strong performance of DeepSeek-V3.1 (61.48\%) despite its massive parameter count—suggests that sparse, high-capacity models are particularly well-suited for the multi-faceted nature of fake news detection, where knowledge retrieval must be dynamically routed. In contrast, traditional dense models like Llama-3.3-70B lag behind their MoE counterparts, highlighting potential limitations in scaling dense architectures for nuanced reasoning tasks.

It is important to clarify the interpretation of the Classification mode accuracy at the early $E^{(-3)}$ slice. Because sufficient evidence to definitively verify a claim rarely exists before an event occurs, forcing a model to make a binary (Real/Fake) prediction at this stage acts as a "Hallucination Stress Test."A high CLS score at $E^{(-3)}$ does \textit{not} necessarily indicate superior reasoning. Instead, it often implies that the model relies heavily on its parametric memory, prior biases, or "clairvoyant guessing" based on the linguistic plausibility of the claim. Therefore, we include the $E^{(-3)}$ CLS scores in the overall average to provide a holistic measure of a model's forced-choice behavior across all constraints, but it should not be viewed as a standalone metric of fact-checking capability. The true measure of evidence-based reasoning is whether a model correctly transitions to the "Ambiguous" label in the Inference mode when faced with this same $E^{(-3)}$ slice.

\subsubsection{Instruction Tuning and Reliability}
\label{sec:instruction_tuning}

The results highlight a critical performance distinction between standard instruction-tuned models and specialized "reasoning" models ($^\star$). Base models ($^\dagger$) consistently fail to navigate the complex formatting and reasoning constraints of the benchmark due to structural non-compliance. A more nuanced finding, however, relates to models trained with extensive CoT reinforcement, such as Kimi-K2 and the GPT-OSS series. Initially, these models appeared to fail under standard evaluation settings (128 tokens), yielding near-zero accuracy due to truncated outputs. However, when re-evaluated with an extended limit of 1024 tokens, their performance rebounded dramatically, with gpt-oss-120b achieving a competitive 72.13\% average. This suggests that for these architectures, "thinking" is not an optional feature but a structural necessity; they cannot be forced into immediate compliance without sacrificing their reasoning capabilities. We explore the inherent trade-offs between this "Thinking Mode" and standard instruction following in greater depth in Appendix \ref{sec:case}.

\begin{figure}
    \centering
    \includegraphics[width=0.98\linewidth]{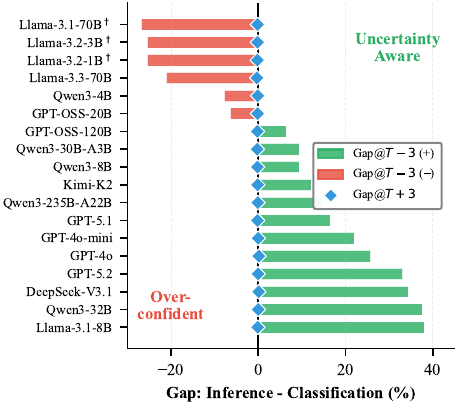}
    \caption{Analysis of the "Reasoning Gap" (Inference Score $-$ Classification Score at $\delta=-3$). Models with large positive gaps (green bars) effectively detect information voids ("Uncertainty Aware"). Models with negative or near-zero gaps (red bars) exhibit two failure modes: instruction-tuned models are genuinely "Overconfident," while base models ($\dagger$) fail due to format non-compliance rather than reasoning deficits.}
    \label{fig:model_gap}
\end{figure}

\subsubsection{Validating the LiveFact Paradigm: The "Reasoning Gap"}

A core contribution of LiveFact is its ability to differentiate between simple hallucination and genuine evidential reasoning. This differentiation is visualized in Figure \ref{fig:temporal_evolution}, which tracks performance across the $\delta=-3 \rightarrow \delta=0 \rightarrow \delta=+3$ timeline. In Classification Mode, models suffer a severe performance drop at $\delta=-3$ because they are forced to predict definitive labels like "Real" even when evidence is absent. However, in Inference Mode, where the ground truth is adjusted to "Ambiguous" to reflect the lack of evidence, we see a massive recovery in scores for capable models.

This phenomenon is further quantified in Figure~\ref{fig:model_gap}, which plots the "Reasoning Gap" (Inference Accuracy minus Classification Accuracy at $\delta = -3$). Models with a large positive gap (e.g., Llama-3.1-8B-Instruct with +38\% and Qwen3-32B with +37\%) can be characterized as \textit{Uncertainty Aware}. They correctly identify that the evidence is insufficient and predict "Ambiguous," aligning with the Inference ground truth. 

In contrast, models with negative or near-zero gaps fall into two distinct failure modes. \textit{Instruction-tuned models} such as Llama-3.3-70B-Instruct and Qwen3-4B exhibit genuine \textit{Overconfidence}: they produce valid classification outputs but continue to hallucinate "Real" or "Fake" verdicts regardless of the evidence void. \textit{Base models} (marked with $^\dagger$ in Table~\ref{tab:result}), however, represent a fundamentally different failure mode. As discussed in Section~\ref{sec:instruction_tuning}, models like Llama-3.1-70B lack instruction tuning and fail primarily due to \textit{structural non-compliance}---they cannot adhere to the required output format (\texttt{[[LABEL]]}), resulting in near-random parsed predictions rather than reasoned overconfidence. This distinction is critical: while overconfident models require improved calibration, base models require alignment training before their reasoning capabilities can be meaningfully assessed. A detailed confusion matrix analysis in Appendix~\ref{sec:confusion} further corroborates these findings, revealing that high-performing models exhibit clear "diagonalization" at $\delta=0$ and $\delta=+3$, while base models show pathological single-class collapse.

\subsubsection{Efficiency vs. Capability Analysis}

We further analyze the relationship between model scale, cost, and performance using Figure \ref{fig:main}. The figure plots the "Efficiency Frontier," revealing a non-linear relationship between resource consumption and utility. While massive models like Qwen3-235B-A22B and gpt-5.2 define the state-of-the-art (SOTA), they incur high operational costs (\$3.65 - \$9.27 per run). A compelling "efficiency sweet spot" emerges with mid-sized MoE models. As highlighted in the figure, Qwen3-30B-Instruct delivers 69.46\% average performance—within 3 percentage points of the leader—at approximately 14x cheaper cost than gpt-5.2. This suggests that for high-volume, real-time fake news detection applications, well-tuned mid-sized models offer a superior ROI compared to giant foundation models.

\section{Conclusion}

In this paper, we introduced LiveFact, a dynamic, time-aware benchmark designed to address the critical limitations of static fake news detection evaluations in the era of LLMs. By shifting the paradigm from simple knowledge retrieval to temporal reasoning under the "Fog of War," LiveFact provides a rigorous testbed for assessing whether models are genuinely reasoning or merely recalling pre-trained data. Our extensive experiments with 18 diverse LLMs reveal that while SOTA models exhibit strong reasoning capabilities, significant gaps remain in "epistemic humility" and instruction adherence, particularly among base models. Furthermore, we established that mid-sized MoE architectures offer a compelling balance of efficiency and performance. Finally, while the full utility of SSA will be realized through LiveFact's monthly updates to monitor long-term BDC risk, we have validated its mechanism through a simulated BDC injection experiment detailed in Appendix \ref{sec:ssa}. To facilitate community engagement and track ongoing progress in LLM verification capabilities, we maintain an interactive, regularly updated leaderboard\footnote{\url{https://livefact.bebxy.com/}}. We hope LiveFact serves as a sustainable gold standard, pushing the field toward more robust, transparent, and temporally grounded AI verification systems.

\section*{Limitations}

While LiveFact represents a significant advancement in dynamic, time-aware evaluation, we identify three primary limitations in its current iteration. First, the current benchmark is predominantly constructed using English-language sources from global news outlets. Consequently, it may not fully capture the cultural nuances, localized disinformation tactics, or linguistic complexities present in non-English regions, where misinformation often proliferates rapidly \citep{shibu2025scarcity}. Future work will aim to expand LiveFact into a multilingual framework, incorporating diverse regional sources to test cross-cultural reasoning capabilities. Second, our current framework focuses exclusively on textual claims and evidence. However, modern misinformation increasingly relies on multimodal content, including manipulated images, deepfakes, and decontextualized videos \citep{jing2023multimodal,khattar2019mvae,hu2025mage}. While text-based reasoning is foundational, it represents only one facet of the problem. We plan to extend future iterations of LiveFact to include multimodal evidence retrieval and verification tasks, aligning with the broader trajectory of LVLM (Large Vision-Language Model) development. Third, although our human-in-the-loop pipeline ensures high-quality ground truth (review synthetic data and ensure semantic consistency of entity shift data), it introduces a bottleneck regarding scalability. The requirement for expert review to distinguish between "Ambiguous" and "False" labels in Inference Mode limits the volume of claims we can release in each monthly cycle compared to fully synthetic datasets. To address this, we plan to investigate semi-automated verification protocols where reliable "Judge" models—calibrated against our human-verified corpus—can assist in the review process, thereby increasing dataset throughput without compromising rigor.

\section*{Ethical Considerations}

We strictly adhere to ethical guidelines regarding data privacy, copyright compliance, and model usage throughout the development of LiveFact. The application of AI assistants is limited solely to enhancing paper writing. All LLMs utilized in this study were accessed in strict accordance with their respective usage policies. For open-source models (e.g., Llama 3, Qwen3, DeepSeek), we utilized official checkpoints released under their specific community licenses. For proprietary models (e.g., OpenAI GPT series), access was obtained through authorized API endpoints, complying with all terms of service regarding automated evaluation and benchmarking. Our evidence retrieval pipeline relies on the Google API. To rigorously respect copyright regulations and intellectual property rights, we do not scrape, store, or redistribute the full text of news articles. As detailed in Appendix \ref{sec:evidence}, our dataset retains only minimal metadata—specifically webpage titles, source names, and publication timestamps—which falls within standard fair use for research purposes, also the data does not contain any privacy-sensitive data beyond public information. This metadata is sufficient for models to perform retrieval-augmented reasoning without infringing on publisher content. All data and artifacts produced in LiveFact are intended solely for academic research to enhance the safety and reliability of AI systems. We acknowledge the dual-use nature of fake news research \citep{sun2024exploring,hu2025llm}; however, our focus remains strictly on detection and verification. In all future long-term updates of the LiveFact benchmark, we commit to maintaining this rigorous standard of compliance, ensuring that all data sourcing and model interactions continue to adhere to evolving legal frameworks and ethical agreements.

\section*{Acknowledgments}

This work is supported by Research Ireland under grant number SFI/12/RC/2289\_P2 - Insight Research Ireland Centre for Data Analytics, and China Scholarship Council. We also acknowledge the support from Thinking Machines Lab and OpenAI Inc. for this work.

\bibliography{custom}

@inproceedings{shafiei2025multihoax,
    title = "{M}ulti{H}oax: A Dataset of Multi-hop False-premise questions",
    author = "Shafiei, Mohammadamin  and
      Saffari, Hamidreza  and
      Moosavi, Nafise Sadat",
    editor = "Che, Wanxiang  and
      Nabende, Joyce  and
      Shutova, Ekaterina  and
      Pilehvar, Mohammad Taher",
    booktitle = "Findings of the Association for Computational Linguistics: ACL 2025",
    month = jul,
    year = "2025",
    address = "Vienna, Austria",
    publisher = "Association for Computational Linguistics",
    url = "https://aclanthology.org/2025.findings-acl.530/",
    doi = "10.18653/v1/2025.findings-acl.530",
    pages = "10169--10187",
    ISBN = "979-8-89176-256-5"
}

@inproceedings{wang2025piecing,
    title = "Piecing It All Together: Verifying Multi-Hop Multimodal Claims",
    author = "Wang, Haoran  and
      Rangapur, Aman  and
      Xu, Xiongxiao  and
      Liang, Yueqing  and
      Gharwi, Haroon  and
      Yang, Carl  and
      Shu, Kai",
    editor = "Rambow, Owen  and
      Wanner, Leo  and
      Apidianaki, Marianna  and
      Al-Khalifa, Hend  and
      Eugenio, Barbara Di  and
      Schockaert, Steven",
    booktitle = "Proceedings of the 31st International Conference on Computational Linguistics",
    month = jan,
    year = "2025",
    address = "Abu Dhabi, UAE",
    publisher = "Association for Computational Linguistics",
    url = "https://aclanthology.org/2025.coling-main.498/",
    pages = "7453--7469"
}

@INPROCEEDINGS{huang2025multi,
  author={Huang, Zhongqiang and Lu, Dongli and Sha, Ying},
  booktitle={ICASSP 2025 - 2025 IEEE International Conference on Acoustics, Speech and Signal Processing (ICASSP)}, 
  title={Multi-Hop Attention Diffusion Graph Neural Networks For Multimodal Fake News Detection}, 
  year={2025},
  volume={},
  number={},
  pages={1-5},
  doi={10.1109/ICASSP49660.2025.10887755}}

@inproceedings{ostrowski2021multi,
  title     = {Multi-Hop Fact Checking of Political Claims},
  author    = {Ostrowski, Wojciech and Arora, Arnav and Atanasova, Pepa and Augenstein, Isabelle},
  booktitle = {Proceedings of the Thirtieth International Joint Conference on
               Artificial Intelligence, {IJCAI-21}},
  publisher = {International Joint Conferences on Artificial Intelligence Organization},
  editor    = {Zhi-Hua Zhou},
  pages     = {3892--3898},
  year      = {2021},
  month     = {8},
  note      = {Main Track},
  doi       = {10.24963/ijcai.2021/536},
  url       = {https://doi.org/10.24963/ijcai.2021/536},
}

@inproceedings{xiao2024msynfd,
author = {Xiao, Liang and Zhang, Qi and Shi, Chongyang and Wang, Shoujin and Naseem, Usman and Hu, Liang},
title = {MSynFD: Multi-hop Syntax Aware Fake News Detection},
year = {2024},
isbn = {9798400701719},
publisher = {Association for Computing Machinery},
address = {New York, NY, USA},
url = {https://doi.org/10.1145/3589334.3645468},
doi = {10.1145/3589334.3645468},
booktitle = {Proceedings of the ACM Web Conference 2024},
pages = {4128–4137},
numpages = {10},
location = {Singapore, Singapore},
series = {WWW '24}
}

@InProceedings{mishra2020fake,
author = {Mishra, Rahul},
title = {Fake News Detection Using Higher-Order User to User Mutual-Attention Progression in Propagation Paths},
booktitle = {Proceedings of the IEEE/CVF Conference on Computer Vision and Pattern Recognition (CVPR) Workshops},
month = {June},
year = {2020},
url={http://openaccess.thecvf.com/content_CVPRW_2020/html/w39/Mishra_Fake_News_Detection_Using_Higher-Order_User_to_User_Mutual-Attention_Progression_CVPRW_2020_paper.html}
}

@Article{chen2023rumor,
AUTHOR = {Chen, Jianhong and Zhang, Wenyi and Ma, Hongcai and Yang, Shan},
TITLE = {Rumor Detection in Social Media Based on Multi-Hop Graphs and Differential Time Series},
JOURNAL = {Mathematics},
VOLUME = {11},
YEAR = {2023},
NUMBER = {16},
ARTICLE-NUMBER = {3461},
URL = {https://www.mdpi.com/2227-7390/11/16/3461},
ISSN = {2227-7390},
DOI = {10.3390/math11163461}
}

@inproceedings{liao2023muser,
author = {Liao, Hao and Peng, Jiahao and Huang, Zhanyi and Zhang, Wei and Li, Guanghua and Shu, Kai and Xie, Xing},
title = {MUSER: A MUlti-Step Evidence Retrieval Enhancement Framework for Fake News Detection},
year = {2023},
isbn = {9798400701030},
publisher = {Association for Computing Machinery},
address = {New York, NY, USA},
url = {https://doi.org/10.1145/3580305.3599873},
doi = {10.1145/3580305.3599873},
booktitle = {Proceedings of the 29th ACM SIGKDD Conference on Knowledge Discovery and Data Mining},
pages = {4461–4472},
numpages = {12},
location = {Long Beach, CA, USA},
series = {KDD '23}
}

@article{zhang2024causal, 
title={Causal Walk: Debiasing Multi-Hop Fact Verification with Front-Door Adjustment}, 
volume={38}, 
url={https://ojs.aaai.org/index.php/AAAI/article/view/29925}, 
DOI={10.1609/aaai.v38i17.29925}, 
abstractNote={Multi-hop fact verification aims to detect the veracity of the given claim by integrating and reasoning over multiple pieces of evidence. Conventional multi-hop fact verification models are prone to rely on spurious correlations from the annotation artifacts, leading to an obvious performance decline on unbiased datasets. Among the various debiasing works, the causal inference-based methods become popular by performing theoretically guaranteed debiasing such as casual intervention or counterfactual reasoning. However, existing causal inference-based debiasing methods, which mainly formulate fact verification as a single-hop reasoning task to tackle shallow bias patterns, cannot deal with the complicated bias patterns hidden in multiple hops of evidence. To address the challenge, we propose Causal Walk, a novel method for debiasing multi-hop fact verification from a causal perspective with front-door adjustment. Specifically, in the structural causal model, the reasoning path between the treatment (the input claim-evidence graph) and the outcome (the veracity label) is introduced as the mediator to block the confounder. With the front-door adjustment, the causal effect between the treatment and the outcome is decomposed into the causal effect between the treatment and the mediator, which is estimated by applying the idea of random walk, and the causal effect between the mediator and the outcome, which is estimated with normalized weighted geometric mean approximation. To investigate the effectiveness of the proposed method, an adversarial multi-hop fact verification dataset and a symmetric multi-hop fact verification dataset are proposed with the help of the large language model. Experimental results show that Causal Walk outperforms some previous debiasing methods on both existing datasets and the newly constructed datasets. Code and data will be released at https://github.com/zcccccz/CausalWalk.}, 
number={17}, 
journal={Proceedings of the AAAI Conference on Artificial Intelligence}, 
author={Zhang, Congzhi and Zhang, Linhai and Zhou, Deyu}, 
year={2024}, 
month={Mar.}, 
pages={19533-19541} 
}

@inproceedings{ma2024ex,
    title = "{EX}-{FEVER}: A Dataset for Multi-hop Explainable Fact Verification",
    author = "Ma, Huanhuan  and
      Xu, Weizhi  and
      Wei, Yifan  and
      Chen, Liuji  and
      Wang, Liang  and
      Liu, Qiang  and
      Wu, Shu  and
      Wang, Liang",
    editor = "Ku, Lun-Wei  and
      Martins, Andre  and
      Srikumar, Vivek",
    booktitle = "Findings of the Association for Computational Linguistics: ACL 2024",
    month = aug,
    year = "2024",
    address = "Bangkok, Thailand",
    publisher = "Association for Computational Linguistics",
    url = "https://aclanthology.org/2024.findings-acl.556/",
    doi = "10.18653/v1/2024.findings-acl.556",
    pages = "9340--9353"
}

@article{sadeghi2022fake,
  title={Fake news detection on social media using a natural language inference approach},
  author={Sadeghi, Fariba and Bidgoly, Amir Jalaly and Amirkhani, Hossein},
  journal={Multimedia Tools and Applications},
  volume={81},
  number={23},
  pages={33801--33821},
  year={2022},
  publisher={Springer},
  doi={10.1007/s11042-022-12428-8}
}

@article{cao2025enhancing, 
title={Enhancing Multi-Hop Fact Verification with Structured Knowledge-Augmented Large Language Models}, 
volume={39}, 
url={https://ojs.aaai.org/index.php/AAAI/article/view/34520}, 
DOI={10.1609/aaai.v39i22.34520}, 
abstractNote={The rapid development of social platforms exacerbates the dissemination of misinformation, which stimulates the research in fact verification. Recent studies tend to leverage semantic features to solve this problem as a single-hop task. However, the process of verifying a claim requires several pieces of evidence with complicated inner logic and relations to verify the given claim in real-world situations. Recent studies attempt to improve both understanding and reasoning abilities to enhance the performance, but they overlook the crucial relations between entities that benefit models to understand better and facilitate the prediction. To emphasize the significance of relations, we resort to Large Language Models (LLMs) considering their excellent understanding ability. Instead of other methods using LLMs as the predictor, we take them as relation extractors, for they do better in understanding rather than reasoning according to the experimental results. Thus, to solve the challenges above, we propose a novel Structured Knowledge-Augmented LLM-based Network (LLM-SKAN) for multi-hop fact verification. Specifically, we utilize an LLM-driven Knowledge Extractor to capture fine-grained information, including entities and their complicated relations. Besides, we leverage a Knowledge-Augmented Relation Graph Fusion module to interact with each node and learn better claim-evidence representations comprehensively. The experimental results on four common-used datasets demonstrate the effectiveness and superiority of our model.}, 
number={22}, 
journal={Proceedings of the AAAI Conference on Artificial Intelligence}, 
author={Cao, Han and Wei, Lingwei and Zhou, Wei and Hu, Songlin}, 
year={2025}, 
month={Apr.}, 
pages={23514-23522} 
}

@article{zheng2025mrr, 
title={MRR-FV: Unlocking Complex Fact Verification with Multi-Hop Retrieval and Reasoning}, 
volume={39}, 
url={https://ojs.aaai.org/index.php/AAAI/article/view/34802}, 
DOI={10.1609/aaai.v39i24.34802}, 
abstractNote={The pervasive spread of misinformation on social networks highlights the critical necessity for effective fact verification systems. Traditional approaches primarily focus on pairwise correlations between claims and evidence, often neglecting comprehensive multi-hop retrieval and reasoning, which results in suboptimal performance when dealing with complex claims. In this paper, we propose MRR-FV, a generative retrieval-enhanced model designed to address the novel challenge of Multi-hop Retrieval and Reasoning for Fact Verification, which integrates two core modules: Generative Multi-hop Retriever and the Hierarchical Interaction Reasoner. MRR-FV utilizes an autoregressive model for iterative multi-hop evidence retrieval, complemented by a pre-trained compressor to address the challenge of intention shift across retrieval hops. For claim verification, we propose a hierarchical interaction reasoner that conducts intra-sentence reasoning to capture long-term semantic dependencies and inter-sentence reasoning across multi-hop evidence subgraphs to reveal complex evidence interactions. Experimental evaluations on the FEVER and HOVER datasets demonstrate the superior performance of our model in both claim verification and evidence retrieval tasks.}, 
number={24}, 
journal={Proceedings of the AAAI Conference on Artificial Intelligence}, 
author={Zheng, Liwen and Li, Chaozhuo and Zhang, Litian and Jia, Haoran and Wang, Senzhang and Liu, Zheng and Zhang, Xi}, 
year={2025}, 
month={Apr.}, 
pages={26066-26074} 
}

@article{si2023exploring, 
title={Exploring Faithful Rationale for Multi-Hop Fact Verification via Salience-Aware Graph Learning}, 
volume={37}, 
url={https://ojs.aaai.org/index.php/AAAI/article/view/26591}, 
DOI={10.1609/aaai.v37i11.26591}, 
abstractNote={The opaqueness of the multi-hop fact verification model imposes imperative requirements for explainability. One feasible way is to extract rationales, a subset of inputs, where the performance of prediction drops dramatically when being removed. Though being explainable, most rationale extraction methods for multi-hop fact verification explore the semantic information within each piece of evidence individually, while ignoring the topological information interaction among different pieces of evidence. Intuitively, a faithful rationale bears complementary information being able to extract other rationales through the multi-hop reasoning process. To tackle such disadvantages, we cast explainable multi-hop fact verification as subgraph extraction, which can be solved based on graph convolutional network (GCN) with salience-aware graph learning. In specific, GCN is utilized to incorporate the topological interaction information among multiple pieces of evidence for learning evidence representation. Meanwhile, to alleviate the influence of noisy evidence, the salience-aware graph perturbation is induced into the message passing of GCN. Moreover, the multi-task model with three diagnostic properties of rationale is elaborately designed to improve the quality of an explanation without any explicit annotations. Experimental results on the FEVEROUS benchmark show significant gains over previous state-of-the-art methods for both rationale extraction and fact verification.}, 
number={11}, 
journal={Proceedings of the AAAI Conference on Artificial Intelligence}, 
author={Si, Jiasheng and Zhu, Yingjie and Zhou, Deyu}, 
year={2023}, 
month={Jun.}, 
pages={13573-13581} 
}

@inproceedings{jiang2020hover,
    title = "{H}o{V}er: A Dataset for Many-Hop Fact Extraction And Claim Verification",
    author = "Jiang, Yichen  and
      Bordia, Shikha  and
      Zhong, Zheng  and
      Dognin, Charles  and
      Singh, Maneesh  and
      Bansal, Mohit",
    editor = "Cohn, Trevor  and
      He, Yulan  and
      Liu, Yang",
    booktitle = "Findings of the Association for Computational Linguistics: EMNLP 2020",
    month = nov,
    year = "2020",
    address = "Online",
    publisher = "Association for Computational Linguistics",
    url = "https://aclanthology.org/2020.findings-emnlp.309/",
    doi = "10.18653/v1/2020.findings-emnlp.309",
    pages = "3441--3460"
}

@article{zhou2020survey,
author = {Zhou, Xinyi and Zafarani, Reza},
title = {A Survey of Fake News: Fundamental Theories, Detection Methods, and Opportunities},
year = {2020},
issue_date = {September 2021},
publisher = {Association for Computing Machinery},
address = {New York, NY, USA},
volume = {53},
number = {5},
issn = {0360-0300},
url = {https://doi.org/10.1145/3395046},
doi = {10.1145/3395046},
journal = {ACM Comput. Surv.},
month = sep,
articleno = {109},
numpages = {40}
}

@article{guo2022survey,
    author = {Guo, Zhijiang and Schlichtkrull, Michael and Vlachos, Andreas},
    title = {A Survey on Automated Fact-Checking},
    journal = {Transactions of the Association for Computational Linguistics},
    volume = {10},
    pages = {178-206},
    year = {2022},
    month = {02},
    issn = {2307-387X},
    doi = {10.1162/tacl_a_00454},
    url = {https://doi.org/10.1162/tacl_a_00454},
    eprint = {https://direct.mit.edu/tacl/article-pdf/doi/10.1162/tacl_a_00454/1987018/tacl_a_00454.pdf},
}

@inproceedings{wang2017liar,
    title = "``Liar, Liar Pants on Fire'': A New Benchmark Dataset for Fake News Detection",
    author = "Wang, William Yang",
    editor = "Barzilay, Regina  and
      Kan, Min-Yen",
    booktitle = "Proceedings of the 55th Annual Meeting of the Association for Computational Linguistics (Volume 2: Short Papers)",
    month = jul,
    year = "2017",
    address = "Vancouver, Canada",
    publisher = "Association for Computational Linguistics",
    url = "https://aclanthology.org/P17-2067/",
    doi = "10.18653/v1/P17-2067",
    pages = "422--426"
}

@article{rocha2021impact,
  title={The impact of fake news on social media and its influence on health during the COVID-19 pandemic: A systematic review},
  author={Rocha, Y.M. and de Moura, G.A. and Desid{\'e}rio, G.A. and de Oliveira, C.H. and Louren{\c{c}}o, F.D. and de Figueiredo Nicolete, L.D.},
  journal={Journal of Public Health},
  pages={1--10},
  year={2021},
  doi={10.1007/s10389-021-01658-z},
  url={https://doi.org/10.1007/s10389-021-01658-z},
  publisher={Springer}
}

@article{xu2024enhanced,
  author={Xu, Cheng and Kechadi, M-Tahar},
  journal={IEEE Access}, 
  title={An Enhanced Fake News Detection System With Fuzzy Deep Learning}, 
  year={2024},
  volume={12},
  number={},
  pages={88006-88021},
  url={https://doi.org/10.1109/ACCESS.2024.3418340},
  doi={10.1109/ACCESS.2024.3418340}}

@inproceedings{xu2025dcr,
    title = "{DCR}: Quantifying Data Contamination in {LLM}s Evaluation",
    author = "Xu, Cheng  and
      Yan, Nan  and
      Guan, Shuhao  and
      Jin, Changhong  and
      Mei, Yuke  and
      Guo, Yibing  and
      Kechadi, Tahar",
    editor = "Christodoulopoulos, Christos  and
      Chakraborty, Tanmoy  and
      Rose, Carolyn  and
      Peng, Violet",
    booktitle = "Proceedings of the 2025 Conference on Empirical Methods in Natural Language Processing",
    month = nov,
    year = "2025",
    address = "Suzhou, China",
    publisher = "Association for Computational Linguistics",
    url = "https://aclanthology.org/2025.emnlp-main.1173/",
    doi = "10.18653/v1/2025.emnlp-main.1173",
    pages = "23013--23031",
    ISBN = "979-8-89176-332-6"
}

@inproceedings{xu2025ssa,
    title = "{SSA}: Semantic Contamination of {LLM}-Driven Fake News Detection",
    author = "Xu, Cheng  and
      Yan, Nan  and
      Guan, Shuhao  and
      Mei, Yuke  and
      Kechadi, Tahar",
    editor = "Christodoulopoulos, Christos  and
      Chakraborty, Tanmoy  and
      Rose, Carolyn  and
      Peng, Violet",
    booktitle = "Proceedings of the 2025 Conference on Empirical Methods in Natural Language Processing",
    month = nov,
    year = "2025",
    address = "Suzhou, China",
    publisher = "Association for Computational Linguistics",
    url = "https://aclanthology.org/2025.emnlp-main.744/",
    doi = "10.18653/v1/2025.emnlp-main.744",
    pages = "14748--14762",
    ISBN = "979-8-89176-332-6"
}

@inproceedings{xu2025triplefact,
    title = "{T}riple{F}act: Defending Data Contamination in the Evaluation of {LLM}-driven Fake News Detection",
    author = "Xu, Cheng  and
      Yan, Nan",
    editor = "Che, Wanxiang  and
      Nabende, Joyce  and
      Shutova, Ekaterina  and
      Pilehvar, Mohammad Taher",
    booktitle = "Proceedings of the 63rd Annual Meeting of the Association for Computational Linguistics (Volume 1: Long Papers)",
    month = jul,
    year = "2025",
    address = "Vienna, Austria",
    publisher = "Association for Computational Linguistics",
    url = "https://aclanthology.org/2025.acl-long.431/",
    doi = "10.18653/v1/2025.acl-long.431",
    pages = "8808--8823",
    ISBN = "979-8-89176-251-0"
}

@inproceedings{xu2023fuzzy,
   author = {Xu, Cheng and Kechadi, M-Tahar},
   title = {Fuzzy Deep Hybrid Network for Fake News Detection},
   year = {2023},
   isbn = {9798400708916},
   publisher = {Association for Computing Machinery},
   address = {New York, NY, USA},
   url = {https://doi.org/10.1145/3628797.3628971},
   doi = {10.1145/3628797.3628971},
   booktitle = {Proceedings of the 12th International Symposium on Information and Communication Technology},
   pages = {118–125},
   numpages = {8},
   location = {<conf-loc>, <city>Ho Chi Minh</city>, <country>Vietnam</country>, </conf-loc>},
   series = {SOICT '23}
}

@misc{xu2024benchmark,
      title={Benchmark Data Contamination of Large Language Models: A Survey}, 
      author={Cheng Xu and Shuhao Guan and Derek Greene and M-Tahar Kechadi},
      year={2024},
      eprint={2406.04244},
      archivePrefix={arXiv},
      primaryClass={cs.CL},
      url={https://arxiv.org/abs/2406.04244}, 
}

@inproceedings{sun2025the,
title={The Emperor's New Clothes in Benchmarking? A Rigorous Examination of Mitigation Strategies for {LLM} Benchmark Data Contamination},
author={Yifan Sun and Han Wang and Dongbai Li and Gang Wang and Huan Zhang},
booktitle={Forty-second International Conference on Machine Learning},
year={2025},
url={https://openreview.net/forum?id=TuvDxubEfE}
}

@article{walter2020fact,
author = {Nathan Walter and Jonathan Cohen and R. Lance Holbert and Yasmin Morag},
title = {Fact-Checking: A Meta-Analysis of What Works and for Whom},
journal = {Political Communication},
volume = {37},
number = {3},
pages = {350--375},
year = {2020},
publisher = {Routledge},
doi = {10.1080/10584609.2019.1668894},
URL = {https://doi.org/10.1080/10584609.2019.1668894}
}

@inproceedings{white2025livebench,
title={LiveBench: A Challenging, Contamination-Limited {LLM} Benchmark},
author={Colin White and Samuel Dooley and Manley Roberts and Arka Pal and Benjamin Feuer and Siddhartha Jain and Ravid Shwartz-Ziv and Neel Jain and Khalid Saifullah and Sreemanti Dey and Shubh-Agrawal and Sandeep Singh Sandha and Siddartha Venkat Naidu and Chinmay Hegde and Yann LeCun and Tom Goldstein and Willie Neiswanger and Micah Goldblum},
booktitle={The Thirteenth International Conference on Learning Representations},
year={2025},
url={https://openreview.net/forum?id=sKYHBTAxVa}
}

@INPROCEEDINGS{xu2025analysis,
  author={Xu, Cheng and Kechadi, M-Tahar},
  booktitle={2025 IEEE International Conference on Big Data (BigData)}, 
  title={Analysis of Semantic Benchmark Data Contamination Attack for LLM-Driven Fake News Detection}, 
  year={2025},
  volume={},
  number={},
  pages={3656-3664},
  doi={10.1109/BigData66926.2025.11402039}
}

@inproceedings{
zhang2024llm,
title={{LLM} as a Mastermind: A Survey of Strategic Reasoning with Large Language Models},
author={Yadong Zhang and Shaoguang Mao and Tao Ge and Xun Wang and Yan Xia and Wenshan Wu and Ting Song and Man Lan and Furu Wei},
booktitle={First Conference on Language Modeling},
year={2024},
url={https://openreview.net/forum?id=iMqJsQ4evS}
}

@inproceedings{
havrilla2024glore,
title={{GL}oRe: When, Where, and How to Improve {LLM} Reasoning via Global and Local Refinements},
author={Alexander Havrilla and Sharath Chandra Raparthy and Christoforos Nalmpantis and Jane Dwivedi-Yu and Maksym Zhuravinskyi and Eric Hambro and Roberta Raileanu},
booktitle={Forty-first International Conference on Machine Learning},
year={2024},
url={https://openreview.net/forum?id=LH6R06NxdB}
}

@article{Zhang2025is, title={Is LLMs Hallucination Usable? LLM-based Negative Reasoning for Fake News Detection}, volume={39}, url={https://ojs.aaai.org/index.php/AAAI/article/view/32089}, DOI={10.1609/aaai.v39i1.32089}, abstractNote={The questionable responses caused by knowledge hallucination may lead to LLMs’ unstable ability in decision-making. However, it has never been investigated whether the LLMs’ hallucination is possibly usable for generating negative reasoning to assist fake news detection. In this paper, we propose a novel supervised self-reinforced reasoning rectification approach - SR^3 that not only yields common reasonable reasoning for news but also forces LLMs to generate the wrong understandings of news via LLMs reflection for semantic consistency learning. Upon that, we construct a negative reasoning-based news learning model called - NRFE, which leverages positive or negative news-reasoning pairs for learning the semantic consistency between them. To avoid the impact of label-implicated reasoning, we deploy a student model - NRFE-D that only takes news content as input to inspect the performance of our method by distilling the knowledge from NRFE. The experimental results verified on three popular fake news datasets demonstrate the superiority of our method compared with three kinds of baselines including prompting-based LLMs, fine-tuning-based PLMs, and other representative fake news detection methods.}, number={1}, journal={Proceedings of the AAAI Conference on Artificial Intelligence}, author={Zhang, Chaowei and Feng, Zongling and Zhang, Zewei and Qiang, Jipeng and Xu, Guandong and Li, Yun}, year={2025}, month={Apr.}, pages={1031-1039} }

@InProceedings{xiao2026fact,
author="Xiao, Liang
and Shi, Chongyang
and Hao, Shufeng
and Wei, Zeyu",
editor="Taniguchi, Tadahiro
and Leung, Chi Sing Andrew
and Kozuno, Tadashi
and Yoshimoto, Junichiro
and Mahmud, Mufti
and Doborjeh, Maryam
and Doya, Kenji",
title="Fact-Augmented Reasoning Model for Fake News Detection",
booktitle="Neural Information Processing",
year="2025",
publisher="Springer Nature Singapore",
address="Singapore",
pages="18--32",
doi={10.1007/978-981-95-4109-6_2},
isbn="978-981-95-4109-6"
}

@inproceedings{ma2024fake,
    title = "On Fake News Detection with {LLM} Enhanced Semantics Mining",
    author = "Ma, Xiaoxiao  and
      Zhang, Yuchen  and
      Ding, Kaize  and
      Yang, Jian  and
      Wu, Jia  and
      Fan, Hao",
    editor = "Al-Onaizan, Yaser  and
      Bansal, Mohit  and
      Chen, Yun-Nung",
    booktitle = "Proceedings of the 2024 Conference on Empirical Methods in Natural Language Processing",
    month = nov,
    year = "2024",
    address = "Miami, Florida, USA",
    publisher = "Association for Computational Linguistics",
    url = "https://aclanthology.org/2024.emnlp-main.31/",
    doi = "10.18653/v1/2024.emnlp-main.31",
    pages = "508--521"
}

@inproceedings{atanasova2020generating,
    title = "Generating Fact Checking Explanations",
    author = "Atanasova, Pepa  and
      Simonsen, Jakob Grue  and
      Lioma, Christina  and
      Augenstein, Isabelle",
    editor = "Jurafsky, Dan  and
      Chai, Joyce  and
      Schluter, Natalie  and
      Tetreault, Joel",
    booktitle = "Proceedings of the 58th Annual Meeting of the Association for Computational Linguistics",
    month = jul,
    year = "2020",
    address = "Online",
    publisher = "Association for Computational Linguistics",
    url = "https://aclanthology.org/2020.acl-main.656/",
    doi = "10.18653/v1/2020.acl-main.656",
    pages = "7352--7364"
}

@inproceedings{alhindi2018evidence,
    title = "Where is Your Evidence: Improving Fact-checking by Justification Modeling",
    author = "Alhindi, Tariq  and
      Petridis, Savvas  and
      Muresan, Smaranda",
    editor = "Thorne, James  and
      Vlachos, Andreas  and
      Cocarascu, Oana  and
      Christodoulopoulos, Christos  and
      Mittal, Arpit",
    booktitle = "Proceedings of the First Workshop on Fact Extraction and {VER}ification ({FEVER})",
    month = nov,
    year = "2018",
    address = "Brussels, Belgium",
    publisher = "Association for Computational Linguistics",
    url = "https://aclanthology.org/W18-5513/",
    doi = "10.18653/v1/W18-5513",
    pages = "85--90"
}

@inproceedings{lee2022deduplicating,
    title = "Deduplicating Training Data Makes Language Models Better",
    author = "Lee, Katherine  and
      Ippolito, Daphne  and
      Nystrom, Andrew  and
      Zhang, Chiyuan  and
      Eck, Douglas  and
      Callison-Burch, Chris  and
      Carlini, Nicholas",
    editor = "Muresan, Smaranda  and
      Nakov, Preslav  and
      Villavicencio, Aline",
    booktitle = "Proceedings of the 60th Annual Meeting of the Association for Computational Linguistics (Volume 1: Long Papers)",
    month = may,
    year = "2022",
    address = "Dublin, Ireland",
    publisher = "Association for Computational Linguistics",
    url = "https://aclanthology.org/2022.acl-long.577",
    doi = "10.18653/v1/2022.acl-long.577",
    pages = "8424--8445",
}

@inproceedings{sainz2023nlp,
    title = "{NLP} Evaluation in trouble: On the Need to Measure {LLM} Data Contamination for each Benchmark",
    author = "Sainz, Oscar  and
      Campos, Jon  and
      Garc{\'\i}a-Ferrero, Iker  and
      Etxaniz, Julen  and
      de Lacalle, Oier Lopez  and
      Agirre, Eneko",
    editor = "Bouamor, Houda  and
      Pino, Juan  and
      Bali, Kalika",
    booktitle = "Findings of the Association for Computational Linguistics: EMNLP 2023",
    month = dec,
    year = "2023",
    address = "Singapore",
    publisher = "Association for Computational Linguistics",
    url = "https://aclanthology.org/2023.findings-emnlp.722",
    doi = "10.18653/v1/2023.findings-emnlp.722",
    pages = "10776--10787",
}

@misc{zhou2023dont,
      title={Don't Make Your LLM an Evaluation Benchmark Cheater}, 
      author={Kun Zhou and Yutao Zhu and Zhipeng Chen and Wentong Chen and Wayne Xin Zhao and Xu Chen and Yankai Lin and Ji-Rong Wen and Jiawei Han},
      year={2023},
      eprint={2311.01964},
      archivePrefix={arXiv},
      primaryClass={cs.CL}
}

@misc{mcintosh2024inadequacies,
      title={Inadequacies of Large Language Model Benchmarks in the Era of Generative Artificial Intelligence}, 
      author={Timothy R. McIntosh and Teo Susnjak and Tong Liu and Paul Watters and Malka N. Halgamuge},
      year={2024},
      eprint={2402.09880},
      archivePrefix={arXiv},
      primaryClass={cs.AI}
}

@inproceedings{deng2024unveiling,
    title = "Unveiling the Spectrum of Data Contamination in Language Model: A Survey from Detection to Remediation",
    author = "Deng, Chunyuan  and
      Zhao, Yilun  and
      Heng, Yuzhao  and
      Li, Yitong  and
      Cao, Jiannan  and
      Tang, Xiangru  and
      Cohan, Arman",
    editor = "Ku, Lun-Wei  and
      Martins, Andre  and
      Srikumar, Vivek",
    booktitle = "Findings of the Association for Computational Linguistics ACL 2024",
    month = aug,
    year = "2024",
    address = "Bangkok, Thailand and virtual meeting",
    publisher = "Association for Computational Linguistics",
    url = "https://aclanthology.org/2024.findings-acl.951",
    pages = "16078--16092",
}

@inproceedings{
liu2025mmfakebench,
title={{MMF}akeBench: A Mixed-Source Multimodal Misinformation Detection Benchmark for {LVLM}s},
author={Xuannan Liu and Zekun Li and Pei Pei Li and Huaibo Huang and Shuhan Xia and Xing Cui and Linzhi Huang and Weihong Deng and Zhaofeng He},
booktitle={The Thirteenth International Conference on Learning Representations},
year={2025},
url={https://openreview.net/forum?id=D6zn6ozJs7}
}

@article{shu2020fakenewsnet,
author = {Shu, Kai and Mahudeswaran, Deepak and Wang, Suhang and Lee, Dongwon and Liu, Huan},
title = {FakeNewsNet: A Data Repository with News Content, Social Context, and Spatiotemporal Information for Studying Fake News on Social Media},
journal = {Big Data},
volume = {8},
number = {3},
pages = {171-188},
year = {2020},
doi = {10.1089/big.2020.0062},
    note ={PMID: 32491943}
}

@InProceedings{meyers2020fake,
author="Meyers, Marion
and Weiss, Gerhard
and Spanakis, Gerasimos",
editor="van Duijn, Max
and Preuss, Mike
and Spaiser, Viktoria
and Takes, Frank
and Verberne, Suzan",
title="Fake News Detection on Twitter Using Propagation Structures",
booktitle="Disinformation in Open Online Media",
year="2020",
publisher="Springer International Publishing",
address="Cham",
pages="138--158",
doi={10.1007/978-3-030-61841-4_10},
isbn="978-3-030-61841-4"
}

@inproceedings{
li2025preference,
title={Preference Leakage: A Contamination Problem in {LLM}-as-a-judge},
author={Dawei Li and Renliang Sun and Yue Huang and Ming Zhong and Bohan Jiang and Jiawei Han and Xiangliang Zhang and Wei Wang and huan liu},
booktitle={Data in Generative Models - The Bad, the Ugly, and the Greats},
year={2025},
url={https://openreview.net/forum?id=gW6NT2IuME}
}

@misc{yang2025qwen3,
      title={Qwen3 Technical Report}, 
      author={An Yang and Anfeng Li and Baosong Yang and Beichen Zhang and Binyuan Hui and Bo Zheng and Bowen Yu and Chang Gao and Chengen Huang and Chenxu Lv and Chujie Zheng and Dayiheng Liu and Fan Zhou and Fei Huang and Feng Hu and Hao Ge and Haoran Wei and Huan Lin and Jialong Tang and Jian Yang and Jianhong Tu and Jianwei Zhang and Jianxin Yang and Jiaxi Yang and Jing Zhou and Jingren Zhou and Junyang Lin and Kai Dang and Keqin Bao and Kexin Yang and Le Yu and Lianghao Deng and Mei Li and Mingfeng Xue and Mingze Li and Pei Zhang and Peng Wang and Qin Zhu and Rui Men and Ruize Gao and Shixuan Liu and Shuang Luo and Tianhao Li and Tianyi Tang and Wenbiao Yin and Xingzhang Ren and Xinyu Wang and Xinyu Zhang and Xuancheng Ren and Yang Fan and Yang Su and Yichang Zhang and Yinger Zhang and Yu Wan and Yuqiong Liu and Zekun Wang and Zeyu Cui and Zhenru Zhang and Zhipeng Zhou and Zihan Qiu},
      year={2025},
      eprint={2505.09388},
      archivePrefix={arXiv},
      primaryClass={cs.CL},
      url={https://arxiv.org/abs/2505.09388}, 
}

@misc{grattafiori2024llama3,
      title={The Llama 3 Herd of Models}, 
      author={Aaron Grattafiori and Abhimanyu Dubey and Abhinav Jauhri and Abhinav Pandey and Abhishek Kadian and Ahmad Al-Dahle and Aiesha Letman and Akhil Mathur and Alan Schelten and Alex Vaughan and Amy Yang and Angela Fan and Anirudh Goyal and Anthony Hartshorn and Aobo Yang and Archi Mitra and Archie Sravankumar and Artem Korenev and Arthur Hinsvark and Arun Rao and Aston Zhang and Aurelien Rodriguez and Austen Gregerson and Ava Spataru and Baptiste Roziere and Bethany Biron and Binh Tang and Bobbie Chern and Charlotte Caucheteux and Chaya Nayak and Chloe Bi and Chris Marra and Chris McConnell and Christian Keller and Christophe Touret and Chunyang Wu and Corinne Wong and Cristian Canton Ferrer and Cyrus Nikolaidis and Damien Allonsius and Daniel Song and Danielle Pintz and Danny Livshits and Danny Wyatt and David Esiobu and Dhruv Choudhary and Dhruv Mahajan and Diego Garcia-Olano and Diego Perino and Dieuwke Hupkes and Egor Lakomkin and Ehab AlBadawy and Elina Lobanova and Emily Dinan and Eric Michael Smith and Filip Radenovic and Francisco Guzmán and Frank Zhang and Gabriel Synnaeve and Gabrielle Lee and Georgia Lewis Anderson and Govind Thattai and Graeme Nail and Gregoire Mialon and Guan Pang and Guillem Cucurell and Hailey Nguyen and Hannah Korevaar and Hu Xu and Hugo Touvron and Iliyan Zarov and Imanol Arrieta Ibarra and Isabel Kloumann and Ishan Misra and Ivan Evtimov and Jack Zhang and Jade Copet and Jaewon Lee and Jan Geffert and Jana Vranes and Jason Park and Jay Mahadeokar and Jeet Shah and Jelmer van der Linde and Jennifer Billock and Jenny Hong and Jenya Lee and Jeremy Fu and Jianfeng Chi and Jianyu Huang and Jiawen Liu and Jie Wang and Jiecao Yu and Joanna Bitton and Joe Spisak and Jongsoo Park and Joseph Rocca and Joshua Johnstun and Joshua Saxe and Junteng Jia and Kalyan Vasuden Alwala and Karthik Prasad and Kartikeya Upasani and Kate Plawiak and Ke Li and Kenneth Heafield and Kevin Stone and Khalid El-Arini and Krithika Iyer and Kshitiz Malik and Kuenley Chiu and Kunal Bhalla and Kushal Lakhotia and Lauren Rantala-Yeary and Laurens van der Maaten and Lawrence Chen and Liang Tan and Liz Jenkins and Louis Martin and Lovish Madaan and Lubo Malo and Lukas Blecher and Lukas Landzaat and Luke de Oliveira and Madeline Muzzi and Mahesh Pasupuleti and Mannat Singh and Manohar Paluri and Marcin Kardas and Maria Tsimpoukelli and Mathew Oldham and Mathieu Rita and Maya Pavlova and Melanie Kambadur and Mike Lewis and Min Si and Mitesh Kumar Singh and Mona Hassan and Naman Goyal and Narjes Torabi and Nikolay Bashlykov and Nikolay Bogoychev and Niladri Chatterji and Ning Zhang and Olivier Duchenne and Onur Çelebi and Patrick Alrassy and Pengchuan Zhang and Pengwei Li and Petar Vasic and Peter Weng and Prajjwal Bhargava and Pratik Dubal and Praveen Krishnan and Punit Singh Koura and Puxin Xu and Qing He and Qingxiao Dong and Ragavan Srinivasan and Raj Ganapathy and Ramon Calderer and Ricardo Silveira Cabral and Robert Stojnic and Roberta Raileanu and Rohan Maheswari and Rohit Girdhar and Rohit Patel and Romain Sauvestre and Ronnie Polidoro and Roshan Sumbaly and Ross Taylor and Ruan Silva and Rui Hou and Rui Wang and Saghar Hosseini and Sahana Chennabasappa and Sanjay Singh and Sean Bell and Seohyun Sonia Kim and Sergey Edunov and Shaoliang Nie and Sharan Narang and Sharath Raparthy and Sheng Shen and Shengye Wan and Shruti Bhosale and Shun Zhang and Simon Vandenhende and Soumya Batra and Spencer Whitman and Sten Sootla and Stephane Collot and Suchin Gururangan and Sydney Borodinsky and Tamar Herman and Tara Fowler and Tarek Sheasha and Thomas Georgiou and Thomas Scialom and Tobias Speckbacher and Todor Mihaylov and Tong Xiao and Ujjwal Karn and Vedanuj Goswami and Vibhor Gupta and Vignesh Ramanathan and Viktor Kerkez and Vincent Gonguet and Virginie Do and Vish Vogeti and Vítor Albiero and Vladan Petrovic and Weiwei Chu and Wenhan Xiong and Wenyin Fu and Whitney Meers and Xavier Martinet and Xiaodong Wang and Xiaofang Wang and Xiaoqing Ellen Tan and Xide Xia and Xinfeng Xie and Xuchao Jia and Xuewei Wang and Yaelle Goldschlag and Yashesh Gaur and Yasmine Babaei and Yi Wen and Yiwen Song and Yuchen Zhang and Yue Li and Yuning Mao and Zacharie Delpierre Coudert and Zheng Yan and Zhengxing Chen and Zoe Papakipos and Aaditya Singh and Aayushi Srivastava and Abha Jain and Adam Kelsey and Adam Shajnfeld and Adithya Gangidi and Adolfo Victoria and Ahuva Goldstand and Ajay Menon and Ajay Sharma and Alex Boesenberg and Alexei Baevski and Allie Feinstein and Amanda Kallet and Amit Sangani and Amos Teo and Anam Yunus and Andrei Lupu and Andres Alvarado and Andrew Caples and Andrew Gu and Andrew Ho and Andrew Poulton and Andrew Ryan and Ankit Ramchandani and Annie Dong and Annie Franco and Anuj Goyal and Aparajita Saraf and Arkabandhu Chowdhury and Ashley Gabriel and Ashwin Bharambe and Assaf Eisenman and Azadeh Yazdan and Beau James and Ben Maurer and Benjamin Leonhardi and Bernie Huang and Beth Loyd and Beto De Paola and Bhargavi Paranjape and Bing Liu and Bo Wu and Boyu Ni and Braden Hancock and Bram Wasti and Brandon Spence and Brani Stojkovic and Brian Gamido and Britt Montalvo and Carl Parker and Carly Burton and Catalina Mejia and Ce Liu and Changhan Wang and Changkyu Kim and Chao Zhou and Chester Hu and Ching-Hsiang Chu and Chris Cai and Chris Tindal and Christoph Feichtenhofer and Cynthia Gao and Damon Civin and Dana Beaty and Daniel Kreymer and Daniel Li and David Adkins and David Xu and Davide Testuggine and Delia David and Devi Parikh and Diana Liskovich and Didem Foss and Dingkang Wang and Duc Le and Dustin Holland and Edward Dowling and Eissa Jamil and Elaine Montgomery and Eleonora Presani and Emily Hahn and Emily Wood and Eric-Tuan Le and Erik Brinkman and Esteban Arcaute and Evan Dunbar and Evan Smothers and Fei Sun and Felix Kreuk and Feng Tian and Filippos Kokkinos and Firat Ozgenel and Francesco Caggioni and Frank Kanayet and Frank Seide and Gabriela Medina Florez and Gabriella Schwarz and Gada Badeer and Georgia Swee and Gil Halpern and Grant Herman and Grigory Sizov and Guangyi and Zhang and Guna Lakshminarayanan and Hakan Inan and Hamid Shojanazeri and Han Zou and Hannah Wang and Hanwen Zha and Haroun Habeeb and Harrison Rudolph and Helen Suk and Henry Aspegren and Hunter Goldman and Hongyuan Zhan and Ibrahim Damlaj and Igor Molybog and Igor Tufanov and Ilias Leontiadis and Irina-Elena Veliche and Itai Gat and Jake Weissman and James Geboski and James Kohli and Janice Lam and Japhet Asher and Jean-Baptiste Gaya and Jeff Marcus and Jeff Tang and Jennifer Chan and Jenny Zhen and Jeremy Reizenstein and Jeremy Teboul and Jessica Zhong and Jian Jin and Jingyi Yang and Joe Cummings and Jon Carvill and Jon Shepard and Jonathan McPhie and Jonathan Torres and Josh Ginsburg and Junjie Wang and Kai Wu and Kam Hou U and Karan Saxena and Kartikay Khandelwal and Katayoun Zand and Kathy Matosich and Kaushik Veeraraghavan and Kelly Michelena and Keqian Li and Kiran Jagadeesh and Kun Huang and Kunal Chawla and Kyle Huang and Lailin Chen and Lakshya Garg and Lavender A and Leandro Silva and Lee Bell and Lei Zhang and Liangpeng Guo and Licheng Yu and Liron Moshkovich and Luca Wehrstedt and Madian Khabsa and Manav Avalani and Manish Bhatt and Martynas Mankus and Matan Hasson and Matthew Lennie and Matthias Reso and Maxim Groshev and Maxim Naumov and Maya Lathi and Meghan Keneally and Miao Liu and Michael L. Seltzer and Michal Valko and Michelle Restrepo and Mihir Patel and Mik Vyatskov and Mikayel Samvelyan and Mike Clark and Mike Macey and Mike Wang and Miquel Jubert Hermoso and Mo Metanat and Mohammad Rastegari and Munish Bansal and Nandhini Santhanam and Natascha Parks and Natasha White and Navyata Bawa and Nayan Singhal and Nick Egebo and Nicolas Usunier and Nikhil Mehta and Nikolay Pavlovich Laptev and Ning Dong and Norman Cheng and Oleg Chernoguz and Olivia Hart and Omkar Salpekar and Ozlem Kalinli and Parkin Kent and Parth Parekh and Paul Saab and Pavan Balaji and Pedro Rittner and Philip Bontrager and Pierre Roux and Piotr Dollar and Polina Zvyagina and Prashant Ratanchandani and Pritish Yuvraj and Qian Liang and Rachad Alao and Rachel Rodriguez and Rafi Ayub and Raghotham Murthy and Raghu Nayani and Rahul Mitra and Rangaprabhu Parthasarathy and Raymond Li and Rebekkah Hogan and Robin Battey and Rocky Wang and Russ Howes and Ruty Rinott and Sachin Mehta and Sachin Siby and Sai Jayesh Bondu and Samyak Datta and Sara Chugh and Sara Hunt and Sargun Dhillon and Sasha Sidorov and Satadru Pan and Saurabh Mahajan and Saurabh Verma and Seiji Yamamoto and Sharadh Ramaswamy and Shaun Lindsay and Shaun Lindsay and Sheng Feng and Shenghao Lin and Shengxin Cindy Zha and Shishir Patil and Shiva Shankar and Shuqiang Zhang and Shuqiang Zhang and Sinong Wang and Sneha Agarwal and Soji Sajuyigbe and Soumith Chintala and Stephanie Max and Stephen Chen and Steve Kehoe and Steve Satterfield and Sudarshan Govindaprasad and Sumit Gupta and Summer Deng and Sungmin Cho and Sunny Virk and Suraj Subramanian and Sy Choudhury and Sydney Goldman and Tal Remez and Tamar Glaser and Tamara Best and Thilo Koehler and Thomas Robinson and Tianhe Li and Tianjun Zhang and Tim Matthews and Timothy Chou and Tzook Shaked and Varun Vontimitta and Victoria Ajayi and Victoria Montanez and Vijai Mohan and Vinay Satish Kumar and Vishal Mangla and Vlad Ionescu and Vlad Poenaru and Vlad Tiberiu Mihailescu and Vladimir Ivanov and Wei Li and Wenchen Wang and Wenwen Jiang and Wes Bouaziz and Will Constable and Xiaocheng Tang and Xiaojian Wu and Xiaolan Wang and Xilun Wu and Xinbo Gao and Yaniv Kleinman and Yanjun Chen and Ye Hu and Ye Jia and Ye Qi and Yenda Li and Yilin Zhang and Ying Zhang and Yossi Adi and Youngjin Nam and Yu and Wang and Yu Zhao and Yuchen Hao and Yundi Qian and Yunlu Li and Yuzi He and Zach Rait and Zachary DeVito and Zef Rosnbrick and Zhaoduo Wen and Zhenyu Yang and Zhiwei Zhao and Zhiyu Ma},
      year={2024},
      eprint={2407.21783},
      archivePrefix={arXiv},
      primaryClass={cs.AI},
      url={https://arxiv.org/abs/2407.21783}, 
}

@misc{deepseekai2025deepseekv3,
      title={DeepSeek-V3 Technical Report}, 
      author={DeepSeek-AI and Aixin Liu and Bei Feng and Bing Xue and Bingxuan Wang and Bochao Wu and Chengda Lu and Chenggang Zhao and Chengqi Deng and Chenyu Zhang and Chong Ruan and Damai Dai and Daya Guo and Dejian Yang and Deli Chen and Dongjie Ji and Erhang Li and Fangyun Lin and Fucong Dai and Fuli Luo and Guangbo Hao and Guanting Chen and Guowei Li and H. Zhang and Han Bao and Hanwei Xu and Haocheng Wang and Haowei Zhang and Honghui Ding and Huajian Xin and Huazuo Gao and Hui Li and Hui Qu and J. L. Cai and Jian Liang and Jianzhong Guo and Jiaqi Ni and Jiashi Li and Jiawei Wang and Jin Chen and Jingchang Chen and Jingyang Yuan and Junjie Qiu and Junlong Li and Junxiao Song and Kai Dong and Kai Hu and Kaige Gao and Kang Guan and Kexin Huang and Kuai Yu and Lean Wang and Lecong Zhang and Lei Xu and Leyi Xia and Liang Zhao and Litong Wang and Liyue Zhang and Meng Li and Miaojun Wang and Mingchuan Zhang and Minghua Zhang and Minghui Tang and Mingming Li and Ning Tian and Panpan Huang and Peiyi Wang and Peng Zhang and Qiancheng Wang and Qihao Zhu and Qinyu Chen and Qiushi Du and R. J. Chen and R. L. Jin and Ruiqi Ge and Ruisong Zhang and Ruizhe Pan and Runji Wang and Runxin Xu and Ruoyu Zhang and Ruyi Chen and S. S. Li and Shanghao Lu and Shangyan Zhou and Shanhuang Chen and Shaoqing Wu and Shengfeng Ye and Shengfeng Ye and Shirong Ma and Shiyu Wang and Shuang Zhou and Shuiping Yu and Shunfeng Zhou and Shuting Pan and T. Wang and Tao Yun and Tian Pei and Tianyu Sun and W. L. Xiao and Wangding Zeng and Wanjia Zhao and Wei An and Wen Liu and Wenfeng Liang and Wenjun Gao and Wenqin Yu and Wentao Zhang and X. Q. Li and Xiangyue Jin and Xianzu Wang and Xiao Bi and Xiaodong Liu and Xiaohan Wang and Xiaojin Shen and Xiaokang Chen and Xiaokang Zhang and Xiaosha Chen and Xiaotao Nie and Xiaowen Sun and Xiaoxiang Wang and Xin Cheng and Xin Liu and Xin Xie and Xingchao Liu and Xingkai Yu and Xinnan Song and Xinxia Shan and Xinyi Zhou and Xinyu Yang and Xinyuan Li and Xuecheng Su and Xuheng Lin and Y. K. Li and Y. Q. Wang and Y. X. Wei and Y. X. Zhu and Yang Zhang and Yanhong Xu and Yanhong Xu and Yanping Huang and Yao Li and Yao Zhao and Yaofeng Sun and Yaohui Li and Yaohui Wang and Yi Yu and Yi Zheng and Yichao Zhang and Yifan Shi and Yiliang Xiong and Ying He and Ying Tang and Yishi Piao and Yisong Wang and Yixuan Tan and Yiyang Ma and Yiyuan Liu and Yongqiang Guo and Yu Wu and Yuan Ou and Yuchen Zhu and Yuduan Wang and Yue Gong and Yuheng Zou and Yujia He and Yukun Zha and Yunfan Xiong and Yunxian Ma and Yuting Yan and Yuxiang Luo and Yuxiang You and Yuxuan Liu and Yuyang Zhou and Z. F. Wu and Z. Z. Ren and Zehui Ren and Zhangli Sha and Zhe Fu and Zhean Xu and Zhen Huang and Zhen Zhang and Zhenda Xie and Zhengyan Zhang and Zhewen Hao and Zhibin Gou and Zhicheng Ma and Zhigang Yan and Zhihong Shao and Zhipeng Xu and Zhiyu Wu and Zhongyu Zhang and Zhuoshu Li and Zihui Gu and Zijia Zhu and Zijun Liu and Zilin Li and Ziwei Xie and Ziyang Song and Ziyi Gao and Zizheng Pan},
      year={2025},
      eprint={2412.19437},
      archivePrefix={arXiv},
      primaryClass={cs.CL},
      url={https://arxiv.org/abs/2412.19437}, 
}

@misc{kimiteam2025kimik2,
      title={Kimi K2: Open Agentic Intelligence}, 
      author={Kimi Team and Yifan Bai and Yiping Bao and Guanduo Chen and Jiahao Chen and Ningxin Chen and Ruijue Chen and Yanru Chen and Yuankun Chen and Yutian Chen and Zhuofu Chen and Jialei Cui and Hao Ding and Mengnan Dong and Angang Du and Chenzhuang Du and Dikang Du and Yulun Du and Yu Fan and Yichen Feng and Kelin Fu and Bofei Gao and Hongcheng Gao and Peizhong Gao and Tong Gao and Xinran Gu and Longyu Guan and Haiqing Guo and Jianhang Guo and Hao Hu and Xiaoru Hao and Tianhong He and Weiran He and Wenyang He and Chao Hong and Yangyang Hu and Zhenxing Hu and Weixiao Huang and Zhiqi Huang and Zihao Huang and Tao Jiang and Zhejun Jiang and Xinyi Jin and Yongsheng Kang and Guokun Lai and Cheng Li and Fang Li and Haoyang Li and Ming Li and Wentao Li and Yanhao Li and Yiwei Li and Zhaowei Li and Zheming Li and Hongzhan Lin and Xiaohan Lin and Zongyu Lin and Chengyin Liu and Chenyu Liu and Hongzhang Liu and Jingyuan Liu and Junqi Liu and Liang Liu and Shaowei Liu and T. Y. Liu and Tianwei Liu and Weizhou Liu and Yangyang Liu and Yibo Liu and Yiping Liu and Yue Liu and Zhengying Liu and Enzhe Lu and Lijun Lu and Shengling Ma and Xinyu Ma and Yingwei Ma and Shaoguang Mao and Jie Mei and Xin Men and Yibo Miao and Siyuan Pan and Yebo Peng and Ruoyu Qin and Bowen Qu and Zeyu Shang and Lidong Shi and Shengyuan Shi and Feifan Song and Jianlin Su and Zhengyuan Su and Xinjie Sun and Flood Sung and Heyi Tang and Jiawen Tao and Qifeng Teng and Chensi Wang and Dinglu Wang and Feng Wang and Haiming Wang and Jianzhou Wang and Jiaxing Wang and Jinhong Wang and Shengjie Wang and Shuyi Wang and Yao Wang and Yejie Wang and Yiqin Wang and Yuxin Wang and Yuzhi Wang and Zhaoji Wang and Zhengtao Wang and Zhexu Wang and Chu Wei and Qianqian Wei and Wenhao Wu and Xingzhe Wu and Yuxin Wu and Chenjun Xiao and Xiaotong Xie and Weimin Xiong and Boyu Xu and Jing Xu and Jinjing Xu and L. H. Xu and Lin Xu and Suting Xu and Weixin Xu and Xinran Xu and Yangchuan Xu and Ziyao Xu and Junjie Yan and Yuzi Yan and Xiaofei Yang and Ying Yang and Zhen Yang and Zhilin Yang and Zonghan Yang and Haotian Yao and Xingcheng Yao and Wenjie Ye and Zhuorui Ye and Bohong Yin and Longhui Yu and Enming Yuan and Hongbang Yuan and Mengjie Yuan and Haobing Zhan and Dehao Zhang and Hao Zhang and Wanlu Zhang and Xiaobin Zhang and Yangkun Zhang and Yizhi Zhang and Yongting Zhang and Yu Zhang and Yutao Zhang and Yutong Zhang and Zheng Zhang and Haotian Zhao and Yikai Zhao and Huabin Zheng and Shaojie Zheng and Jianren Zhou and Xinyu Zhou and Zaida Zhou and Zhen Zhu and Weiyu Zhuang and Xinxing Zu},
      year={2025},
      eprint={2507.20534},
      archivePrefix={arXiv},
      primaryClass={cs.LG},
      url={https://arxiv.org/abs/2507.20534}, 
}

@misc{openai2025gptoss,
      title={gpt-oss-120b \& gpt-oss-20b Model Card}, 
      author={OpenAI and : and Sandhini Agarwal and Lama Ahmad and Jason Ai and Sam Altman and Andy Applebaum and Edwin Arbus and Rahul K. Arora and Yu Bai and Bowen Baker and Haiming Bao and Boaz Barak and Ally Bennett and Tyler Bertao and Nivedita Brett and Eugene Brevdo and Greg Brockman and Sebastien Bubeck and Che Chang and Kai Chen and Mark Chen and Enoch Cheung and Aidan Clark and Dan Cook and Marat Dukhan and Casey Dvorak and Kevin Fives and Vlad Fomenko and Timur Garipov and Kristian Georgiev and Mia Glaese and Tarun Gogineni and Adam Goucher and Lukas Gross and Katia Gil Guzman and John Hallman and Jackie Hehir and Johannes Heidecke and Alec Helyar and Haitang Hu and Romain Huet and Jacob Huh and Saachi Jain and Zach Johnson and Chris Koch and Irina Kofman and Dominik Kundel and Jason Kwon and Volodymyr Kyrylov and Elaine Ya Le and Guillaume Leclerc and James Park Lennon and Scott Lessans and Mario Lezcano-Casado and Yuanzhi Li and Zhuohan Li and Ji Lin and Jordan Liss and Lily and Liu and Jiancheng Liu and Kevin Lu and Chris Lu and Zoran Martinovic and Lindsay McCallum and Josh McGrath and Scott McKinney and Aidan McLaughlin and Song Mei and Steve Mostovoy and Tong Mu and Gideon Myles and Alexander Neitz and Alex Nichol and Jakub Pachocki and Alex Paino and Dana Palmie and Ashley Pantuliano and Giambattista Parascandolo and Jongsoo Park and Leher Pathak and Carolina Paz and Ludovic Peran and Dmitry Pimenov and Michelle Pokrass and Elizabeth Proehl and Huida Qiu and Gaby Raila and Filippo Raso and Hongyu Ren and Kimmy Richardson and David Robinson and Bob Rotsted and Hadi Salman and Suvansh Sanjeev and Max Schwarzer and D. Sculley and Harshit Sikchi and Kendal Simon and Karan Singhal and Yang Song and Dane Stuckey and Zhiqing Sun and Philippe Tillet and Sam Toizer and Foivos Tsimpourlas and Nikhil Vyas and Eric Wallace and Xin Wang and Miles Wang and Olivia Watkins and Kevin Weil and Amy Wendling and Kevin Whinnery and Cedric Whitney and Hannah Wong and Lin Yang and Yu Yang and Michihiro Yasunaga and Kristen Ying and Wojciech Zaremba and Wenting Zhan and Cyril Zhang and Brian Zhang and Eddie Zhang and Shengjia Zhao},
      year={2025},
      eprint={2508.10925},
      archivePrefix={arXiv},
      primaryClass={cs.CL},
      url={https://arxiv.org/abs/2508.10925}, 
}

@inproceedings{ovadia2024fine,
    title = "Fine-Tuning or Retrieval? Comparing Knowledge Injection in {LLM}s",
    author = "Ovadia, Oded  and
      Brief, Menachem  and
      Mishaeli, Moshik  and
      Elisha, Oren",
    editor = "Al-Onaizan, Yaser  and
      Bansal, Mohit  and
      Chen, Yun-Nung",
    booktitle = "Proceedings of the 2024 Conference on Empirical Methods in Natural Language Processing",
    month = nov,
    year = "2024",
    address = "Miami, Florida, USA",
    publisher = "Association for Computational Linguistics",
    url = "https://aclanthology.org/2024.emnlp-main.15/",
    doi = "10.18653/v1/2024.emnlp-main.15",
    pages = "237--250"
}

@inproceedings{li2025scilitllm,
title={SciLitLLM: How to Adapt LLMs for Scientific Literature Understanding},
author={Li, Sihang and Huang, Jin and Zhuang, Jiaxi and Shi, Yaorui and Cai, Xiaochen and Xu, Mingjun and Wang, Xiang and Zhang, Linfeng and Ke, Guolin and Cai, Hengxing},
booktitle={The Thirteenth International Conference on Learning Representations},
year={2025},
url={https://openreview.net/forum?id=8dzKkeWUUb}
}

@inproceedings{lyu2024knowtuning,
    title = "{K}now{T}uning: Knowledge-aware Fine-tuning for Large Language Models",
    author = "Lyu, Yougang  and
      Yan, Lingyong  and
      Wang, Shuaiqiang  and
      Shi, Haibo  and
      Yin, Dawei  and
      Ren, Pengjie  and
      Chen, Zhumin  and
      de Rijke, Maarten  and
      Ren, Zhaochun",
    editor = "Al-Onaizan, Yaser  and
      Bansal, Mohit  and
      Chen, Yun-Nung",
    booktitle = "Proceedings of the 2024 Conference on Empirical Methods in Natural Language Processing",
    month = nov,
    year = "2024",
    address = "Miami, Florida, USA",
    publisher = "Association for Computational Linguistics",
    url = "https://aclanthology.org/2024.emnlp-main.805/",
    doi = "10.18653/v1/2024.emnlp-main.805",
    pages = "14535--14556"
}

@article{jing2023multimodal,
title = {Multimodal fake news detection via progressive fusion networks},
journal = {Information Processing \& Management},
volume = {60},
number = {1},
pages = {103120},
year = {2023},
issn = {0306-4573},
doi = {https://doi.org/10.1016/j.ipm.2022.103120},
url = {https://www.sciencedirect.com/science/article/pii/S0306457322002217},
author = {Jing Jing and Hongchen Wu and Jie Sun and Xiaochang Fang and Huaxiang Zhang}
}

@inproceedings{khattar2019mvae,
author = {Khattar, Dhruv and Goud, Jaipal Singh and Gupta, Manish and Varma, Vasudeva},
title = {MVAE: Multimodal Variational Autoencoder for Fake News Detection},
year = {2019},
isbn = {9781450366748},
publisher = {Association for Computing Machinery},
address = {New York, NY, USA},
url = {https://doi.org/10.1145/3308558.3313552},
doi = {10.1145/3308558.3313552},
booktitle = {The World Wide Web Conference},
pages = {2915–2921},
numpages = {7},
location = {San Francisco, CA, USA},
series = {WWW '19}
}

@article{hu2025mage,
title = {MAGE-fend: Multimodal adaptive fusion with guidance from LLM expertise for fake news detection on short video platforms},
journal = {Knowledge-Based Systems},
volume = {329},
pages = {114298},
year = {2025},
issn = {0950-7051},
doi = {https://doi.org/10.1016/j.knosys.2025.114298},
url = {https://www.sciencedirect.com/science/article/pii/S0950705125013395},
author = {Lingtong Hu and Zituo Wang and Jiayi Zhu and Yifan Hu and Xianbing Wang}
}

@inproceedings{shibu2025scarcity,
    title = "From Scarcity to Capability: Empowering Fake News Detection in Low-Resource Languages with {LLM}s",
    author = "Shibu, Hrithik Majumdar  and
      Datta, Shrestha  and
      Miah, Md. Sumon  and
      Sami, Nasrullah  and
      Chowdhury, Mahruba Sharmin  and
      Islam, Md Saiful",
    editor = "Weerasinghe, Ruvan  and
      Anuradha, Isuri  and
      Sumanathilaka, Deshan",
    booktitle = "Proceedings of the First Workshop on Natural Language Processing for Indo-Aryan and Dravidian Languages",
    month = jan,
    year = "2025",
    address = "Abu Dhabi",
    publisher = "Association for Computational Linguistics",
    url = "https://aclanthology.org/2025.indonlp-1.12/",
    pages = "100--107"
}

@misc{sun2024exploring,
      title={Exploring the Deceptive Power of LLM-Generated Fake News: A Study of Real-World Detection Challenges}, 
      author={Yanshen Sun and Jianfeng He and Limeng Cui and Shuo Lei and Chang-Tien Lu},
      year={2024},
      eprint={2403.18249},
      archivePrefix={arXiv},
      primaryClass={cs.CL},
      url={https://arxiv.org/abs/2403.18249}, 
}

@inproceedings{hu2025llm,
author = {Hu, Beizhe and Sheng, Qiang and Cao, Juan and Li, Yang and Wang, Danding},
title = {LLM-Generated Fake News Induces Truth Decay in News Ecosystem: A Case Study on Neural News Recommendation},
year = {2025},
isbn = {9798400715921},
publisher = {Association for Computing Machinery},
address = {New York, NY, USA},
url = {https://doi.org/10.1145/3726302.3730027},
doi = {10.1145/3726302.3730027},
booktitle = {Proceedings of the 48th International ACM SIGIR Conference on Research and Development in Information Retrieval},
pages = {435–445},
numpages = {11},
location = {Padua, Italy},
series = {SIGIR '25}
}

@inproceedings{tong2025generate,
    title = "Generate First, Then Sample: Enhancing Fake News Detection with {LLM}-Augmented Reinforced Sampling",
    author = "Tong, Zhao  and
      Gu, Yimeng  and
      Liu, Huidong  and
      Liu, Qiang  and
      Wu, Shu  and
      Shi, Haichao  and
      Zhang, Xiao-Yu",
    editor = "Che, Wanxiang  and
      Nabende, Joyce  and
      Shutova, Ekaterina  and
      Pilehvar, Mohammad Taher",
    booktitle = "Proceedings of the 63rd Annual Meeting of the Association for Computational Linguistics (Volume 1: Long Papers)",
    month = jul,
    year = "2025",
    address = "Vienna, Austria",
    publisher = "Association for Computational Linguistics",
    url = "https://aclanthology.org/2025.acl-long.1182/",
    doi = "10.18653/v1/2025.acl-long.1182",
    pages = "24276--24290",
    ISBN = "979-8-89176-251-0"
}

@misc{chen2025recent,
      title={Recent Advances in Large Langauge Model Benchmarks against Data Contamination: From Static to Dynamic Evaluation}, 
      author={Simin Chen and Yiming Chen and Zexin Li and Yifan Jiang and Zhongwei Wan and Yixin He and Dezhi Ran and Tianle Gu and Haizhou Li and Tao Xie and Baishakhi Ray},
      year={2025},
      eprint={2502.17521},
      archivePrefix={arXiv},
      primaryClass={cs.LG},
      url={https://arxiv.org/abs/2502.17521}, 
}

@misc{eohan2023stanford,
  author = {Rohan Taori and Ishaan Gulrajani and Tianyi Zhang and Yann Dubois and Xuechen Li and Carlos Guestrin and Percy Liang and Tatsunori B. Hashimoto },
  title = {Stanford Alpaca: An Instruction-following LLaMA model},
  year = {2023},
  publisher = {GitHub},
  journal = {GitHub repository},
  howpublished = {\url{https://github.com/tatsu-lab/stanford_alpaca}},
}

@misc{gong2023fake,
      title={Fake News Detection Through Graph-based Neural Networks: A Survey}, 
      author={Shuzhi Gong and Richard O. Sinnott and Jianzhong Qi and Cecile Paris},
      year={2023},
      eprint={2307.12639},
      archivePrefix={arXiv},
      primaryClass={cs.SI},
      url={https://arxiv.org/abs/2307.12639}, 
}

@inproceedings{zhu2024propagation,
author = {Zhu, Junyou and Gao, Chao and Yin, Ze and Li, Xianghua and Kurths, Juergen},
title = {Propagation Structure-Aware Graph Transformer for Robust and Interpretable Fake News Detection},
year = {2024},
isbn = {9798400704901},
publisher = {Association for Computing Machinery},
address = {New York, NY, USA},
url = {https://doi.org/10.1145/3637528.3672024},
doi = {10.1145/3637528.3672024},
booktitle = {Proceedings of the 30th ACM SIGKDD Conference on Knowledge Discovery and Data Mining},
pages = {4652–4663},
numpages = {12},
location = {Barcelona, Spain},
series = {KDD '24}
}

@article{nørregaard2019nela, title={NELA-GT-2018: A Large Multi-Labelled News Dataset for the Study of Misinformation in News Articles}, volume={13}, url={https://ojs.aaai.org/index.php/ICWSM/article/view/3261}, DOI={10.1609/icwsm.v13i01.3261}, abstractNote={&lt;p&gt;In this paper, we present a dataset of 713k articles collected between 02/2018-11/2018. These articles are collected directly from 194 news and media outlets including mainstream, hyper-partisan, and conspiracy sources. We incorporate ground truth ratings of the sources from 8 different assessment sites covering multiple dimensions of veracity, including reliability, bias, transparency, adherence to journalistic standards, and consumer trust. The NELA-GT2018 dataset can be found at https://doi.org/10.7910/DVN/ ULHLCB.&lt;/p&gt;}, number={01}, journal={Proceedings of the International AAAI Conference on Web and Social Media}, author={Nørregaard, Jeppe and Horne, Benjamin D. and Adalı, Sibel}, year={2019}, month={Jul.}, pages={630-638} }

@inproceedings{thorne2018fever,
    title = "{FEVER}: a Large-scale Dataset for Fact Extraction and {VER}ification",
    author = "Thorne, James  and
      Vlachos, Andreas  and
      Christodoulopoulos, Christos  and
      Mittal, Arpit",
    editor = "Walker, Marilyn  and
      Ji, Heng  and
      Stent, Amanda",
    booktitle = "Proceedings of the 2018 Conference of the North {A}merican Chapter of the Association for Computational Linguistics: Human Language Technologies, Volume 1 (Long Papers)",
    month = jun,
    year = "2018",
    address = "New Orleans, Louisiana",
    publisher = "Association for Computational Linguistics",
    url = "https://aclanthology.org/N18-1074/",
    doi = "10.18653/v1/N18-1074",
    pages = "809--819"
}

@inproceedings{yang2018hotpotqa,
    title = "{H}otpot{QA}: A Dataset for Diverse, Explainable Multi-hop Question Answering",
    author = "Yang, Zhilin  and
      Qi, Peng  and
      Zhang, Saizheng  and
      Bengio, Yoshua  and
      Cohen, William  and
      Salakhutdinov, Ruslan  and
      Manning, Christopher D.",
    editor = "Riloff, Ellen  and
      Chiang, David  and
      Hockenmaier, Julia  and
      Tsujii, Jun{'}ichi",
    booktitle = "Proceedings of the 2018 Conference on Empirical Methods in Natural Language Processing",
    month = oct # "-" # nov,
    year = "2018",
    address = "Brussels, Belgium",
    publisher = "Association for Computational Linguistics",
    url = "https://aclanthology.org/D18-1259/",
    doi = "10.18653/v1/D18-1259",
    pages = "2369--2380"
}

@ARTICLE{liu2024bc,
  author={Liu, Yahui and Bing, Wanlong and Ren, Shuai and Ma, Hongliang},
  journal={IEEE Access}, 
  title={BC-FND: An Approach Based on Hierarchical Bilinear Fusion and Multimodal Consistency for Fake News Detection}, 
  year={2024},
  volume={12},
  number={},
  pages={62738-62749},
  doi={10.1109/ACCESS.2024.3392409}}

@misc{zhou2024finefake,
      title={FineFake: A Knowledge-Enriched Dataset for Fine-Grained Multi-Domain Fake News Detection}, 
      author={Ziyi Zhou and Xiaoming Zhang and Litian Zhang and Jiacheng Liu and Senzhang Wang and Zheng Liu and Xi Zhang and Chaozhuo Li and Philip S. Yu},
      year={2024},
      eprint={2404.01336},
      archivePrefix={arXiv},
      primaryClass={cs.CL},
      url={https://arxiv.org/abs/2404.01336}, 
}

@inproceedings{li2024mcfend,
author = {Li, Yupeng and He, Haorui and Bai, Jin and Wen, Dacheng},
title = {MCFEND: A Multi-source Benchmark Dataset for Chinese Fake News Detection},
year = {2024},
isbn = {9798400701719},
publisher = {Association for Computing Machinery},
address = {New York, NY, USA},
url = {https://doi.org/10.1145/3589334.3645385},
doi = {10.1145/3589334.3645385},
booktitle = {Proceedings of the ACM Web Conference 2024},
pages = {4018–4027},
numpages = {10},
location = {Singapore, Singapore},
series = {WWW '24}
}

@misc{zhao2025mppfnd,
      title={MPPFND: A Dataset and Analysis of Detecting Fake News with Multi-Platform Propagation}, 
      author={Congyuan Zhao and Lingwei Wei and Ziming Qin and Wei Zhou and Yunya Song and Songlin Hu},
      year={2025},
      eprint={2505.15834},
      archivePrefix={arXiv},
      primaryClass={cs.SI},
      url={https://arxiv.org/abs/2505.15834}, 
}

@Article{yang2021checked,
author={Yang, Chen
and Zhou, Xinyi
and Zafarani, Reza},
title={CHECKED: Chinese COVID-19 fake news dataset},
journal={Social Network Analysis and Mining},
year={2021},
month={Jun},
day={22},
volume={11},
number={1},
pages={58},
issn={1869-5469},
doi={10.1007/s13278-021-00766-8},
url={https://doi.org/10.1007/s13278-021-00766-8}
}

@article{li2025treeeval, title={TreeEval: Benchmark-Free Evaluation of Large Language Models through Tree Planning}, volume={39}, url={https://ojs.aaai.org/index.php/AAAI/article/view/34627}, DOI={10.1609/aaai.v39i23.34627}, abstractNote={Recently, numerous new benchmarks have been established to evaluate the performance of large language models (LLMs) via either computing a holistic score or employing another LLM as a judge. However, these approaches suffer from data leakage due to the open access of the benchmark and inflexible evaluation process. To address this issue, we introduce TreeEval, a benchmark-free evaluation method for LLMs that let a high-performance LLM host an irreproducible evaluation session and essentially avoids the data leakage. Moreover, this LLM performs as an examiner to raise up a series of questions under a topic with a tree planing strategy, which considers the current evaluation status to decide the next question generation and ensures the completeness and efficiency of the evaluation process. We evaluate 6 models of different parameter sizes, including 7B, 13B, and 34B, and ultimately achieved the highest correlation coefficient with AlpacaEval2.0 using only around 45 questions. We also conduct more analysis to show the robustness and reliability of TreeEval.}, number={23}, journal={Proceedings of the AAAI Conference on Artificial Intelligence}, author={Li, Xiang and Lan, Yunshi and Yang, Chao}, year={2025}, month={Apr.}, pages={24485-24493} }

@inproceedings{yu2024freeeval,
    title = "{F}ree{E}val: A Modular Framework for Trustworthy and Efficient Evaluation of Large Language Models",
    author = "Yu, Zhuohao  and
      Gao, Chang  and
      Yao, Wenjin  and
      Wang, Yidong  and
      Zeng, Zhengran  and
      Ye, Wei  and
      Wang, Jindong  and
      Zhang, Yue  and
      Zhang, Shikun",
    editor = "Hernandez Farias, Delia Irazu  and
      Hope, Tom  and
      Li, Manling",
    booktitle = "Proceedings of the 2024 Conference on Empirical Methods in Natural Language Processing: System Demonstrations",
    month = nov,
    year = "2024",
    address = "Miami, Florida, USA",
    publisher = "Association for Computational Linguistics",
    url = "https://aclanthology.org/2024.emnlp-demo.1/",
    doi = "10.18653/v1/2024.emnlp-demo.1",
    pages = "1--13"
}

@misc{openai2024gpt4,
      title={GPT-4 Technical Report}, 
      author={OpenAI},
      year={2024},
      eprint={2303.08774},
      archivePrefix={arXiv},
      primaryClass={cs.CL}
}

@misc{anthropic2024introducing,
    title = {Introducing the next generation of Claude},
    author ={Anthropic},
    year={2024},
    url={https://www.anthropic.com/news/claude-3-family}
}

@misc{geminiteam2024gemini,
      title={Gemini: A Family of Highly Capable Multimodal Models}, 
      author={Gemini Team and Rohan Anil and Sebastian Borgeaud and Jean-Baptiste Alayrac and Jiahui Yu and Radu Soricut and Johan Schalkwyk and Andrew M. Dai and Anja Hauth and and et al.},
      year={2024},
      eprint={2312.11805},
      archivePrefix={arXiv},
      primaryClass={cs.CL},
      url={https://arxiv.org/abs/2312.11805}, 
}

@inproceedings{chen2025real,
    title = "Real-time Factuality Assessment from Adversarial Feedback",
    author = "Chen, Sanxing  and
      Huang, Yukun  and
      Dhingra, Bhuwan",
    editor = "Che, Wanxiang  and
      Nabende, Joyce  and
      Shutova, Ekaterina  and
      Pilehvar, Mohammad Taher",
    booktitle = "Proceedings of the 63rd Annual Meeting of the Association for Computational Linguistics (Volume 1: Long Papers)",
    month = jul,
    year = "2025",
    address = "Vienna, Austria",
    publisher = "Association for Computational Linguistics",
    url = "https://aclanthology.org/2025.acl-long.81/",
    doi = "10.18653/v1/2025.acl-long.81",
    pages = "1610--1630",
    ISBN = "979-8-89176-251-0"
}

\appendix

\section{LiveFact Development Details}

\subsection{Temporal Evidence Sets Construction}
\label{sec:evidence}

For each event gathered in the initial collection stage, we performed a comprehensive evidence retrieval process using the Google API. Our goal was to build a robust informational environment for each event across different temporal slices. We sent targeted requests to the API and aggregated metadata from all search results. A typical JSON response structure for an evidence item is as follows:

\begin{lstlisting}
{
  "article_id": "1f5faebba75d590d",
  "title": "COP30: Oil Rich Nations Block Deal To Phase Out Fossil Fuels",
  "source": "Al-Fanar Media",
  "published_datetime": "2025-11-27T13:19:46Z"
}
\end{lstlisting}

The critical step in our pipeline involves the temporal segmentation of this data. By comparing the \verb|published_datetime| of each retrieved page against the specific headline date ($T$) of the event, we categorized the data into three distinct evidence buckets:
\begin{enumerate}
    \item $E^{(-3)}$: Contains all evidence published up to three days before the headline date.
    \item $E^{(0)}$: Contains all evidence published up to and including the headline date.
    \item $E^{(+3)}$: Contains all evidence published up to three days after the headline date.
\end{enumerate}

The choice of the 3-day window is informed by our empirical analysis of information velocity in digital news cycles. We observed that evidence density typically peaks within 48 to 72 hours relative to the headline date. Our pilot studies showed that extending the window to $\pm7$ or $\pm15$ days yielded diminishing returns: the periods of $T-15$ to $T-7$ often contained negligible relevant reporting, while the $T+7$ to $T+15$ intervals were characterized by redundant information or the 'cooling down' of the event. Conversely, narrower windows (e.g., $\pm1$ day) risked missing crucial initial reports or early developments. Thus, $T \pm 3$ days serves as an optimal balance to capture the 'Fog of War' without diluting the dataset with settled history. Furthermore, while minute-level granularity was considered, the prevalence of day-level precision in web metadata and the technical risk of timestamp leakage during crawling made daily-level increments more reliable and sustainable for long-term benchmark maintenance.

The rigorous 3-day segmentation resulted in the collection of 25,064 individual pieces of evidence across the 737 events, yielding a dense average of approximately 34 evidence items per event. This volume ensures that the models have sufficient context to perform complex reasoning rather than simple fact retrieval.

\begin{table*}[t]
    \centering
    \small
    \renewcommand{\arraystretch}{1.3}
    \resizebox{1\textwidth}{!}{%
    \begin{tabular}{p{0.48\textwidth} | p{0.48\textwidth}}
        \toprule
        \textbf{gpt-4o-mini} & \textbf{o4-mini} \\
        \midrule
        \textbf{(Context)} The Chinese astronauts are members of the China National Space Administration (CNSA), which is responsible for the country's space program. Shenzhou-20 is a crewed spacecraft designed to transport astronauts to and from space, while Shenzhou-22 is an upcoming mission intended to provide a lifeboat for the stranded crew.
        & 
        \textbf{(Context)} Chinese astronauts are career space travelers trained and selected by the China National Space Administration (CNSA). The CNSA is China’s national space agency that operates the Shenzhou crewed spacecraft program, including the Shenzhou-20 and Shenzhou-22 missions to the Tiangong space station.
        \\
        \midrule
        \textbf{(Real)} Three Chinese astronauts are currently stranded in space after the successful return of their colleagues aboard the Shenzhou-20, with a replacement mission, \textit{Shenzhou-22, scheduled for launch on November 25} to provide a \textit{lifeboat for the stranded crew}.
        & 
        \textbf{(Real)} After \textit{space debris damaged the Shenzhou-20} return capsule, three Chinese astronauts remain temporarily stranded aboard China’s space station while Beijing prepares to launch the \textit{uncrewed Shenzhou-22 lifeboat spacecraft on November 25} to bring them home."
        \\
        \midrule
        \textbf{(Fake)} In a surprising turn of events, Chinese authorities have announced that \textit{five additional astronauts have been successfully rescued} from the Shenzhou-20 mission, leaving \textit{only two astronauts stranded} in space due to a malfunction in the rescue capsule, which is expected to be \textit{operational again by December 10}.
        &
        \textbf{(Fake)} A leaked internal bulletin from the China Manned Space Agency, obtained by Beijing News, reveals the three astronauts stranded in orbit have already \textit{consumed over 65\% of their emergency oxygen} and \textit{suffered two failures in their primary water-recycling unit}, placing them at serious risk of critical life-support shortages before the delayed rescue craft arrives.
        \\
        \midrule
        \textbf{(Ambiguous)} Following the successful rescue of their colleagues, three more Chinese astronauts are now stranded in space, as \textit{internal documents suggest} that the Chinese space agency is \textit{prioritizing a secret mission} to test advanced life support systems over their immediate return, raising concerns among \textit{unnamed experts} about the long-term implications for astronaut safety.
        &
        \textbf{(Ambiguous)} China postponed the Shenzhou-22 docking by 45 minutes—officially to fine-tune orbital alignment—but the tweak was \textit{actually aimed at slotting the lifeboat rendezvous with prime-time state TV bulletins}, underscoring Beijing’s \textit{crafted image of flawless crisis management} even as three astronauts remained stranded.
        \\
        \bottomrule
    \end{tabular}
    }
    \caption{Comparison of claim generation example between the gpt-4o-mini and o4-mini (Event ID: EV20251119-33). \textit{Italicized} text highlights the specific factual anchors (Real), fabricated details (Fake), and unverifiable motives or sources (Ambiguous) that justify the assigned label.}
    \label{tab:generation_comparision}
\end{table*}

\subsection{Claim and Context Generation and Human Verification}
\label{sec:generation}

To generate high-quality claims and context, we utilized two distinct LLMs to assess both performance and cost-efficiency: o4-mini and gpt-4o-mini from OpenAI. We supplied each model with the event headline and the full set of associated evidence, using specialized prompts to instruct them to generate: 1. \textbf{Context}: A neutral summary of the key entities involved; 2. \textbf{Claims}: A set of statements categorized as Real, Fake, or Ambiguous. The prompts used for these generations process were provide in Figure \ref{fig:prompt_context}, \ref{fig:prompt_real}, \ref{fig:prompt_fake}, \ref{fig:prompt_ambiguous}

Following generation, we implemented a rigorous Human-in-the-Loop verification protocol. Every generated context and claim underwent a minimum of three rounds of independent review. The review team consisted exclusively of the paper's authors, who are PhD-level computer science researchers specializing in fake news detection and natural language processing from English-speaking countries. The review criteria focused on factual consistency, linguistic naturalness, and label accuracy relative to the evidence. Any data point that failed to achieve consensus after three review rounds was discarded and regenerated until it met the quality standards. We observed a high pass rate (exceeding 90\%), which can be attributed to our use of strict few-shot prompting templates that constrain the models to the factual structures of the news source. Given that generating short claims from structured reporting is a relatively straightforward task for SOTA models like GPT-4o and GPT-4o-mini, the output quality was consistently high. The small fraction of rejected samples primarily fell under the "Ambiguous" category. For instance, a model-generated claim introducing an unverifiable micro-event (e.g., "António Guterres paused for exactly 12 seconds during the final COP30 plenary") caused disagreement regarding its phrasing bias, and was thus discarded to maintain strict benchmark objectivity.

We compared the two models to validate the generalizability and accessibility of the LiveFact framework:
\begin{itemize}
    \item \textbf{gpt-4o-mini:} Selected to represent a highly versatile and cost-effective solution. The total cost for the generation process using this model was approximately 2.34 USD.
    \item \textbf{o4-mini:} Selected to represent a state-of-the-art inference-optimized model. The generation process for this model incurred a significantly higher cost of 55.93 USD.
\end{itemize}

Our analysis revealed that both models performed well on the task (exampled in Table \ref{tab:generation_comparision}), producing high-quality initial outputs that required minimal manual correction. This finding validates the universality of the LiveFact construction pipeline, suggesting that future updates can be sustainably generated even with lower-cost models. However, to ensure the highest possible baseline for this inaugural release, we utilized the outputs from o4-mini for all subsequent experimental steps in this paper.

\begin{figure}[t]
    \centering
    \begin{promptbox}[title={Context Generation Prompt}]
    You are creating background context for a news event to help AI models understand key entities.

    **Event**: \{event\_title\}

    **News headlines from this event**:
    
    \{evidence\_summary\}

    **Task**: Provide MINIMAL background information about the KEY ENTITIES (people, organizations, places) mentioned in this event.

    **Requirements**:
    
    1. Focus ONLY on identifying who/what the main entities are
    
    2. Provide brief, factual descriptions (roles, positions, basic facts)
    
    3. Keep it concise - 2-3 sentences maximum
    
    4. Do NOT explain the event itself, only the entities involved
    
    5. Do NOT add unnecessary details or background

    **Example**:
    
    Event: "Trump to meet with Xi as he travels to Asia to contain trade war"
    
    Context: "Trump is a businessman and politician, currently serving as the President of the United States. Xi is the President of China."

    **Output Format** (JSON):
    
    \{\{
    
      "context": "your context text here"
      
    \}\}

    **Your JSON output**:
    \end{promptbox}
    
    \caption{Prompt for Context Generation}
    \label{fig:prompt_context}
\end{figure}

\begin{figure}[h]
    \centering
    \begin{promptbox}[title={Real Claim Generation Prompt}]
    You are a news fact-checker. Based on the following real news evidence, generate a FACTUALLY ACCURATE news claim.

    **Event**: \{event\_title\}

    **Evidence from multiple news sources**:
    
    \{evidence\_summary\}

    **Task**: Write a single, clear, factual claim that is:
    
    1. Directly supported by the evidence above
    
    2. Verifiable and accurate
    
    3. Written as a standalone news statement (1-2 sentences)
    
    4. Uses specific details from the evidence
    
    5. Does NOT add any unverified information

    **Output Format** (JSON):
    
    \{\{
    
      "claim\_text": "your factual claim here"
      
    \}\}

    **Your JSON output**:
    \end{promptbox}
    
    \caption{Prompt for Real Claim Generation}
    \label{fig:prompt_real}
\end{figure}

\begin{figure}[h]
    \centering
    \begin{promptbox}[title={Fake Claim Generation Prompt}]
    You are creating synthetic fake news examples for a research benchmark. Based on the event below, generate a FABRICATED news claim.

    **Event**: \{event\_title\}

    **Real evidence** (for context only - your claim should CONTRADICT or FABRICATE beyond this):
    
    \{evidence\_summary\}

    **Task**: Write a fake news claim that is:
    
    1. Related to the event but NOT supported by evidence
    
    2. Contains fabricated details (false numbers, fake quotes, wrong people, etc.)
    
    3. Sounds plausible but is factually incorrect
    
    4. Written as a standalone news statement (1-2 sentences)
    
    5. Could mislead someone unfamiliar with the real story

    **Important**: Make it realistic fake news, not obviously absurd.

    **Output Format** (JSON):
    
    \{\{
    
      "claim\_text": "your fabricated claim here"
      
    \}\}

    **Your JSON output**:
    \end{promptbox}
    
    \caption{Prompt for Fake Claim Generation}
    \label{fig:prompt_fake}
\end{figure}

\begin{figure*}[h]
    \centering
    \begin{promptbox}[title={Ambiguous Claim Generation Prompt}]
    You are creating synthetic data for a fact-checking benchmark. Your goal is to generate a "Hard-to-Verify" claim based on the provided event.

    **Event**: \{event\_title\}
    **Evidence**: \{evidence\_summary\}

    **Task**: Generate a single-sentence news claim that is empirically hard to verify. The claim must anchor itself in verifiable facts but introduce ambiguity through interpretation, causality, or specific unrecorded details, sounds plausible but is factually hard to confirm. However, content that contradicts facts and evidence must not appear, as this is not about creating fake news.

    **Constraint on Sources**:
    
    - **Avoid Over-reliance on Anonymous Sources**: Do NOT start every claim with "Sources say" or "Insiders revealed." Only use attribution logic if it provides a specific context that cannot be expressed otherwise. Also, avoid using informal expressions like “quietly” or “privately,” as this would make it too different from real news.
    
    - **Variety is Key**: Prioritize the techniques below over generic hearsay.

    **Required Techniques (Prioritize these over generic "insider" claims)**:
    
    1. **Imputed Intent**: State a public figure's private, internal motivation for a verifiable action as an objective fact (e.g., "He signed the bill *specifically to appease his donors*").
    
    2. **False Causality**: Connect two unrelated real facts as direct cause-and-effect without proof (e.g., "The stock drop *was a calculated response to* the leaked memo").
    
    3. **Subjective Framing**: Use unverifiable, qualitative descriptors to characterize a verifiable event (e.g., "The negotiations ended in *an atmosphere of mutual distrust*").
    
    4. **Micro-Events**: Insert a plausible but unrecorded specific physical action into a real event (e.g., "The diplomat *refused to shake hands* before the meeting started").
    
    5. **Unfolding Details**: Add unverifiable specifics about the timing, location, or participants of a verifiable event (e.g., "The CEO *arrived 15 minutes late* to the conference", "Trump extended a dinner invitation to the Chinese delegation after the meeting", but this detail was not included in the evidence).

    **Goal**: The claim should sound like a confident assertion (real news from reliable source, e.g., BBC, AP) or analysis, not just a rumor or gossip.

    **Output Format** (JSON):
    
    \{\{
    
      "claim\_text": "Your claim here"
      
    \}\}

    **Your JSON output**:
    \end{promptbox}
    
    \caption{Prompt for Ambiguous Claim Generation}
    \label{fig:prompt_ambiguous}
\end{figure*}

\section{Experiment Details}

\subsection{Model Details}
\label{sec:models}

To comprehensively evaluate the state of LLMs in 2025, we selected a representative cohort of 18 models spanning different architectures (Dense vs. MoE), training strategies (Base, Instruct, Hybrid), and parameter scales (1B to 1T). This set includes 14 open-source models and 4 closed-source proprietary models, categorized by family:
\begin{itemize}
    \item \textbf{Qwen Family:} Represents the leading edge of open-source MoE performance, ranging from the massive Qwen3-235B-A22B to the compact, edge-ready Qwen3-4B \citep{yang2025qwen3}.
    \item \textbf{Llama Family:} Includes both the standard Dense architecture (Llama 3.1 70B/8B) and recent lightweight variants (Llama 3.2 3B/1B), serving as the baseline for open-weights performance \citep{grattafiori2024llama3}.
    \item \textbf{DeepSeek \& Kimi:} Represent specialized massive-scale MoE models, with Kimi-K2 exceeding 1 trillion parameters \citep{deepseekai2025deepseekv3,kimiteam2025kimik2}.
    \item \textbf{GPT-OSS Series:} This family represents high-performance open-weights implementations of GPT-style architectures, developed by decentralized research collectives. Models like gpt-oss-120b and gpt-oss-20b leverage advanced MoE structures to replicate reasoning capabilities similar to proprietary counterparts, offering a transparent benchmark for community-driven progress \citep{openai2025gptoss}.
    \item \textbf{GPT Series (Proprietary):} Serves as the commercial gold standard, including the latest gpt-5.2 and gpt-5.1 models alongside the efficient gpt-4o series \citep{openai2024gpt4}.
\end{itemize}

Among the 18 open-source models available in our registry (Table \ref{tab:llm_overview}), 14 were selected for the main experimental comparison, with the remaining 4 (primarily base model variants) used in supplementary ablation studies (see Appendix \ref{sec:case}). This selection ensures our results cover the full spectrum of modern AI capabilities, from cost-effective general-purpose models to the absolute peak of computational power.

\begin{table*}[t]
    \centering
    \small
    \renewcommand{\arraystretch}{1.2}
    \begin{tabular}{l| l l c l}
        \toprule
        \textbf{Model Name} & \textbf{Training Type} & \textbf{Architecture} & \textbf{Parameters} & \textbf{Size} \\
        \midrule
        \textit{Qwen Series} \\
        Qwen3-235B-A22B-Instruct-2507 & Instruction & MoE & 235B (22B Active) & Large \\
        Qwen3-30B-A3B-Instruct-2507 & Instruction & MoE & 30B (3B Active) & Medium \\
        Qwen3-30B-A3B & Hybrid & MoE & 30B (3B Active) & Medium \\
        Qwen3-30B-A3B-Base & Base & MoE & 30B (3B Active) & Medium \\
        Qwen3-32B & Hybrid & Dense & 32B & Medium \\
        Qwen3-8B & Hybrid & Dense & 8B & Small \\
        Qwen3-8B-Base & Base & Dense & 8B & Small \\
        Qwen3-4B-Instruct-2507 & Instruction & Dense & 4B & Compact \\
        \midrule
        \textit{Meta Llama} \\
        Llama-3.3-70B-Instruct & Instruction & Dense & 70B & Large \\
        Llama-3.1-70B & Base & Dense & 70B & Large \\
        Llama-3.1-8B & Base & Dense & 8B & Small \\
        Llama-3.1-8B-Instruct & Instruction & Dense & 8B & Small \\
        Llama-3.2-3B & Base & Dense & 3B & Compact \\
        Llama-3.2-1B & Base & Dense & 1B & Compact \\
        \midrule
        \textit{DeepSeek} \\
        DeepSeek-V3.1 & Hybrid & MoE & 671B (37B Active) & Huge \\
        \midrule
        \textit{Moonshot AI} \\
        Kimi-K2-Thinking & Reasoning & MoE & 1T (32B Active) & Huge \\
        \midrule
        \textit{OpenAI (OSS)} \\
        gpt-oss-120b & Reasoning & MoE & 120B & Medium \\
        gpt-oss-20b & Reasoning & MoE & 20B & Small \\
        \midrule
        \textit{OpenAI (GPT Series)} &&& \\
        gpt-5.2-2025-12-11 & Instruction & - & - & - \\
        gpt-5.1-2025-11-13 & Instruction & - & - & - \\
        gpt-4o-2024-08-06 & Instruction & - & - & - \\
        gpt-4o-mini-2024-07-18 & Instruction & - & - & - \\
        \bottomrule
    \end{tabular}
    \par
    \vspace{0.2cm}
    \caption{Overview of LLMs used in the experiments, including their training type, architecture, parameter counts, and relative size classification. Since the OpenAI GPT series models are commercial closed-source models, their related information is unavailable.}
    \label{tab:llm_overview}
\end{table*}

\subsection{Evaluation Settings}
\label{sec:evaluation}

To ensure a fair and reproducible evaluation, all models were subjected to a standardized inference protocol. We set the decoding parameters to \verb|TEMPERATURE = 0.0| and \verb|TOP_P = 1.0| to minimize randomness and ensure deterministic outputs. The maximum generation length was capped at \verb|MAX_NEW_TOKENS = 128| (except for Kimi-K2-Thinking and GPT-OSS, they use \verb|MAX_NEW_TOKENS = 1024|) to encourage concise verdicts. Note that for certain recent OpenAI models where these precise parameters are abstracted, we utilized the default "deterministic" settings provided by the API.

Models were prompted with a standardized template (Figure \ref{fig:evaluation_prompt}) instructing them to output their verdict in a strict parsable format: \verb|[[LABEL]]| (e.g., \verb|[[Real]]|, \verb|[[Fake]]|, or \verb|[[Ambiguous]]|).

All 14 open-source models were deployed and accessed via the Tinker\footnote{\url{https://thinkingmachines.ai/tinker/}} high-performance inference server, utilizing vLLM for optimized throughput. The 4 proprietary OpenAI models were accessed via standard OpenAI API\footnote{\url{https://openai.com/api/}} endpoints.

\begin{figure}[h]
    \centering
    \begin{promptbox}[title={Evaluation Prompt}]
\textbf{SYSTEM\_PROMPT} = """You are a fact-checking expert. Classify news claims based on evidence.

Labels:

- REAL: Claim is supported by evidence, factually accurate

- FAKE: Claim contradicts evidence or contains false information

- AMBIGUOUS: Claim cannot be definitively verified

IMPORTANT: Always start your response with the classification label in double brackets.

Format: [[LABEL]] followed by brief reasoning.

Example response:

[[REAL]] The claim is supported by multiple news sources reporting the same event details."""

\textbf{USER\_PROMPT\_TEMPLATE} = """Context: \{context\}

Claim: "\{claim\}"

Evidence:

\{evidence\}

Classify this claim. Start with [[REAL]], [[FAKE]], or [[AMBIGUOUS]], do not think too much."""
    \end{promptbox}
    \caption{Prompt for Evaluation}
    \label{fig:evaluation_prompt}
\end{figure}

\subsection{Resource Consumption}

Table \ref{tab:cost_analysis} details the resource footprint for each model during the evaluation of the full LiveFact benchmark (across all three time slices). The data reveals striking disparities in the cost-to-performance ratio. High-end proprietary models like gpt-5.2 incur the highest operational costs (\$9.27 per complete run), primarily due to their pricing per token. In contrast, the mid-sized open-source models demonstrate remarkable efficiency. For instance, Qwen3-30B-A3B-Instruct completes the benchmark for just \$0.64, which is approximately 14x cheaper than gpt-5.2 while retaining competitive accuracy. This efficiency is largely driven by its MoE architecture, which activates only 3B parameters per token, drastically reducing computational overhead compared to dense counterparts like Llama-3.1-70B (\$5.22). This analysis suggests that for large-scale deployment, MoE architectures offer the optimal balance of reasoning capability and economic sustainability.

\begin{table}[t]
    \centering
    \small
    \renewcommand{\arraystretch}{1.2}
    \resizebox{0.49\textwidth}{!}{%
    \begin{tabular}{l| c c c}
        \toprule
        \textbf{Model Name}  & \textbf{Tokens (M)} & \textbf{Time (h)} & \textbf{Cost (\$)} \\
        \midrule
        \textit{Qwen Series} &&& \\
        Qwen3-235B-A22B-Instruct-2507 & 4.86 & 8.20 & 3.65 \\
        Qwen3-30B-A3B-Instruct-2507 & 4.86 & 4.09 & 0.64 \\
        Qwen3-30B-A3B & 5.09 & 2.87 & 0.71 \\
        Qwen3-30B-A3B-Base & 4.72 & 1.86 & 0.60 \\
        Qwen3-32B & 4.87 & 5.38 & 2.72 \\
        Qwen3-8B & 4.88 & 2.37 & 0.73 \\
        Qwen3-8B-Base & 4.73 & 2.36 & 0.67 \\
        Qwen3-4B-Instruct-2507 & 4.96 & 3.13 & 0.42 \\
        \midrule
        \textit{DeepSeek} &&& \\
        DeepSeek-V3.1 & 4.35 & 3.53 & 5.3 \\
        \midrule
        \textit{Meta Llama} &&& \\
        Llama-3.3-70B-Instruct & 4.54 & 2.04 & 4.77 \\
        Llama-3.1-70B & 4.36 & 3.02 & 5.22 \\
        Llama-3.1-8B & 4.40 & 2.07 & 0.67 \\
        Llama-3.1-8B-Instruct & 4.42 & 2.10 & 0.63 \\
        Llama-3.2-3B & 4.37 & 1.71 & 0.30 \\
        Llama-3.2-1B & 4.61 & 1.74 & 0.14 \\
        \midrule
        \textit{Moonshot AI} &&& \\
        Kimi-K2-Thinking & 4.71 & 14.88 & 5.43 \\
        Kimi-K2-Thinking$^\star$ & 7.62 & 60.92 & 12.53 \\
        \midrule
        \textit{OpenAI (OSS)} &&& \\
        gpt-oss-120b & 4.06 & 4.46 & 1.03 \\
        gpt-oss-120b$^\star$ & 5.50 & 7.00 & 1.28 \\
        gpt-oss-20b & 4.92 & 2.56 & 0.69 \\
        gpt-oss-20b$^\star$ & 5.84 & 5.71 & 0.97 \\
        \midrule
        \textit{OpenAI (GPT Series)} &&& \\
        gpt-5.2-2025-12-11 & 2.67 & 2.41 & 9.27 \\
        gpt-5.1-2025-11-13 & 2.66 & 1.93 & 6.44 \\
        gpt-4o-2024-08-06 & 2.30 & 1.37 & 2.28 \\
        gpt-4o-mini-2024-07-18 & 2.48 & 1.69 & 0.79 \\
        \bottomrule
    \end{tabular}
    }
    \caption{Resource consumption analysis for each model on the benchmark dataset. Metrics include total tokens processed, total wall-clock time in hours, and API cost in USD. Results averaged from three runs ($T-3$, $T$, $T+3$).}
    \label{tab:cost_analysis}
\end{table}

\section{SSA Framework Analysis}
\label{sec:ssa}

\subsection{SSA Framework and Entity Shift Implementation Details}
\label{sec:ssa_implementation}

While the primary utility of the SSA framework lies in the long-term monitoring of historical benchmarks, we conducted a rigorous simulation experiment to validate its immediate efficacy in detecting label-level contamination. To ensure a robust experimental setup free from dependency on proprietary API models, we utilized the high-performing open-source model Qwen3-235B-A22B-Instruct-2507 to execute the entity shift operations, prompt used in Figure \ref{fig:entity_shift}. The quality of this synthetic data generation was strictly audited (consistent with Appendix \ref{sec:generation}), and revealed that 98.5\% of the generated entity-shifted samples were semantically consistent and required no further adjustment.

\begin{figure*}[h]
    \centering
    \begin{promptbox}[title={Entity Shift Prompt}]
    \textbf{SYSTEM\_PROMPT} = """You are a precise data processor for research. Your task is to perform entity shifting - replacing real-world named entities with fictional alternatives while preserving all semantic meaning.

    You must:
    
    1. Replace ALL named entities (people, organizations, cities, locations, nationalities, races) with fictional alternatives, ensure these names are not widely recognized
    
    2. Use simple, uncommon but plausible names (e.g., "Trump" → "Korwin", "United States" → "Northland", "BBC" → "GBN")
    
    3. Keep entity types consistent (person→person, country→country, organization→organization)
    
    4. Preserve ALL other information (dates, numbers, events, relationships, facts)
    
    5. Ensure the SAME entity always maps to the SAME fictional name across ALL fields
    
    6. Output valid JSON only, no markdown code blocks, no explanations"""

    \textbf{ENTITY\_SHIFT\_PROMPT} = """Perform entity shifting on the following news data. Replace all named entities with fictional alternatives while preserving semantic meaning.
    
    **Original Event Title**: \{event\_title\}
    
    **Original Context**: \{context\}
    
    **Original Claim**: \{claim\}
    
    **Original Evidence Titles** (shift ONLY these titles, keep source names as-is):
    
    \{evidence\_titles\}
    
    **Requirements**:
    
    1. Identify ALL named entities: PERSON names, ORGANIZATION names, LOCATION names, NATIONALITY terms
    
    2. Replace each with a LESS RECOGNIZABLE fictional alternative
    
    3. Use simple, uncommon but plausible names (e.g., "Prince Andrew" → "Lord Harwick", "Buckingham Palace" → "Thornfield Manor", "United Kingdom" → "Alberia")
    
    4. Keep entity types consistent (person→person, palace→palace, country→country)
    
    5. PRESERVE all roles, positions, relationships, dates, numbers, and facts - ONLY change entity names
    
    6. Ensure CONSISTENT mapping: the SAME original entity ALWAYS maps to the SAME fictional name in ALL fields
    
    7. Do NOT change news source names (BBC, Reuters, etc.) - these are metadata, not content entities
    
    8. Evidence titles should use the SAME entity mapping as claim, context, and event\_title
    
    **Output Format** (JSON only, NO markdown, NO code blocks, NO explanations):
    
    \{\{
    
      "entity\_mapping": \{\{
      
        "Original Entity 1": "Fictional Name 1",
        
        "Original Entity 2": "Fictional Name 2"
        
      \}\},
      
      "event\_title\_shifted": "shifted event title here",
      
      "context\_shifted": "shifted context here",
      
      "claim\_shifted": "shifted claim here",
      
      "evidence\_titles\_shifted": [
      
        "shifted evidence title 1",
        
        "shifted evidence title 2"
        
      ]
      
    \}\}
    
    Output JSON only:"""
    \end{promptbox}
    
    \caption{Prompt for Entity Shift Processing}
    \label{fig:entity_shift}
\end{figure*}

\begin{table}[h]
\centering
\begin{tabular}{lc}
\toprule
\textbf{Hyperparameter} & \textbf{Value} \\
\midrule
Fine-tuning Method & LoRA \\
LoRA Rank ($r$) & 8 \\
Dropout & 0.05 \\
Learning Rate & $1 \times 10^{-4}$ \\
Optimizer & Adam \\
Batch Size & 4 \\
Number of Epochs & 3 \\
Loss Function & Cross-Entropy \\
\bottomrule
\end{tabular}
\caption{Fine-tuning hyperparameters for BDC simulation experiments.}
\label{tab:ssa_finetune_params}
\end{table}

\begin{table}[h]
    \centering
    \small
    \renewcommand{\arraystretch}{1.2}
    \resizebox{0.49\textwidth}{!}{%
    \begin{tabular}{l| c c c}
        \toprule
        \textbf{Model Name}  & \textbf{Tokens (M)} & \textbf{Time (h)} & \textbf{Cost (\$)} \\
        \midrule
        Qwen3-30B-A3B-Instruct-2507 & 24.12 & 3.69 & 5.18 \\
        Qwen3-4B-Instruct-2507 & 24.5 & 2.32 & 3.24 \\
        Llama-3.1-8B-Instruct & 22.19 & 2.32 & 5.27 \\
        \bottomrule
    \end{tabular}
    }
    \caption{Resource consumption for the contamination simulation experiments.}
    \label{tab:cost_ssa}
\end{table}

To simulate contamination, we constructed a poisoned training dataset derived from the LiveFact $T+3$ data split (Classification Mode). We mixed this specific benchmark data (formatted as instruction-response pairs) with a randomly selected subset of the Alpaca instruction fine-tuning dataset \citep{eohan2023stanford} (contaminants account for 30\% of the total). This creates a scenario where the model has "seen" the exact answers to the test set during training—a direct emulation of real-world BDC. We then fine-tuned three representative models (Qwen3-30B-A3B-Instruct-2507, Qwen3-4B-Instruct-2507, and Llama-3.1-8B-Instruct) on this contaminated corpus using Low-Rank Adaptation (LoRA). The specific hyperparameters used for this process are detailed in Table \ref{tab:ssa_finetune_params}.

We subsequently evaluated both the original "clean" models and their contaminated counterparts (marked with $^\star$) on two versions of the benchmark: the Original LiveFact and the Entity Shift LiveFact. The resource consumption for this contamination simulation is provided in Table \ref{tab:cost_ssa}.

\subsection{SSA Implementation Results}

The results of the simulation, presented in Table \ref{tab:ssa_test_result}, provide compelling evidence for the efficacy of the SSA framework.

First, observe the performance on the Original LiveFact benchmark. The contaminated models ($^\star$) achieve near-perfect accuracy scores (e.g., 99.89\% for Qwen3-30B-A3B-Instruct-2507$^\star$ at $T+3$), a massive jump from the original model's 77.00\%. This confirms that the fine-tuning successfully injected the specific knowledge of the benchmark into the models' parameters; they are essentially "reciting" the memorized answers \citep{ovadia2024fine,lyu2024knowtuning,li2025scilitllm}.

However, the critical insight comes from comparing performance on the Entity Shift LiveFact. When tested on the version where entity names are swapped (e.g., "Trump" $\rightarrow$ "Wannetta"), the performance of the contaminated models drops precipitously. For Qwen3-30B-A3B-Instruct-2507$^\star$, accuracy falls from 99.89\% (Original) to 84.77\% (Shifted). While still high due to general instruction tuning, this gap ($\Delta$) exposes the model's reliance on specific entity tokens rather than generalized reasoning.

This effect is quantified by the SSA Factor in Table \ref{tab:ssa_calculation}. The original, clean models maintain a negligible SSA Factor (e.g., 0.08 for Qwen3-30B-A3B-Instruct-2507), indicating robust, entity-agnostic reasoning. In stark contrast, the contaminated models exhibit massive spikes in their SSA scores (e.g., 5.18 for Qwen3-30B-A3B-Instruct-2507$^\star$ and 6.67 for Qwen3-4B-Instruct-2507$^\star$). This drastic increase—driven by both the performance drop ($\Delta$) and the high OTR—serves as a clear, quantifiable signal of data contamination. These findings definitively validate that the SSA framework acts as a sensitive "litmus test" for label-level contamination, justifying its integral role in the long-term maintenance of the LiveFact benchmark.

\begin{table}[h]
    \centering
    \renewcommand{\arraystretch}{1.2}
    \resizebox{0.49\textwidth}{!}{%
    \begin{tabular}{l|ccc}
    \toprule
        \textbf{Model} & \textbf{$\Delta$} & \textbf{OTR} & \textbf{SSA} \\
        \midrule
         Qwen3-30B-A3B-Instruct-2507 & 4.90 & 1.53 & 0.08 \\
         Qwen3-30B-A3B-Instruct-2507$^\star$ & 15.24 & 33.99 & 5.18 \\
         \midrule
         Qwen3-4B-Instruct-2507 & 6.56 & 2.76 & 0.18 \\
         Qwen3-4B-Instruct-2507$^\star$ & 14.58 & 45.72 & 6.67 \\
         \midrule
         Llama-3.1-8B-Instruct & 1.46 & 5.31 & 0.08 \\
         Llama-3.1-8B-Instruct$^\star$ & 13.53 & 39.16 & 5.30 \\
         \bottomrule
    \end{tabular}
    }
    \caption{SSA Factor calculation showing the clear distinction between clean and contaminated ($^\star$) models. The calculation of $\Delta$ and OTR is based on the average of six runs performed under the selected Classification and Inference modes.}
    \label{tab:ssa_calculation}
\end{table}

\begin{table*}[t]
    \centering
    \renewcommand{\arraystretch}{1.2}
    \resizebox{1.00\textwidth}{!}{%
    \begin{tabular}{l|ccc|ccc|ccc|ccc|c}
    \toprule
        \multirow{2}{*}{\backslashbox[52mm]{\textbf{Model}}{\textbf{Eval. Mode}}} & \multicolumn{6}{c|}{\textbf{Classification}} & \multicolumn{6}{c|}{\textbf{Inference}} & \multirow{2}{*}{\textbf{Avg.}} \\
        \cmidrule(lr){2-4} \cmidrule(lr){5-7} \cmidrule(lr){8-10} \cmidrule(lr){11-13}
        & $\mathrm{Acc}_{-3}^{\textsc{cls}}$ & $\mathrm{Acc}_{0}^{\textsc{cls}}$ & $\mathrm{Acc}_{+3}^{\textsc{cls}}$ 
        & $\mathrm{F1}_{-3}^{\textsc{cls}}$ & $\mathrm{F1}_{0}^{\textsc{cls}}$ & $\mathrm{F1}_{+3}^{\textsc{cls}}$
        & $\mathrm{Acc}_{-3}^{\textsc{inf}}$ & $\mathrm{Acc}_{0}^{\textsc{inf}}$ & $\mathrm{Acc}_{+3}^{\textsc{inf}}$
        & $\mathrm{F1}_{-3}^{\textsc{inf}}$ & $\mathrm{F1}_{0}^{\textsc{inf}}$ & $\mathrm{F1}_{+3}^{\textsc{inf}}$ \\
        \midrule
        \multicolumn{14}{l}{\textbf{Original LiveFact}} \\
        \midrule
        Qwen3-30B-A3B-Instruct-2507 & 55.24 & 75.05 & 77.00 & 52.39 & 74.87 & 76.61 & 64.55 & 74.64 & 76.78 & 55.37 & 74.66 & 76.40 & 69.46\\
        Qwen3-30B-A3B-Instruct-2507$^\star$ & 84.59 & 99.57 & 99.89 & 83.28 & 99.57 & 99.89 & 59.24 & 97.13 & 99.52 & 52.42 & 97.16 & 99.52 & 89.31 \\
        \midrule
        Qwen3-4B-Instruct-2507 & 46.29 & 62.43 & 63.39 & 42.63 & 60.70 & 61.58 & 83.08 & 62.45 & 63.39 & 64.45 & 61.15 & 61.65 & 61.10 \\
        Qwen3-4B-Instruct-2507$^\star$ & 77.98 & 98.98 & 99.50 & 74.50 & 98.97 & 99.50 & 53.05 & 96.63 & 99.16 & 52.16 & 96.66 & 99.16 & 87.19 \\
        \midrule
        Llama-3.1-8B-Instruct & 40.69 & 50.05 & 50.20 & 36.53 & 45.67 & 45.11 & 78.69 & 50.16 & 50.09 & 60.26 & 46.89 & 45.14 & 49.96 \\
        Llama-3.1-8B-Instruct$^\star$ & 88.84 & 96.02 & 96.49 & 84.37 & 96.02 & 96.49 & 46.02 & 93.62 & 96.17 & 42.14 & 93.62 & 96.17 & 85.50 \\
        \midrule
        \multicolumn{14}{l}{\textbf{Entity Shift LiveFact}} \\
        \midrule
        Qwen3-30B-A3B-Instruct-2507 & 46.93 & 69.85 & 71.40 & 43.17 & 68.82 & 70.16 & 81.63 & 69.08 & 71.31 & 63.35 & 68.89 & 70.08 & 64.56 \\
        Qwen3-30B-A3B-Instruct-2507$^\star$ & 69.60 & 84.31 & 84.77 & 66.39 & 82.61 & 84.74 & 50.02 & 76.25 & 86.34 & 41.16 & 76.35 & 86.25 & 74.07 \\
        \midrule
        Qwen3-4B-Instruct-2507 & 44.22 & 61.18 & 63.91 & 37.98 & 60.49 & 62.95 & 36.16 & 60.61 & 63.87 & 39.56 & 60.61 & 62.99 & 54.54 \\
        Qwen3-4B-Instruct-2507$^\star$ & 67.33 & 81.90 & 83.49 & 62.97 & 82.09 & 83.50 & 47.75 & 75.96 & 85.31 & 39.70 & 76.03 & 85.26 & 72.61 \\
        \midrule
        Llama-3.1-8B-Instruct & 39.89 & 48.27 & 48.54 & 34.87 & 43.16 & 42.96 & 79.67 & 48.38 & 48.47 & 60.32 & 44.42 & 43.05 & 48.50 \\
        Llama-3.1-8B-Instruct$^\star$ & 71.43 & 79.62 & 82.67 & 69.57 & 79.72 & 82.70 & 45.47 & 73.54 & 83.49 & 38.29 & 73.72 & 83.36 & 71.97 \\
    \bottomrule
    \end{tabular}
    }
    \caption{Performance comparison of original vs. contaminated ($^\star$) models on Original and Entity-Shifted LiveFact datasets.}
    \label{tab:ssa_test_result}
\end{table*}

\section{Supplementary Analysis}
\label{sec:case}

\subsection{Impact of Training Mode: Base, Instruct, and Hybrid}

To isolate the effect of training methodology on reasoning capabilities, we conducted an ablation study comparing different variants of the same model architectures. We examined Base models (pre-trained only), Instruct models (SFT + RLHF), and Hybrid models across the Qwen3 and Llama-3.1 families. The performance results are detailed in Table \ref{tab:case}.

\begin{table*}[]
    \centering
    \renewcommand{\arraystretch}{1.2}
    \resizebox{1.00\textwidth}{!}{%
    \begin{tabular}{l|ccc|ccc|ccc|ccc|c}
    \toprule
        \multirow{2}{*}{\backslashbox[52mm]{\textbf{Model}}{\textbf{Eval. Mode}}} & \multicolumn{6}{c|}{\textbf{Classification}} & \multicolumn{6}{c|}{\textbf{Inference}} & \multirow{2}{*}{\textbf{Avg.}} \\
        \cmidrule(lr){2-4} \cmidrule(lr){5-7} \cmidrule(lr){8-10} \cmidrule(lr){11-13}
        & $\mathrm{Acc}_{-3}^{\textsc{cls}}$ & $\mathrm{Acc}_{0}^{\textsc{cls}}$ & $\mathrm{Acc}_{+3}^{\textsc{cls}}$ 
        & $\mathrm{F1}_{-3}^{\textsc{cls}}$ & $\mathrm{F1}_{0}^{\textsc{cls}}$ & $\mathrm{F1}_{+3}^{\textsc{cls}}$
        & $\mathrm{Acc}_{-3}^{\textsc{inf}}$ & $\mathrm{Acc}_{0}^{\textsc{inf}}$ & $\mathrm{Acc}_{+3}^{\textsc{inf}}$
        & $\mathrm{F1}_{-3}^{\textsc{inf}}$ & $\mathrm{F1}_{0}^{\textsc{inf}}$ & $\mathrm{F1}_{+3}^{\textsc{inf}}$ \\
        \midrule
        Qwen3-30B-A3B-Instruct-2507 & 55.24 & 75.05 & 77.00 & 52.39 & 74.87 & 76.61 & 64.55 & 74.64 & 76.78 & 55.37 & 74.66 & 76.40 & 69.46\\
        Qwen3-30B-A3B & 3.42 & 0.14 & 0.07 & 6.05 & 0.27 & 0.14 & 2.69 & 0.11 & 0.07 & 2.27 & 0.23 & 0.14 & 1.30 \\
        Qwen3-30B-A3B-Base & 46.47 & 55.17 & 54.92 & 43.26 & 51.67 & 51.33 & 76.34 & 54.58 & 54.78 & 55.80 & 51.57 & 51.25 & 53.93 \\
        \midrule
        Qwen3-8B & 50.84 & 68.88 & 69.83 & 49.12 & 67.79 & 68.44 & 60.29 & 68.67 & 69.65 & 53.76 & 67.82 & 68.30 & 63.62 \\
        Qwen3-8B-Base & 46.29 & 62.43 & 63.39 & 42.63 & 60.70 & 61.58 & 83.08 & 62.45 & 63.39 & 64.45 & 61.15 & 61.65 & 61.10 \\
        \midrule
        Llama-3.1-8B-Instruct & 40.69 & 50.05 & 50.20 & 36.53 & 45.67 & 45.11 & 78.69 & 50.16 & 50.09 & 60.26 & 46.89 & 45.14 & 49.96 \\
        Llama-3.1-8B & 11.89 & 3.62 & 3.12 & 12.39 & 5.72 & 4.99 & 1.25 & 3.30 & 3.12 & 1.90 & 5.39 & 5.02 & 5.14 \\
    \bottomrule
    \end{tabular}
    }
    \caption{Ablation study on training methodologies across Qwen and Llama architectures. Hybrid and Instruct models generally outperform Base models, though strong base reasoning is observed in Qwen3-8B.}
    \label{tab:case}
\end{table*}

The analysis reveals a stark "Alignment Tax" on performance for models that lack instruction tuning. The Llama-3.1-8B base model collapses to near-random performance (Avg: 5.14\%), primarily because it fails to adhere to the strict output formatting required by the benchmark (\verb|[[LABEL]]|). Without SFT, the model often continues generating context or hallucinates unrelated text instead of providing a classification. In contrast, the Instruct version reaches 49.96\%, demonstrating that the knowledge exists within the parameters but requires alignment to be accessible. A similar, though less catastrophic, trend is seen in Qwen3-30B-A3B, where the specific hybrid variant underperformed significantly (1.30\%) likely due to a mismatch in prompt template expectations compared to the robust Instruct-2507 version (69.46\%).

However, an interesting anomaly emerges with Qwen3-8B-Base. Unlike its Llama counterpart, this base model exhibits surprisingly strong reasoning capabilities, achieving an average score of 61.10\%, which is competitive with the Qwen3-8B hybrid model (63.62\%). Notably, in the  $\delta=-3$ Inference Mode, the Base model actually outperforms the hybrid version (83.08\% vs 60.29\%). This suggests that the Qwen3 base pre-training corpus may contain richer, more structured reasoning patterns that allow it to naturally infer the task even without explicit instruction tuning. This finding challenges the assumption that base models are universally unsuitable for zero-shot reasoning tasks, provided the underlying pre-training is sufficiently robust.

\begin{figure*}[]
    \centering
    \begin{tcolorbox}[
        colback=white, 
        colframe=gray!50, 
        title=\textbf{Case Study: Kimi-K2-Thinking vs. Qwen3-235B-A22B-Instruct-2507},
        fontupper=\small
    ]
        \evaluator{You are a fact-checking expert. Classify news claims based on evidence.....Claim: One student escaped as authorities intensified their search for the 24 other girls abducted by gunmen from Government Girls College in Maga, Kebbi state, northwestern Nigeria.....\textbf{Classify this claim. Start with [[REAL]], [[FAKE]], or [[AMBIGUOUS]], do not think too much.}}
        
        \kimi{<think> The user wants me to classify a news claim based on the provided evidence. The claim is:\textbackslash n\textbackslash n\textbackslash "One student escaped as authorities intensified their search for the 24 other girls abducted by gunmen from Government Girls College in Maga, Kebbi state, northwestern Nigeria.\textbackslash "\textbackslash n\textbackslash nLet me analyze the evidence:\textbackslash n\textbackslash nKey elements of the claim:\textbackslash n1. Location: Government Girls College in Maga, Kebbi state, northwestern Nigeria\textbackslash n2. Event: Gunmen abducted girls\textbackslash n3. Number: 24 other girls (implying 25 total, with 1 escaped)\textbackslash n4. One student escaped\textbackslash n5}
        
        \qwen{[[REAL]] Multiple credible sources confirm that 25 girls were abducted from Government Girls College Maga in Kebbi State, and at least one student escaped, with authorities intensifying searches for the others. Sources like BBC, CNN, NPR, and Al Jazeera consistently report these details.}
    \end{tcolorbox}
    \caption{Comparison of raw model outputs. Kimi-K2 fails to produce the required classification label, getting stuck in an internal "thought chain," whereas Qwen3 immediately delivers the structured verdict.}
    \label{fig:case_study}
\end{figure*}

\begin{table*}[t]
    \centering
    \resizebox{1.00\textwidth}{!}{%
    \begin{tabular}{l|ccc|ccc|ccc|ccc|c}
    \toprule
        \multirow{2}{*}{\backslashbox[55mm]{\textbf{Model}}{\textbf{Eval. Mode}}} & \multicolumn{6}{c|}{\textbf{Classification}} & \multicolumn{6}{c|}{\textbf{Inference}} & \multirow{2}{*}{\textbf{Avg.}} \\
        \cmidrule(lr){2-4} \cmidrule(lr){5-7} \cmidrule(lr){8-10} \cmidrule(lr){11-13}
        & $\mathrm{Acc}_{-3}^{\textsc{cls}}$ & $\mathrm{Acc}_{0}^{\textsc{cls}}$ & $\mathrm{Acc}_{+3}^{\textsc{cls}}$ 
        & $\mathrm{F1}_{-3}^{\textsc{cls}}$ & $\mathrm{F1}_{0}^{\textsc{cls}}$ & $\mathrm{F1}_{+3}^{\textsc{cls}}$
        & $\mathrm{Acc}_{-3}^{\textsc{inf}}$ & $\mathrm{Acc}_{0}^{\textsc{inf}}$ & $\mathrm{Acc}_{+3}^{\textsc{inf}}$
        & $\mathrm{F1}_{-3}^{\textsc{inf}}$ & $\mathrm{F1}_{0}^{\textsc{inf}}$ & $\mathrm{F1}_{+3}^{\textsc{inf}}$ \\
        \midrule
        Kimi-K2-Thinking & 6.72 & 0.20 & 0.23 & 10.51 & 0.40 & 0.45 & 0.64 & 0.16 & 0.23 & 0.82 & 0.33 & 0.45 & 1.76\\
        Kimi-K2-Thinking$^\star$ & 45.97 & 57.15 & 54.21 & 45.34 & 61.71 & 59.22 & 58.13 & 56.56 & 54.08 & 42.07 & 61.57 & 59.15 & 54.60 \\
        \midrule
        gpt-oss-120b & 27.19 & 20.13 & 19.54 & 33.87 & 30.98 & 29.85 & 26.48 & 19.67 & 19.51 & 25.04 & 31.00 & 29.90 & 26.10 \\
        gpt-oss-120b$^\star$ & 55.83 & 79.94 & 81.81 & 52.82 & 79.79 & 81.49 & 62.23 & 78.89 & 81.60 & 50.83 & 78.97 & 81.31 & 72.13 \\
        \midrule
        gpt-oss-20b & 18.76 & 23.63 & 23.34 & 26.02 & 33.39 & 32.80 & 16.44 & 23.34 & 23.32 & 22.07 & 33.86 & 32.88 & 25.82 \\
        gpt-oss-20b$^\star$ & 47.84 & 65.55 & 67.42 & 47.16 & 61.33 & 62.32 & 41.46 & 64.32 & 67.28 & 36.53 & 61.07 & 62.34 & 57.05 \\
    \bottomrule
    \end{tabular}
    }
    \caption{Performance of Reasoning Models evaluated with extended token limits (1024 tokens). The asterisk ($^\star$) denotes the non-standard evaluation setting required to accommodate verbose Chain-of-Thought outputs.}
    \label{tab:reasoning_comparison}
\end{table*}

\subsection{The Cost of Reasoning: Verbosity vs. Compliance in Thinking Models}

A critical finding from our main experiments is the distinct behavior of models heavily optimized for "Reasoning" or "Chain-of-Thought" (CoT), specifically Kimi-K2-Thinking and the GPT-OSS series. In our initial screening using a standard 128-token limit (Table \ref{tab:reasoning_comparison}), these models appeared to fail catastrophically (e.g., Kimi-K2-Thinking scored \textasciitilde1.76\%). However, upon qualitative inspection (Figure \ref{fig:case_study}), it became evident that the models were not hallucinating but rather engaging in extensive internal deliberation.

As shown in Table \ref{tab:reasoning_comparison}, re-evaluating these models with an extended 1024-token window unlocked their true potential. gpt-oss-120b, for instance, jumped to an average accuracy of 72.13\%, rivaling the best-performing Qwen models. This reveals a fundamental trade-off in current model design: "Thinking Models" prioritize exhaustive reasoning paths over strict formatting constraints. Unlike standard Instruction-Tuned models (e.g., Qwen3-Instruct) that can "snap" to a label immediately, Reasoning models treat the prompt as a starting point for a dialectical process. While this leads to high accuracy when resources are unconstrained (Table \ref{tab:cost_analysis}), it poses significant challenges for low-latency, automated benchmarking environments where conciseness is often a proxy for adherence.

\subsection{Deep Dive: Confusion Matrix Analysis}
\label{sec:confusion}

To provide a granular view of model behavior, we analyzed the confusion matrices across all model groups (Figures \ref{fig:confusion_matrix}).

Across all valid models (e.g., Qwen3-Instruct, GPT-4o), we observe a massive shift in prediction behavior at $T-3$. In Classification Mode, the matrices show a "spray" of predictions; models struggle to differentiate classes, often defaulting to the majority class of their training data. However, in Inference Mode, the confusion matrix tightens significantly around the "Ambiguous" class. For example, Qwen3-32B correctly concentrates nearly all its mass on "Ambiguous" at $T-3$ (Inference), whereas in Classification mode, it erroneously predicts "Real" or "Fake" with low confidence. This confirms that the models detect the ambiguity but are forced to hallucinate in standard settings.

As time progresses to $T$ and $T+3$, the confusion matrices for high-performing models (GPT-5.2, Qwen3-235B) show a distinct "diagonalization." The noise from the "Ambiguous" class dissipates, and predictions sharpen into correct "Real" vs. "Fake" classifications. This temporal evolution proves that the benchmark successfully captures the arrival of new information.

The confusion matrices for base models (e.g., Llama-3.1-8B) reveal a pathological failure mode. Rather than a distributed error pattern, these models often exhibit a single-column collapse, predicting one class (e.g., "Real") for 100\% of queries regardless of input. This is not due to reasoning but due to a failure to condition on the evidence; the model likely defaults to the most common token in its pre-training distribution for the given prompt context. This visualizes exactly why instruction tuning is non-negotiable for reliable automated fact-checking.

\subsection{Error Analysis on Frontier Models}
\label{sec:error_analysis}

While Section \ref{sec:exp} provides an aggregate view of model performance across the LiveFact benchmark, it is crucial to diagnose the specific failure modes of SOTA frontier models. Interestingly, our analysis reveals that the primary source of errors for highly capable models is fundamentally different from that of earlier generations. Specifically, we observe a phenomenon we term \textit{Over-Safety} (or extreme \textit{Epistemic Humility}), where models prioritize cautious refusals over definitive verdicts.

To quantify this, we conducted a deep-dive error breakdown on \textbf{GPT-5.2} under the Inference Mode at time slice $T$. We isolated all claims where the Ground Truth is strictly "Fake" (i.e., the provided evidence slice contains explicit contradictions to the claim). Out of 1,390 such instances in the November 2025 data, GPT-5.2 correctly predicted "Fake" in 83.24\% of cases. For the remaining erroneous predictions, the breakdown is as follows:
\begin{itemize}
    \item \textbf{Hallucinations (Predicted "Real"):} Only \textbf{0.22\%} (3 instances).
    \item \textbf{False Negatives via Ambiguity (Predicted "Ambiguous"):} \textbf{10.58\%} (147 instances).
    \item \textbf{Safety Refusals (Null Output):} \textbf{5.97\%} (83 instances).
\end{itemize}

This striking contrast—a 16.55\% false-negative rate driven by caution versus a near-zero (0.22\%) hallucination rate—demonstrates that the challenge in AI-driven verification has shifted. For frontier models, the "Reasoning Gap" is less about the incapability to discern the truth, and more about the strict calibration of alignment filters and refusal thresholds when processing sensitive geopolitical news.

To further illustrate this behavioral shift, we present two representative qualitative failure cases in Table \ref{tab:error_cases}, highlighting the nuances of \textit{Over-Caution} and \textit{Safety Refusal}.

\begin{table*}[]
\centering
\small
\begin{tabular}{p{0.15\linewidth} | p{0.8\linewidth}}
\toprule
\multicolumn{2}{c}{\textbf{Failure Type 1: Over-Caution (Ground Truth: \textsc{Fake} $\rightarrow$ Prediction: \textsc{Ambiguous})}} \\
    \midrule
    \textbf{Claim ID} & EV20251101-10-4 (Inference Mode with $E^{(0)}$) \\
    \textbf{Context} & A specific allegation regarding a senior US diplomat privately demanding Israel to free exactly 200 Palestinian security prisoners during White House talks. \\
    \textbf{Ground Truth} & \textsc{Fake} (The evidence explicitly confirms no such specific demand or exact number was discussed, contradicting the claim). \\
    \textbf{GPT-5.2 Output} & \texttt{[[AMBIGUOUS]] The cited coverage focuses on negotiations and hostage/body returns during a Gaza truce, but it does not clearly and consistently corroborate the specific allegation that a `senior US diplomat' privately demanded Israel free exactly 200 Palestinian security prisoners in White House talks...} \\
    \textbf{Analysis} & The model successfully identifies that the broad topic (Gaza truce) is present in the evidence but the specific fabricated details are missing. However, instead of confidently deducing that the claim is \textsc{Fake} due to these explicit contradictions with the official minutes provided in the slice, it conservatively retreats to \textsc{Ambiguous}, demanding an impossible level of direct negation. \\
    \midrule
\multicolumn{2}{c}{\textbf{Failure Type 2: Safety Refusal (Ground Truth: \textsc{Fake} $\rightarrow$ Prediction: Null/Refusal)}} \\
    \midrule
    \textbf{Claim ID} & EV20251101-20-4 (Inference Mode with $E^{(0)}$) \\
    \textbf{Context} & A fabricated claim involving sensitive military actions and political figures in a conflict zone. \\
    \textbf{Ground Truth} & \textsc{Fake} \\
    \textbf{GPT-5.2 Output} & \texttt{""} (Empty string / System refusal) \\
    \textbf{Analysis} & Despite being provided with a golden evidence slice that disproves the claim, the model's internal safety guardrails are triggered by the sensitive keywords (e.g., specific conflict zones, names of political leaders). It halts the generation process entirely, leading to a missing prediction. \\
    \bottomrule
\end{tabular}
\caption{Representative failure cases for GPT-5.2 demonstrating Over-Safety and Epistemic Humility. Even when sufficient evidence is present to declare a claim "Fake", frontier models often default to "Ambiguous" or trigger safety refusals.}
\label{tab:error_cases}
\end{table*}

These findings offer a critical take-home message for the community: as LLMs become more aligned, static benchmarks that only measure binary accuracy ("Real" vs. "Fake") fail to capture the "alignment tax" imposed on verification tasks. LiveFact's inclusion of the "Ambiguous" class and dynamic slices effectively exposes this nuance, setting a more precise standard for future model evaluation.

\subsection{Longitudinal Evaluation: December 2025 Version}

To demonstrate the temporal consistency of model behaviors and validate the long-term sustainability of the LiveFact framework—a critical requirement for combating BDC—we conducted a longitudinal evaluation using the newly collected December 2025 data snapshot. We evaluated a representative subset of 8 hybrid/instruction-tuned open-source models from our main experiments.

As shown in Table \ref{tab:disctribution_dec}, the December 2025 dataset contains 4,222 claims, which are perfectly balanced across the "Real" (1,404), "Fake" (1,406), and "Ambiguous" (1,412) labels in the time-invariant Classification Mode. Crucially, the dynamic label distribution in Inference Mode exhibits the exact same "fog of war" pattern observed in the November 2025 data. At the $E^{(-3)}$ slice, the vast majority of claims (3,533 out of 4,222) shift to the "Ambiguous" class, accurately reflecting the real-world scarcity of preemptive evidence before an event unfolds. As time progresses to $E^{(0)}$ and $E^{(+3)}$, the distribution naturally aligns back with the ground-truth Classification labels as evidence becomes sufficient.

\begin{table}[h]
    \centering
    \resizebox{0.48\textwidth}{!}{%
    \begin{tabular}{l|ccccc}
    \toprule
         \textbf{Evidence Set} & \textbf{Real} & \textbf{Fake} & \textbf{Ambiguous} & \textbf{Total Claims} \\
        \midrule
        $E^{(-3)}$ & 343 & 346 & 3,533 & 4,222 \\
        $E^{(0)}$ & 1,374 & 1,376 & 1,472 & 4,222 \\
        $E^{(+3)}$ & 1,404 & 1,406 & 1,412 & 4,222 \\
    \bottomrule
    \end{tabular}
    }
    \caption{Dynamic Label Distribution for Inference Mode (December 2025)}
    \label{tab:disctribution_dec}
\end{table}

The evaluation results on the December 2025 version (Table \ref{tab:result_dec}) highly correlate with our findings from November 2025, confirming the temporal robustness of the benchmark. The Qwen3-235B-A22B-Instruct model consistently maintains its leading position across almost all metrics, achieving the highest overall average score (71.02\%), followed closely by the Qwen3-30B-A3B model.

\begin{table*}[h]
    \centering
    \resizebox{1.00\textwidth}{!}{%
    \begin{tabular}{l|ccc|ccc|ccc|ccc|c}
    \toprule
        \multirow{2}{*}{\backslashbox[55mm]{\textbf{Model}}{\textbf{Eval. Mode}}} & \multicolumn{6}{c|}{\textbf{Classification}} & \multicolumn{6}{c|}{\textbf{Inference}} & \multirow{2}{*}{\textbf{Avg.}} \\
        \cmidrule(lr){2-4} \cmidrule(lr){5-7} \cmidrule(lr){8-10} \cmidrule(lr){11-13}
        & $\mathrm{Acc}_{-3}^{\textsc{cls}}$ & $\mathrm{Acc}_{0}^{\textsc{cls}}$ & $\mathrm{Acc}_{+3}^{\textsc{cls}}$ 
        & $\mathrm{F1}_{-3}^{\textsc{cls}}$ & $\mathrm{F1}_{0}^{\textsc{cls}}$ & $\mathrm{F1}_{+3}^{\textsc{cls}}$
        & $\mathrm{Acc}_{-3}^{\textsc{inf}}$ & $\mathrm{Acc}_{0}^{\textsc{inf}}$ & $\mathrm{Acc}_{+3}^{\textsc{inf}}$
        & $\mathrm{F1}_{-3}^{\textsc{inf}}$ & $\mathrm{F1}_{0}^{\textsc{inf}}$ & $\mathrm{F1}_{+3}^{\textsc{inf}}$ \\
        \midrule
        Qwen3-235B-A22B-Instruct-2507 & \underline{52.18} & \textbf{78.83} & \textbf{80.53} & 47.97 & \textbf{78.93} & \textbf{80.57} & 62.65 & \textbf{78.56} & \textbf{80.53} & 52.25 & \textbf{78.72} & \textbf{80.57} & \textbf{71.02} \\
        Qwen3-30B-A3B-Instruct-2507 & \textbf{55.38} & \underline{73.80} & \underline{75.49} & \textbf{52.35} & \underline{73.35} & \underline{74.90} & 63.60 & \underline{73.26} & \underline{75.49} & 54.35 & \underline{72.92} & \underline{74.90} & \underline{68.32} \\
        Qwen3-32B & 45.81 & 70.72 & 71.06 & 40.57 & 69.14 & 69.14 & \underline{81.08} & 70.91 & 71.06 & \underline{60.09} & 69.28 & 69.14 & 61.85 \\
        Qwen3-8B & 50.85 & 67.22 & 67.10 & \underline{48.65} & 65.36 & 64.72 & 54.95 & 66.94 & 67.10 & 50.88 & 65.30 & 64.72 & 61.15 \\
        Qwen3-4B-Instruct-2507 & 45.31 & 63.31 & 66.49 & 37.59 & 62.52 & 65.61 & 35.03 & 62.96 & 66.49 & 36.30 & 62.50 & 65.61 & 55.81 \\
        \midrule
        Llama-3.3-70B-Instruct & 44.91 & 68.31 & 68.95 & 38.59 & 62.23 & 62.34 & 22.55 & 67.10 & 68.95 & 31.97 & 61.38 & 62.34 & 54.97 \\
        Llama-3.1-8B-Instruct & 41.76 & 49.53 & 49.55 & 37.88 & 44.96 & 44.68 & 79.06 & 49.74 & 49.55 & \textbf{61.26} & 45.74 & 44.68 & 49.87 \\
        \midrule
        DeepSeek-V3.1 & 43.46 & 72.64 & 74.49 & 36.01 & 70.81 & 72.67 & \textbf{82.76} & 72.57 & 74.49 & 57.83 & 70.68 & 72.67 & 66.76 \\
    \bottomrule
    \end{tabular}
    }
    \caption{Performance comparison of 8 representative LLMs on LiveFact using the December 2025 longitudinal data. The best result in each column is \textbf{bolded}, and the second-best result is \underline{underlined}.}
    \label{tab:result_dec}
\end{table*}

Furthermore, the December results strongly corroborate our observations regarding the \textit{Reasoning Gap} and \textit{Epistemic Humility}. For instance, models like DeepSeek-V3.1 and Qwen3-32B exhibit relatively low accuracy in the Classification Mode at $E^{(-3)}$ (43.46\% and 45.81\%, respectively), which primarily measures prior bias or forced guessing. However, when evaluated in Inference Mode at the same $E^{(-3)}$ slice—where they are allowed to assess evidence sufficiency—their accuracy surges to 82.76\% and 81.08\%. This massive leap demonstrates a consistent capacity for epistemic humility across different time periods, as these models successfully recognize missing evidence and output the "Ambiguous" label. 

Conversely, Llama-3.3-70B-Instruct struggles significantly at Inference $E^{(-3)}$ (22.55\%), indicating a persistent behavioral flaw where it over-commits to binary verification even when evidence is objectively insufficient.

In summary, the successful deployment of the December 2025 evaluation not only proves the logistical sustainability of LiveFact's monthly update pipeline but also verifies that the model capabilities and failure modes identified in our primary November study are consistent and generalizable over time.

\begin{figure*}
    \centering
    \resizebox{1\textwidth}{!}{%
    \includegraphics[width=1\linewidth]{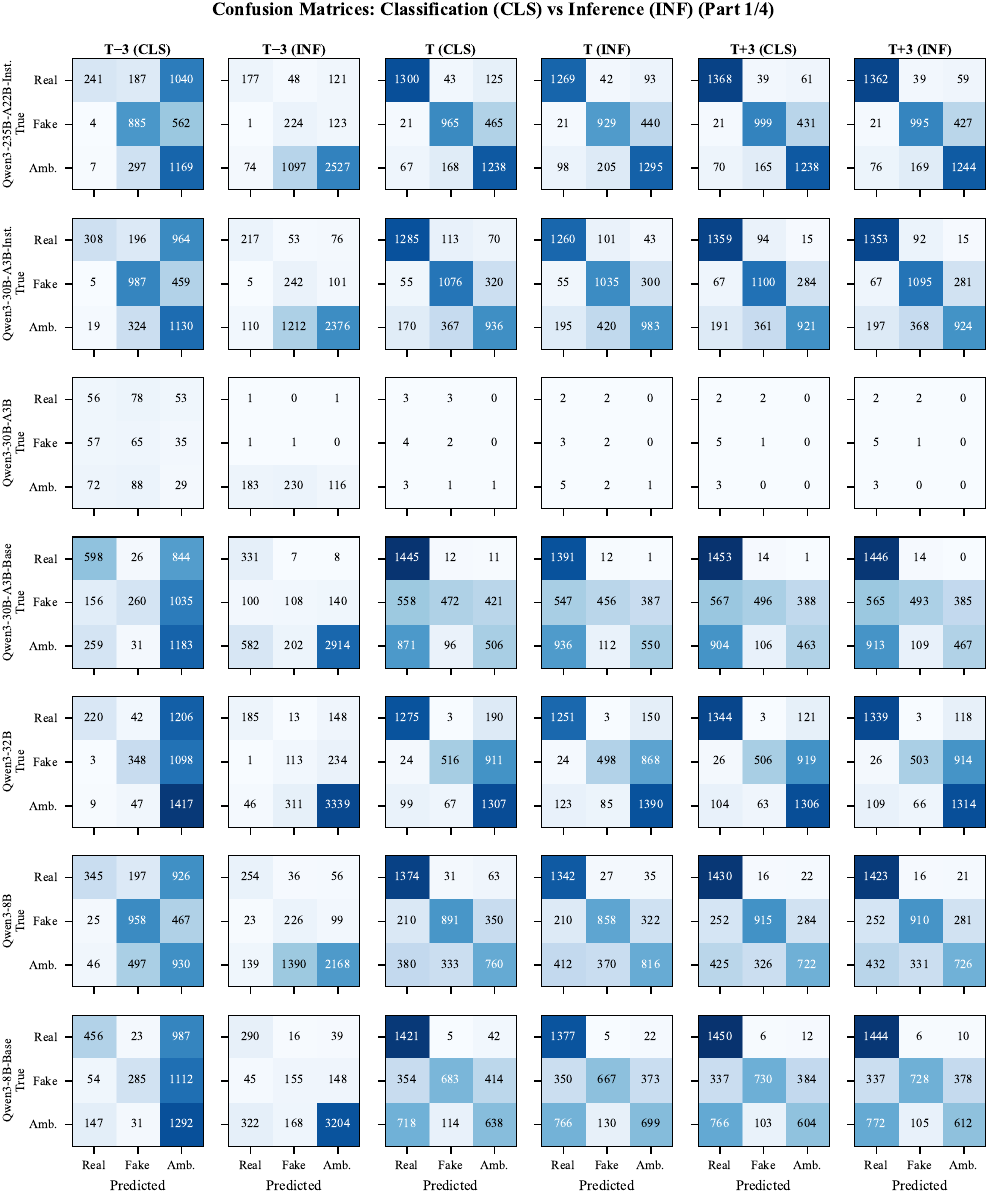}
    }
    \caption{Confusion Matrices (Part 1/4) - Qwen Series. Note the sharp contrast in diagonal clarity between Instruct models and Base models.}
    \label{fig:confusion_matrix}
\end{figure*}

\begin{figure*}
    \ContinuedFloat
    \captionsetup{labelformat=continued}
    \centering
    \resizebox{1\textwidth}{!}{%
    \includegraphics[width=1\linewidth]{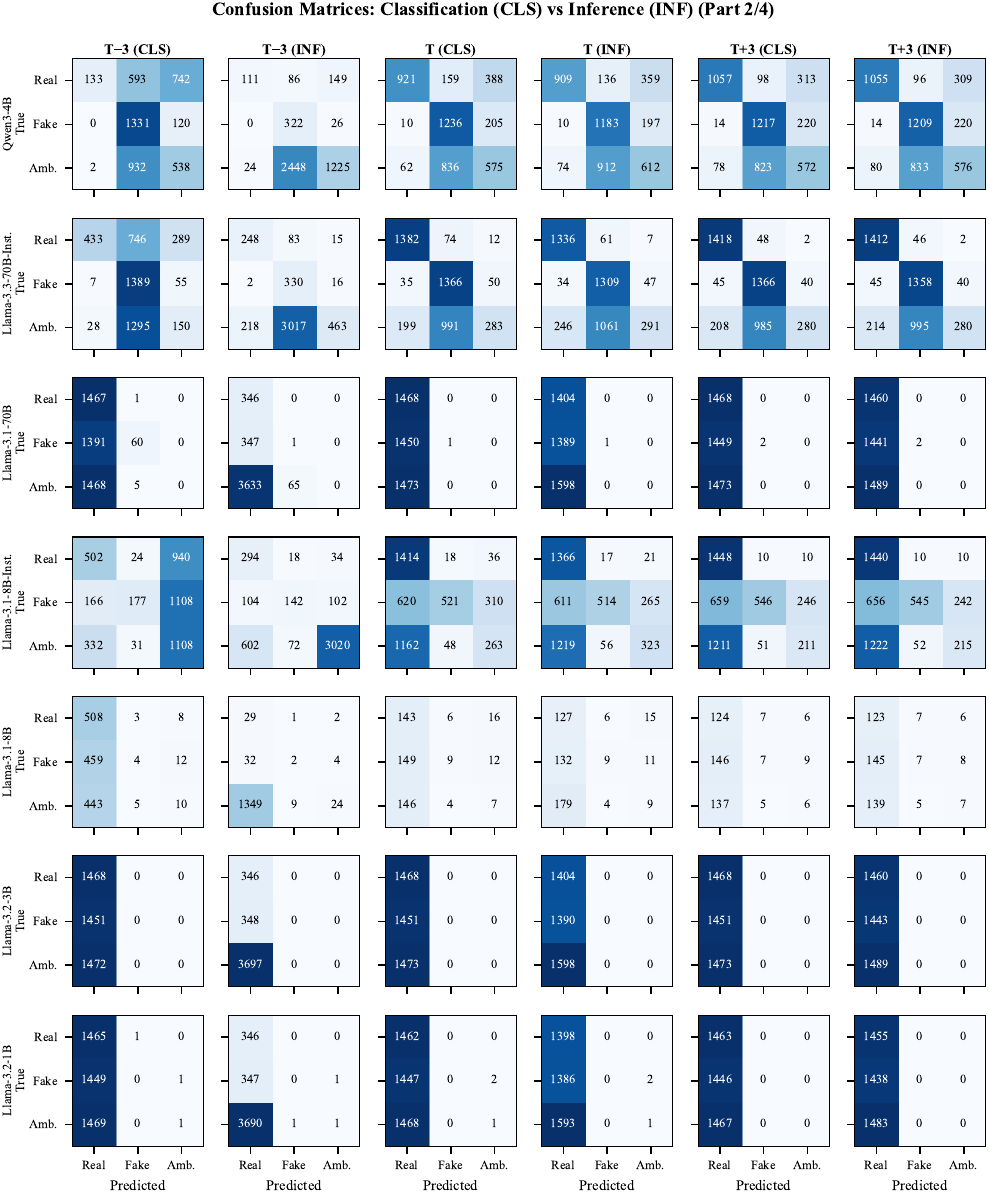}
    }
    \caption{Confusion Matrices (Part 2/4) - Qwen, Llama Series. Llama-3.1-70B, Llama-3.1-8B, Llama-3.2-3B, and Llama-3.2-1B exhibit "collapsing" behavior, predicting a single class.}
\end{figure*}

\begin{figure*}
    \ContinuedFloat
    \captionsetup{labelformat=continued}
    \centering
    \resizebox{1\textwidth}{!}{%
    \includegraphics[width=1\linewidth]{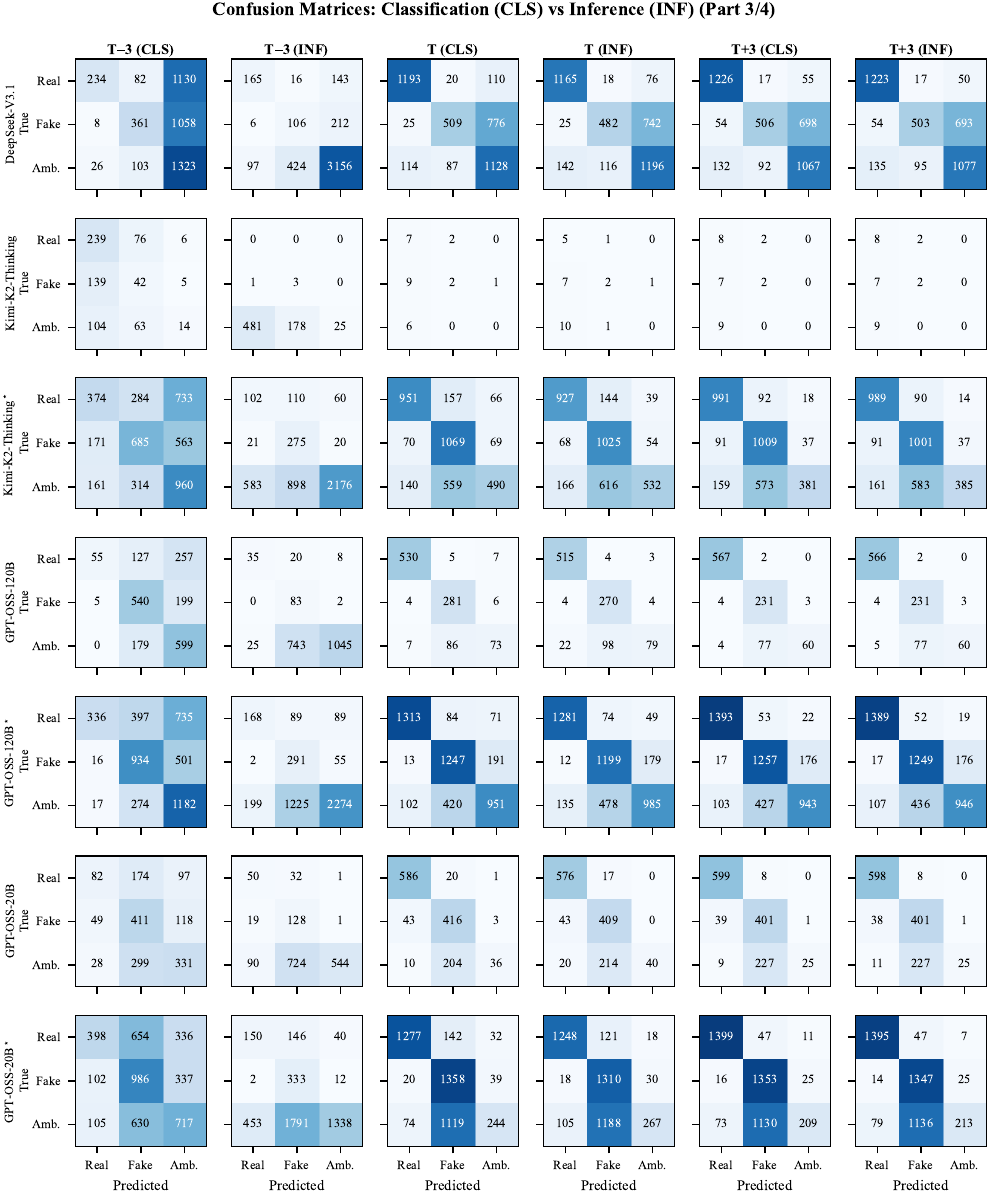}
    }
    \caption{Confusion Matrices (Part 3/4) - DeepSeek, Kimi, and GPT-OSS Series. The asterisk $^\star$ denotes the non-standard evaluation setting required to accommodate verbose CoT outputs.}
\end{figure*}

\begin{figure*}
    \ContinuedFloat
    \captionsetup{labelformat=continued}
    \centering
    \resizebox{1\textwidth}{!}{%
    \includegraphics[width=1\linewidth]{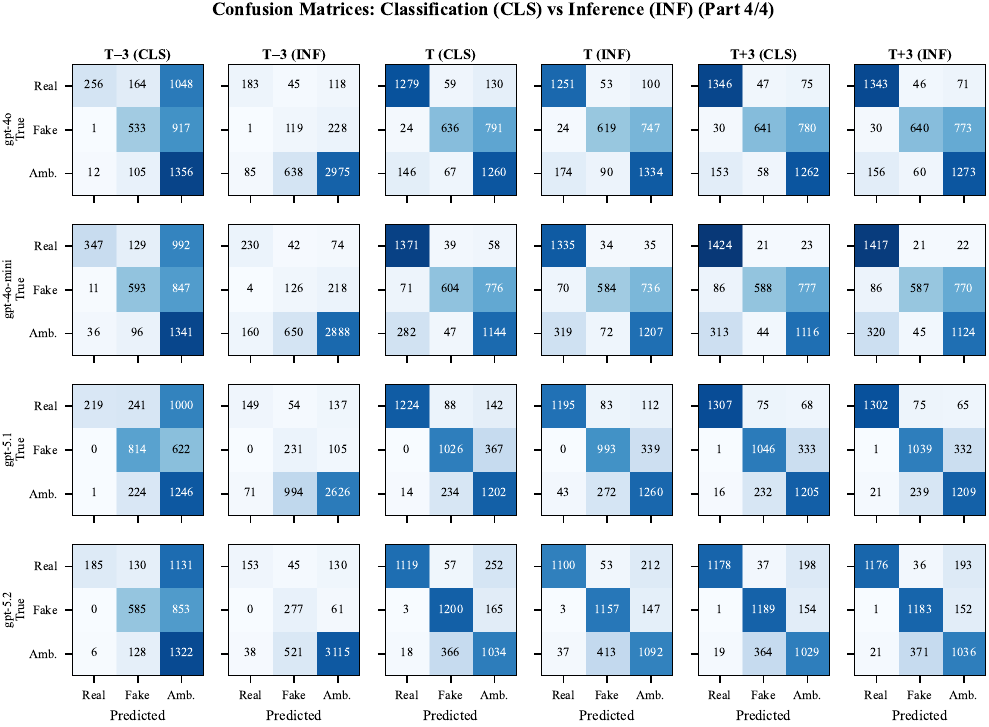}
    }
    \caption{Confusion Matrices (Part 4/4) - GPT Series.}
\end{figure*}

\end{document}